\documentclass{article}
\usepackage[
  left=2.2cm,
  right=2.2cm,
  top=2.0cm,
  bottom=1.8cm,
  includefoot,
  footskip=30pt,
]{geometry}
\counterwithin{figure}{section}
\usepackage{tikz}
\usetikzlibrary{matrix, arrows}
\usepackage{amsmath,amssymb}

\usepackage{dingbat}
\usepackage{amsthm}
\usepackage{mathtools}
\usepackage{xspace}
\usepackage[noend]{algorithmic}
\usepackage[ruled,vlined]{algorithm2e}
\usepackage{url}
\usepackage{makeidx}
\usepackage{enumerate}
\usepackage{epstopdf}
\usepackage{booktabs}
\usepackage{xltabular}
\usepackage{color}
\usepackage[utf8]{inputenc}
\usepackage{thm-restate}
\usepackage{scalerel,stackengine}
\usepackage[shortlabels]{enumitem}
\usepackage{xr}
\usepackage{fancyvrb}
\usepackage{xcolor}
\usepackage{bold-extra}
\usepackage[most]{tcolorbox}
\usepackage{fvextra}
\usepackage{float}
\usepackage{alltt}
\usepackage{soul}
\usepackage{fancyvrb}
\usepackage{multirow}
\usepackage[final]{hyperref}
\usepackage[bottom]{footmisc}
\usepackage{natbib}
\usepackage{linguex}
\usepackage{ulem}
\usepackage[T1]{fontenc}

\usepackage[all]{nowidow}
\usepackage{hyperref}

\usepackage{caption}
\renewenvironment{abstract}
  {\small
   \begin{center}
   {\large\bfseries \abstractname\vspace{-.5em}\vspace{0pt}}
   \end{center}
   \quotation}
  {\endquotation}

\usepackage{listings}
\let\oldbibliography\thebibliography
\renewcommand{\thebibliography}[1]{%
  \oldbibliography{#1}%
  \setlength{\itemsep}{2pt}%
  \setlength{\parskip}{0pt}%
}

\usepackage{tikz}
\usetikzlibrary{shapes,calc,positioning}

\usepackage{array}
\newcolumntype{L}[1]{>{\raggedright\let\newline\\\arraybackslash\hspace{0pt}}m{#1}}
\newcolumntype{C}[1]{>{\centering\let\newline\\\arraybackslash\hspace{0pt}}m{#1}}
\newcolumntype{R}[1]{>{\raggedleft\let\newline\\\arraybackslash\hspace{0pt}}m{#1}}

\begin{document}

\title{
\vspace*{-10pt}
\textbf{The Emergent Symbolic Structure \\ of Artificial Neural Networks}}

\author{\normalsize R. Thomas McCoy\textsuperscript{\hyperlink{affiliation}{1}} \\ \normalsize Yale University
\and \normalsize Paul Soulos\textsuperscript{\hyperlink{affiliation}{1}} \\ \normalsize Johns Hopkins University
\and \normalsize Tal Linzen\textsuperscript{\hyperlink{affiliation}{1}} \\ \normalsize New York University
\and \normalsize Paul Smolensky\textsuperscript{\hyperlink{affiliation}{1}} \\ \normalsize Microsoft Research
}

\date{}
\maketitle

\begingroup
\renewcommand{\thefootnote}{}
\footnotetext{%
  \hypertarget{affiliation}{\textsuperscript{1}At the project's start, all authors were affiliated with Johns Hopkins, and Smolensky was also affiliated with Microsoft Research. For parts of the project, McCoy was affiliated with Princeton or with Yale and Microsoft. Soulos's current affiliation is Microsoft.}}
\addtocounter{footnote}{1}
\endgroup

\vspace{-11pt}

\begin{abstract}
\normalsize

\noindent
Modern systems in artificial intelligence (AI) somehow excel in domains for which they seem 
poorly suited.
Intelligence has traditionally been modeled as operating over structured combinations of symbols, such as logical formulas.
However, the strongest modern AI systems are based on neural networks, which instead represent information in continuous vectors.
Vectors seem inadequate for capturing the structure of language, logic, and other cognitive domains, yet neural networks achieve impressive performance in these areas. How do they do it?
In this work, we propose a potential answer: Despite appearances, perhaps the internal representations of neural networks implicitly realize symbolic structure. In support of this hypothesis, we show that the vector representations of a variety of neural networks can be closely approximated with symbolic structures: we can replace the network's entire representation-generating process with a closed-form equation instantiating a symbolic structure, and the network's behavior remains largely unchanged.
This finding holds for both small-scale neural networks trained to manipulate lists as well as large language models (LLMs) operating in four domains that are central in symbolic traditions: arithmetic, logic, computer code, and language. Further, our symbolic approximation allows us to modify an LLM’s behavior in targeted ways via precise interventions on its internal representations, showing that the LLM's behavior is reliant on the symbolic structures we have identified.
This work provides a potential way to reconcile longstanding symbolic conceptions of intelligence with the vector-based nature of modern AI.

\end{abstract}


\setlength{\Exlabelwidth}{0.7em}
\setlength{\SubExleftmargin}{1.35em}
\setlength{\SubSubExleftmargin}{1.5em}
\renewcommand{\firstrefdash}{} 
\renewcommand{\secondrefdash}{.} 

\section{Introduction}

\begin{figure}
    \centering
    \includegraphics[width=0.95\linewidth]{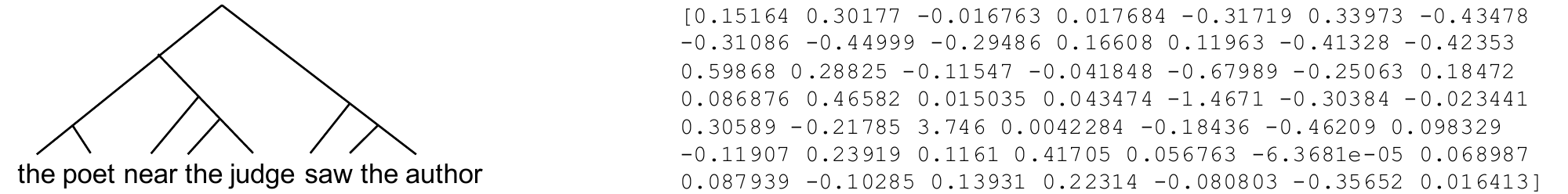}
    \caption{The apparent incompatibility of neural and symbolic representations. \textbf{Left:} A symbolic sentence representation, specifically a (simplified) syntax tree. \textbf{Right:} A neural-network sentence representation, specifically a continuous vector. These two representations encode the same sentence, 
    but they appear starkly different from each other.
    Nonetheless, neural networks are the state of the art in many areas traditionally viewed as symbolic. 
    In this work, we study how vector representations can support such strong performance in domains that appear to require symbolic structure.
    }
    \label{fig:sentence_tree_vector}
\end{figure}

In both AI and cognitive science, many domains of intelligence have traditionally been modeled using symbolic systems, in which discrete units (symbols) are combined in structured ways \citep{boole1854investigation,turing1950computing,mccarthy1959programs,chomsky1965aspects,newell1972human,Quilty-Dunn_Porot_Mandelbaum_2023}. 
For instance, in linguistics, sentences are typically represented with trees composed of discrete words
(Figure~\ref{fig:sentence_tree_vector}, left). Symbolic analyses have a long history: P\=a\d{n}ini developed symbolic theories of language around 500 BCE \citep{kiparsky1993panini}, Aristotle formalized symbolic logic around 350 BCE \citep{smith1989prior}, and Ada Lovelace used symbol manipulation as the basis of the first computer program in 1843  \citep{bowden1953faster}. 

More recently, however, major advances in AI have been driven by neural networks \citep{rumelhart1986general}---computational systems that 
encode information in a very different way: using continuous vectors (Figure~\ref{fig:sentence_tree_vector}, right). 
Such vectors seem qualitatively different from symbol structures, yet neural networks excel in domains traditionally viewed as symbolic, including language \citep{brown2020language}, mathematics \citep{glazer2024frontiermath}, and coding \citep{li2022competition}. 
How can systems that use vector representations perform so well in symbolic domains? 
Answering this question is an important goal for mechanistic interpretability \citep{olah2018building,elhage2021mathematical,marks2025sparse}, the growing area of research that aims to understand the inner workings of AI systems such as large language models (LLMs).

The hypothesis we investigate is that, although the \textit{explicit} structure of neural networks appears very different from symbol systems, perhaps they \textit{implicitly} realize symbolic structure inside their vector representations. 
To test this hypothesis, we use an analysis method called DISCOVER (DISsecting COmpositionality in VEctor Representations) which aims to approximate a neural network’s vector representations with vectors that explicitly encode symbolic structure.

This approach requires a proposal about how vectors could capture such structure---a consideration that is far from trivial given the apparent incompatibility of vectors and symbols. For this purpose, we adopt a mathematical formalism from cognitive science called Tensor Product Representations \citep[TPRs;][]{smolensky1990tensor}. In a TPR, a symbolic structure is framed as a collection of \textbf{fillers}---the elements of the structure---each of which is paired with a \textbf{role} that denotes its position. 
For instance, to encode the sentence \textit{cats chase dogs}, we might use the fillers \texttt{cats}, \texttt{chase}, and \texttt{dogs} paired with the roles \texttt{subject}, \texttt{verb}, and \texttt{object}, respectively. The collection of role-filler pairs can then be translated into a vector using an approach described in Section~\ref{sec:tprs}. 
TPRs were introduced as a proposal about how neural networks could be designed to capture symbolic structure. Our hypothesis is that, even without being explicitly designed in this way, standard neural networks naturally converge to representations that are structured as TPRs.

We apply DISCOVER to a variety of neural networks and find that their representations can be closely approximated by the type of symbolic structure that we have hypothesized. Specifically, \textbf{we can replace each neural network's entire representation-generating process with a single, interpretable, TPR-based equation, and the network's behavior changes minimally if at all}.
Our core findings are the following:
\begin{enumerate}[itemsep=0pt]
    \item In analyses of models trained on synthetic sequence-manipulation tasks, we find emergent symbolic structure across all three classes of neural networks that we consider---multi-layer perceptrons \citep{rosenblatt1962principles}, recurrent neural networks \citep{elman1990finding}, and Transformers \citep{vaswani2017attention}. Thus, this type of emergent structure is general enough to hold across several different classes of neural systems.
    \item We also analyze seven LLMs: 
    Gemma-3-27b \citep{team2025gemma}, GPT-2-XL \citep{radford2019language}, GPT-OSS-20b \citep{agarwal2025gpt}, Pythia-12b \citep{biderman2023pythia}, Qwen3-14b \citep{yang2025qwen3}, OLMo-2-13B \citep{walsh2025}, and Llama-3.1-8b \citep{grattafiori2024llama}.
    We find emergent symbolic structure in all of these LLMs, showing that such structure arises in large-scale systems trained on naturalistic data. 
    \item We conduct an extensive analysis of one LLM (GPT-OSS) performing tasks in four domains long viewed as symbolic (arithmetic, logic, coding, and language). Its representations are well-approximated with TPRs, providing evidence that LLMs draw on implicit symbolic structure when performing symbolic tasks.
    \item Our analyses enable us to intervene on neural network representations in ways that lead the network’s behavior to change in the expected ways, showing that the representational structure we have identified is implicated in model behavior. For instance, if a network's input is \textit{the clever doctor helped the lawyer}, we can edit the representation by changing the position of \textit{clever} from ``subject adjective'' to ``object adjective.'' The network then behaves as if the input had been \textit{the doctor helped the clever lawyer}. 
    \item Our analyses can generalize to new combinations of fillers and roles. E.g., if the DISCOVER approach is trained on sentences that include the word \textit{scientist} but never occurring in the role of ``subject of the sentence'', DISCOVER then successfully generalizes to sentences with \textit{scientist} as the subject. This finding shows that the networks we analyze are systematic in how they compose fillers with their roles.
\end{enumerate}
These findings support the hypothesis that, at least in many cases, the vector representations of standard neural networks have implicit symbolic structure. 
Though advances in neural networks make it tempting to think that symbolic structure is unnecessary, these results instead suggest that such structure is important for intelligent behavior, since even systems that are not inherently symbolic learn to use symbolic structure on their own.\footnote{A partial codebase is available at \url{https://github.com/tommccoy1/discover/}. The complete code will be made available soon (pending employer approval).}

\section{Background}

\subsection{Understanding the representations of neural networks}

Though neural networks are widely used, 
the internal mechanisms that drive their behavior remain poorly understood.
An area of research called \textit{interpretability} or \textit{mechanistic interpretability} aims to address this issue by figuring out how neural networks operate internally.
Our work is situated in the side of interpretability that analyzes the vectors that serve as a network's representations.
Prior work has made important progress in this area
by identifying how interpretable features are encoded inside these vectors \citep{elman1990finding,bau2017network,adi2017fine,rogers2020primer,ravfogel2021counterfactual,hernandez2021low,elhage2022toy,nanda2023emergent,engels2025not,todd2024function,bhalla2026sparse}. For instance, one analysis identified a vector component that encoded the concept ``Golden Gate Bridge,'' such that emphasizing this feature led the network to fixate on the Golden Gate Bridge \citep{templeton2024scaling}. 

Most prior representational analyses can be unified under a proposal called the Linear Representation Hypothesis \citep{park2024the}. Under one framing of this proposal, each vector inside a neural network is assumed to capture some set of concepts $\{c_1, c_2, ..., c_n\}$; each possible concept $c_i$ has a vector $e(c_i)$  that encodes it, and the overall representation is the sum\footnote{Some approaches---e.g., standard sparse autoencoders---instead use a weighted sum of concept vectors, where the weights might correspond to, e.g., the model's confidence that the concept is present. For this discussion we stick with an unweighted sum because we are analyzing settings in which it is reasonable to assume that the relevant concepts are either fully present or fully absent.} of these concept vectors: $\sum_i e(c_i)$.  
For example, a representation for the sentence \textit{dogs jump} might encode two concepts---\textit{dogs} and \textit{jump}; if the concept vector for \textit{dogs} is $e(dogs)$ = [2.0, --0.3] and the concept vector for \textit{jump} is $e(jump) = $ [0.6, 1.8], then the encoding for \textit{dogs jump} would be [2.6, 1.5]. The Linear Representation Hypothesis has been shown to unify a range of popular representational interpretability methods including additive analogies \citep{mikolov2013linguistic}, linear probes \citep{ettinger2016probing,alain2017understanding,hupkes2018visualisation,belinkov2022probing}, and sparse autoencoders \citep{cunningham2023sparseautoencodershighlyinterpretable,templeton2024scaling}, each of which has significantly advanced our understanding of neural representations.

\subsection{Structure and the binding problem}\label{sec:bindingproblem}

The Linear Representation Hypothesis has supported important progress in understanding AI systems, but it faces a challenge for explaining situations where structure matters---that is, when the system must consider not only which concepts are present but also how they are arranged. Consider the sentence \textit{cats chase dogs}. Under the framing described in the previous paragraph, we would have $e(\textit{cats chase dogs}) = e(cats) + e(chase) + e(dogs)$. 
The problem is that addition is not sensitive to order, so \textit{dogs chase cats} would receive the same encoding as \textit{cats chase dogs} even though these sentences mean different things. 
Structure is important across the domains that are traditionally analyzed as symbolic. In math, $4 - 7$ means something different from $7 - 4$; in logic, \textit{If \texttt{P}, then \texttt{Q}} means something different from \textit{If \texttt{Q}, then \texttt{P}}; in computer code, \texttt{concatenate(x, y)} means something different from \texttt{concatenate(y,x)}; and in language---as noted by Lewis Carroll (\citeyear{carroll1865alice})---\textit{I breathe when I sleep} means something different from \textit{I sleep when I breathe}. 
Therefore, approaches that do not capture such structure are missing an important part of how AI systems can succeed in these domains.

Addressing this issue requires some way to indicate which position each element occupies (e.g., indicating that \textit{cats} is in the subject position while \textit{dogs} is in the object position). This goal is at the heart of a long line of literature in cognitive science, where the issue goes under the name of the binding problem \citep{von1981correlation,von1999binding,treisman1999solutions,roskies1999binding,greff2020binding}: how can a neural network bind together different pieces of information such as features and positions? (In cognitive science, the neural network in question is usually the brain rather than an artificial neural network.) 
As a hypothesis about how AI systems encode structure, the current paper adopts one prominent solution to the binding problem: the Tensor Product Representation formalism.

\subsection{Tensor Product Representations}\label{sec:tprs}

\begin{figure}[]
    \centering
    \includegraphics[width=\textwidth]{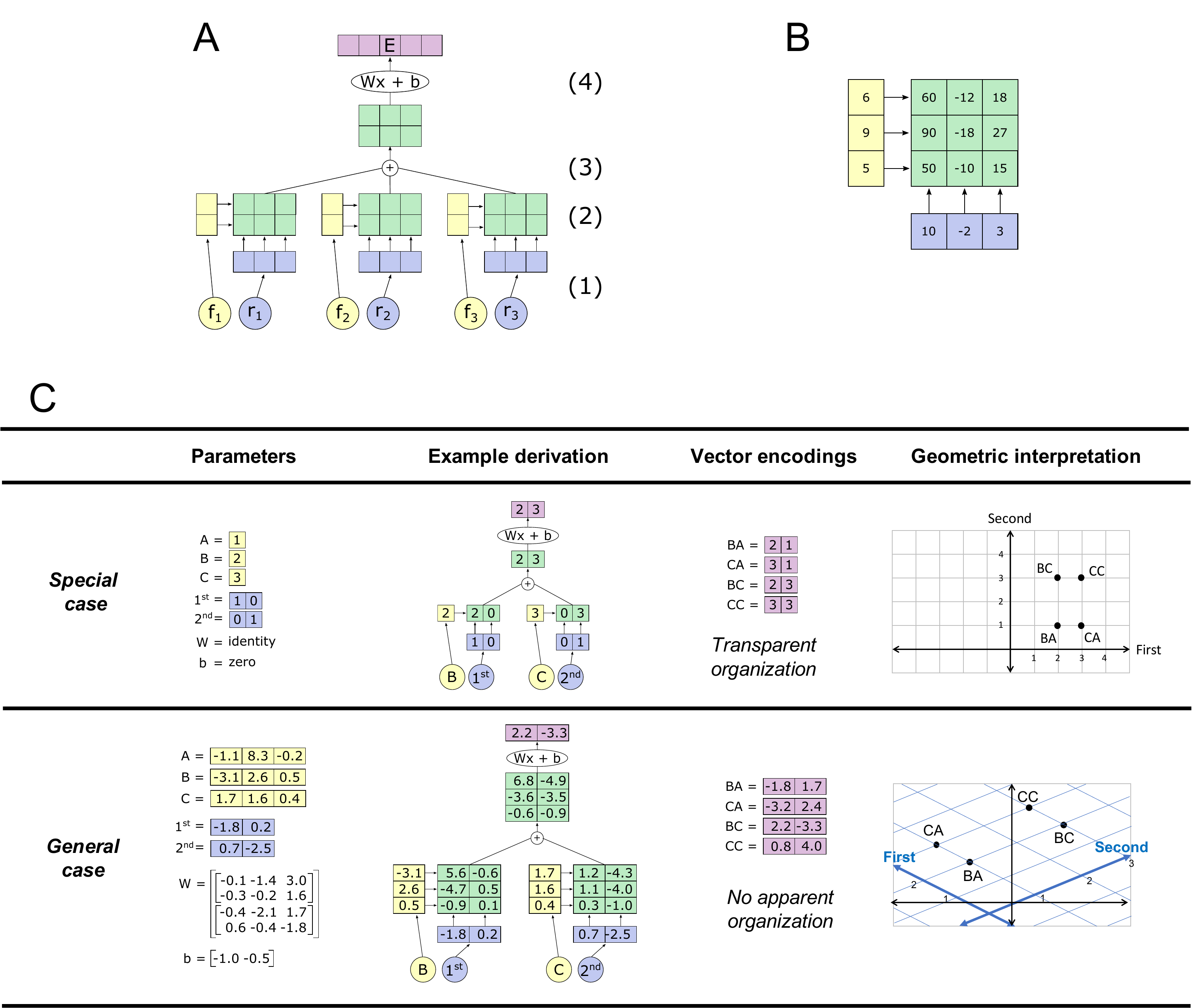}
    \caption{Linearly-transformed Tensor Product Representations (TPRs). \textbf{A.} Structure of a linearly-transformed TPR. \textit{(1)}. Each filler and each role is encoded with a vector that stands for it. \textit{(2)} Each filler and its corresponding role are bound together using the tensor product, illustrated in \textbf{B}. \textit{(3)} The tensor products are summed. \textit{(4)} A linear transformation is applied to give the final encoding, $E$. \textbf{C.} Illustrations of TPRs used to encode pairs of letters with the roles $1^{\text{st}}$ and $2^{\text{nd}}$. In the special case, the encodings have a human-interpretable structure both when construed as lists of numbers and when construed as points in space. In the general case, the list of numbers is not readily interpretable, but the geometric arrangement remains intact (in a stretched, rotated, and translated coordinate system), illustrating how vectors can have systematic compositional structure even if no such structure is apparent to human observers. The example TPRs shown here are only two-dimensional for ease of visualization, but most of the TPRs we deal with in this paper are much larger (with hundreds or thousands of dimensions).}
\label{fig:tpr_structure}
\end{figure}

Tensor Product Representations \citep[TPRs;][]{smolensky1990tensor} are a proposal about how symbolic structure could be realized in vector space. In a TPR, a symbol structure is viewed as a set of fillers (the elements of the structure) paired with roles (the positions of those elements). For instance, the fraction $\frac{3}{7}$ uses the fillers $3$ and $7$ in the roles of \textit{numerator} and \textit{denominator}. This role-filler structure is then translated into vector space (Figure~\ref{fig:tpr_structure}A): First, each filler has some vector that encodes it (e.g., the filler \textit{3} would be encoded with a vector $f_{3}$), and similarly for the roles (e.g., the role of \textit{numerator} would be encoded with a vector $r_{numerator}$); these vectors can also be referred to as filler embeddings and role embeddings. Each filler vector is combined with the corresponding role vector using the tensor product (Figure~\ref{fig:tpr_structure}B), which takes in the two vectors and returns a matrix that encodes the role-filler pair (e.g., a matrix that encodes ``\textit{3} in the role of \textit{numerator}''). The matrices for all role-filler pairs are summed to give an encoding of the entire structure.\footnote{Such role-filler pairings could be captured under the Linear Representation Hypothesis as described above by using a separate concept for each role-filler pair (e.g., \textit{cats-as-subject} and \textit{cats-as-object}) such that the representation of \textit{cats chase dogs} might be $e(\textit{cats-as-subject}) + e(\textit{chase-as-verb}) + e(\textit{dogs-as-object})$. 
However, it is unlikely that LLMs encode information in this way because it requires an independent representation for every possible combination of a word and a position, which would require the LLMs to have learned a very large concept vocabulary.
As further reason to doubt such an account, see Section~\ref{sec:ood} for empirical evidence that LLMs do not encode role-filler pairs in an atomic way but rather have systematic strategies for combining roles and fillers.} 
We use a variant of TPRs that we call linearly-transformed TPRs, in which the summed matrix $x$ is passed through an affine transformation $x \mapsto Wx + b$ to resize and reshape the matrix into a vector, 
which serves as our final encoding of the structure.\footnote{
In linear algebra terminology, linear transformations cannot include a bias term $b$, making this transformation affine but not linear.
Our use of the term \textit{linearly-transformed} follows calculus terminology, in which linear functions can include bias terms.}
The overall representation is then $W (\sum_i f_i \otimes r_i) + b$. See Appendix~\ref{app:need_for_linear_transformation} for why we use linearly-transformed TPRs rather than basic TPRs; in brief, the linear transformation allows for superficial changes to the vector space that do not meaningfully change the TPR properties that are relevant to our work.

Figure~\ref{fig:tpr_structure}C provides some intuition for how the role-filler structure is captured in vector space when we encode letter pairs as linearly-transformed TPRs. 
The ``special case'' shows a situation in which TPRs have clear human-interpretable organization: each letter is encoded with a number (A=1, B=2, C=3) and the TPR for the letter pair is simply a length-2 vector containing the encodings for the two letters in order---e.g., the vector [3,1] for \texttt{CA}. This special case arises when $W$, $b$, and the role and filler vectors have a trivial structure.
When these components are non-trivial, the TPR vector space becomes stretched, rotated, and translated in ways that prevent the vectors from having obvious structure when viewed as lists of numbers. 
However, the vector space still displays a consistent geometric organization---the stretching, rotation, and translation impede human understanding, but they do not necessarily alter the basic structure of the vector space (particularly the structure that is accessible to neural networks, since neural networks can straightforwardly undo these transformations).
This illustrates how vectors can have systematic structure even when humans cannot readily observe that structure.

TPRs resolve the binding problem by providing a principled way to pair fillers with roles. Given a TPR, it is possible to determine which filler occupies each role through a simple linear procedure called unbinding, a procedure which is exact when the role vectors are linearly independent; in practice, unbinding can still be possible with high accuracy even when linear independence does not hold \citep{haley2020invertible}. 

Thus, TPRs are one way in which symbolic structure can be realized in vector space (though see Section~\ref{sec:other_vsas} for other formalisms that achieve the same goal). Our core hypothesis is that this strategy is the one implicitly used by neural networks when they perform symbolic tasks; the rest of this paper focuses on testing this hypothesis.

\section{The central technique: DISCOVER}

To test the hypothesis that neural networks implicitly use TPR structure, we apply an analysis method that we call DISCOVER (\textbf{DIS}secting \textbf{CO}mpositionality in \textbf{VE}ctor \textbf{R}epresentations). To make the DISCOVER procedure concrete, we first introduce a specific neural network that we will analyze as a running example and then describe DISCOVER by showing how it applies to this running example. In later sections, we then apply DISCOVER to a broad range of neural networks. 
We only describe this running example at a high level; see the next section and the associated Appendix~\ref{app:letter_sequence_technical_details} for more details.
DISCOVER was introduced by \citet{mccoy2018rnns}, who used it to analyze recurrent neural networks; the current paper extends DISCOVER to a much broader range of models, including large-scale LLMs. See Section~\ref{sec:relatedwork} for more discussion of the history of DISCOVER and its relation to other analysis methods.

\subsection{Running example: A recurrent neural network reversing a list}

As our running example, we consider a neural network that is trained to reverse lists of letters (e.g., turning \texttt{Q M Z} into \texttt{Z M Q}).
The type of neural network that we use is a sequence-to-sequence recurrent neural network \citep{sutskever2014sequence}, which is made of two sub-components called the encoder and the decoder, each of which is a gated recurrent unit (GRU) network \citep{cho2014learning}. The encoder  takes in the input sequence and converts it to a single vector $E$; the decoder then takes in $E$ and uses it to produce the output. 

We train 10 copies of this type of network. At the end of training, all copies achieve an accuracy of 100\% on a test set made of items that were not seen by the models during training. Therefore, these networks have clearly succeeded at learning the task of reversal, meaning that the encoding vector $E$ must encode all the information that the decoder needs to produce the output. Although reversal is a simple task, performing it requires $E$ to capture both which letters appear in the input and which positions they appear in. The goal of our analysis will be to understand how $E$ captures this information; our hypothesis is that it does so by using TPR structure. 

\subsection{The DISCOVER procedure}

\begin{figure}
    \centering
    \includegraphics[width=0.5\linewidth]{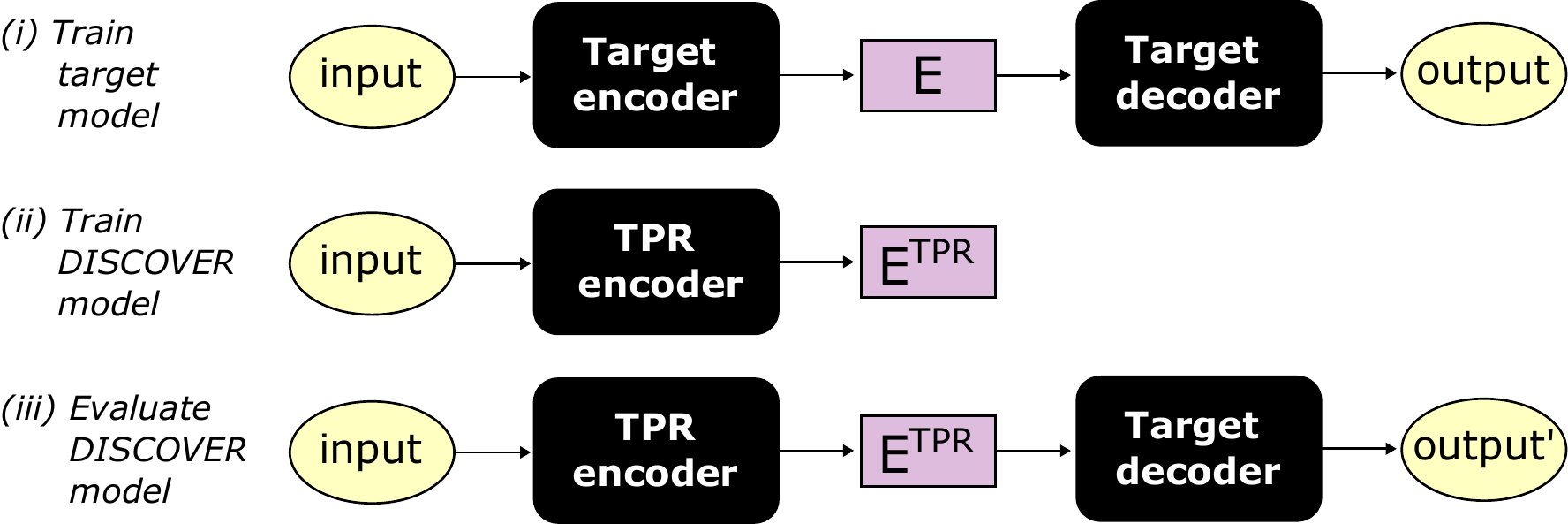} \hspace{1.5cm}
    \includegraphics[width=0.3\textwidth]{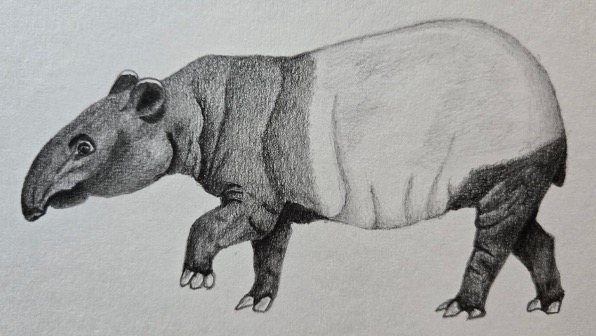}
    \caption{\textbf{Left:} The DISCOVER process. (i) We train the target model, whose encodings $E$ we wish to understand. (ii) We then train a DISCOVER model, which is explicitly structured as a linearly-transformed TPR, so that, for each input, its encoding $E^{\text{TPR}}$ is as close as possible to the target model's encoding $E$. (iii) We then evaluate the DISCOVER model by plugging its encoding $E^{\text{TPR}}$ into the decoder of the original target model, to see if the target model still produces the right output when provided with $E^{\text{TPR}}$ instead of $E$. 
    \textbf{Right:} A tapir (\textit{Tapirus indicus}). Tapirs are the mascot of DISCOVER because the word \textit{tapir} can be created by interleaving \textit{TPR} with \textit{AI}. Tapir image \copyright~2026 by Victoria E. McCoy under CC BY-SA 4.0 (\url{https://creativecommons.org/licenses/by-sa/4.0/}).
    }
    \label{fig:discover_process}
\end{figure}

To test the hypothesis that a network's encodings are \textit{implicitly} structured as TPRs, we see whether those encodings can be approximated using vectors that are \textit{explicitly} TPRs (Figure~\ref{fig:discover_process}); this is the procedure that we refer to as DISCOVER. That is, we create a neural network that is explicitly structured as a linearly-transformed TPR, and we train it so that the encodings it produces are as close as possible (i.e., minimizing mean squared error) to the encodings of the \textbf{target model}---the black-box model that we are trying to understand. The free parameters in this TPR model are the vectors that it uses to represent fillers and roles, and the parameters $W$ and $b$ that define its linear transformation;
thus, training the TPR model amounts to finding values for these parameters.
Once the TPR model is done training, we evaluate the quality of the TPR approximations by feeding them into the target model's decoder. This manipulation amounts to replacing the target model's \textbf{entire encoding process} with a closed-form TPR expression that comes from DISCOVER. If the target model continues to produce the correct answer when fed the TPR approximations, we conclude that the TPR has succeeded at approximating the target model's encodings, since the TPR approximations are functionally equivalent to these target encodings. 
This criterion for success is quantified with the metric of \textit{approximation accuracy}, which is the proportion of test-set examples on which the target model's decoder produces the entire correct output sequence when given a DISCOVER model's TPR-based encoding of the input.

\subsection{Role schemes}\label{sec:running_example_role_schemes}

To train our DISCOVER model to approximate the encoding $E_S$ for some sequence $S$, we need to have a hypothesis about what fillers and roles are present in $S$. We assume that the fillers are the letters that are present in the sequence; e.g., if the sequence is \texttt{Q M Z}, the fillers would be the letters \texttt{Q}, \texttt{M}, and \texttt{Z}. The roles, however, are trickier: there are many reasonable hypotheses about what sorts of positions could be used as roles. For this experiment, we will consider 5 possible types of roles (which we refer to as \textbf{role schemes}):
\begin{itemize}[itemsep=0pt]
    \item \textbf{Left-to-right:} Each letter's role is its left-to-right position (\textit{first}, \textit{second}, etc.)
    \item \textbf{Right-to-left:} Each letter's role is its right-to-left position (\textit{last}, \textit{second-to-last}, etc.)
    \item \textbf{Bidirectional:} Each letter's role is the concatenation of its left-to-right and right-to-left position (e.g., (0,3) means ``1st-from-left-and-4th-from-right''). Prior work has found that humans encode letter positions in a bidirectional way in the spellings of words \citep{fischer2010representation}.
    \item \textbf{Wickelroles:} Each letter's role indicates what letter appears before it and what letter appears after it \citep{wickelgren1969context}. This role scheme was used by \citet{rumelhart1986learning} in their influential neural-network model of learning the English past tense.
    \item \textbf{Bag-of-words:} All letters are given the same role as each other.
\end{itemize}
See Figure~\ref{fig:role_schemes_tpe_acc_gru_rev} (left) for an example of each role scheme.

\subsection{Results}\label{sec:running_example_results}

\begin{figure}
    \centering
    \begin{minipage}[c]{0.7\textwidth}
    \begin{tabular}{ccccc} \toprule
        & Q & M & Z & C \\ \midrule
        Left-to-right & first & second & third & fourth  \\
        Right-to-left & fourth-to-last & third-to-last & second-to-last & last \\
        Bidirectional & (0,3) & (1,2) & (2,1) & (3,0) \\
        Wickelroles & \#\_M & Q\_Z & M\_C & Z\_\# \\
        Bag-of-words & present & present & present & present \\ \bottomrule
    \end{tabular}
    \end{minipage}
    \begin{minipage}[c]{0.25\textwidth}
    \includegraphics[width=\linewidth]{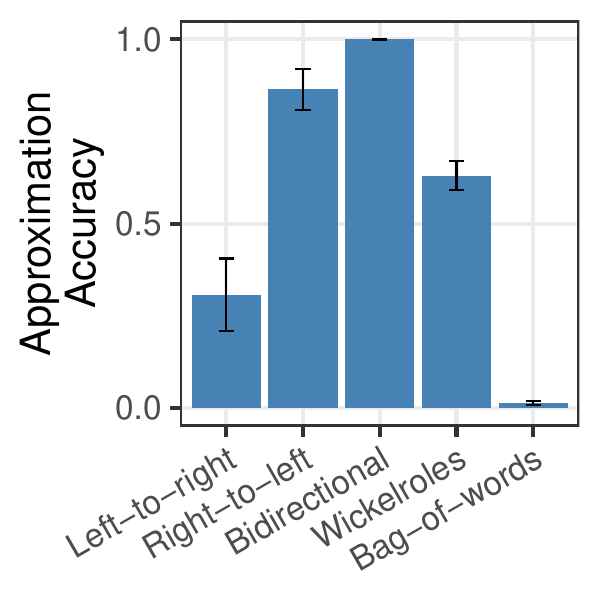}
    \end{minipage}
    \caption{\textbf{Left:} The role schemes that we use for analyzing letter-sequence-processing neural networks, with an example of the roles assigned by each role scheme to the sequence \texttt{Q M Z C}.
    \textbf{Right:} Performance of DISCOVER with various role schemes at approximating the encodings of GRUs trained to reverse letter sequences. The bars show the mean across 10 reruns, with error bars showing two standard deviations above and below the mean.}
    \label{fig:role_schemes_tpe_acc_gru_rev}
\end{figure}

Figure~\ref{fig:role_schemes_tpe_acc_gru_rev} (right) shows the results. A TPR using a \texttt{bidirectional} role scheme provides perfect or close-to-perfect accuracy at approximating the target model in all 10 reruns of the model (the lowest approximation accuracy across the 10 reruns is 99.98\%). This provides strong evidence that the representations of this model indeed have symbolic structure of the sort realized in a TPR (specifically, a \texttt{bidirectional} TPR). 

Among the other role schemes, the \texttt{right-to-left} one does nearly as well as \texttt{bidirectional}, while \texttt{left-\allowbreak to-\allowbreak right} provides a poor approximation. This asymmetry makes sense: intuitively, right-to-left positions are more important for the task of reversal than left-to-right positions. 
The \texttt{bag-of-words} role scheme is essentially a null hypothesis corresponding to a lack of structure; the fact that it performs so poorly validates the assumption that the target model indeed encodes structure.

\subsection{Notes on the scope and interpretation of DISCOVER}\label{sec:discover_comments}

With the DISCOVER method described, there are several points about it that are now worth making.

First, although the models we analyze are black-box neural networks that are notorious for being challenging to understand, DISCOVER produces an interpretable closed-form equation for the representations of these models. Thus, providing a DISCOVER approximation of a network's representations substantially increases the extent to which the structure of those encodings is human-understandable.

Second, there is no guarantee that a given neural network can be approximated by DISCOVER. The neural architectures that we analyze are complex, nonlinear systems that (under certain idealizations) can approximate any function \citep{hornik1989multilayer,siegelmann1992computational,perez2021attention}. In contrast, TPRs have a simple, bilinear structure, meaning that (once the role scheme has been specified) a TPR is highly constrained and can only implement a narrow class of functions.\footnote{E.g., any linearly-transformed TPR using left-to-right roles must obey the property that $TPR(\texttt{Q M Z}) - TPR(\texttt{Q S}) + TPR(\texttt{V S}) = TPR(\texttt{V M Z})$---a property we leverage in the experiments in Section~\ref{sec:causal_interventions}. Such constraints severely limit the set of functions that may be implemented by a TPR with a given role scheme.} As an illustration of how constrained TPRs are, a left-to-right role scheme is a completely reasonable way to encode a sequence, yet TPRs with this role scheme achieved a low approximation accuracy in Figure~\ref{fig:role_schemes_tpe_acc_gru_rev} (right).
The fact that DISCOVER is not at all guaranteed to work means that, when it does work, its success is meaningful.

Third, a successful DISCOVER result indicates that the DISCOVER model captures all information that is present in the target model, but it does not necessarily indicate that the target model captures all information that is present in the DISCOVER model.
Suppose the target model implements TPRs with left-to-right roles. This target model could be perfectly approximated by a DISCOVER model using bidirectional roles: bidirectional roles are strictly more powerful than left-to-right roles, so a bidirectional role scheme can be realized in a degenerate way that is equivalent to a left-to-right role scheme.
To use DISCOVER to identify \textit{only} what information is present in the target model and nothing more, one would need to (e.g.) pare down the successful DISCOVER model until nothing more could be removed without harming its approximation accuracy. 
Such paring down is an important direction, but we do not pursue it here because our goal is to test the broad hypothesis that neural networks use implicit TPRs, rather than aiming to conclusively identify \textit{which} TPRs are used. We consider this goal to be achieved as long as we can find any TPR that closely approximates a target model, even if that TPR is not as simple as it strictly could be. For more discussion, see Section~7.1.6 of \citet{mccoy2022implicit}.

Fourth, the version of DISCOVER that we use in this paper is supervised---it requires the human experimenter to make a hypothesis about which role scheme to search for.
This aspect of DISCOVER could be made unsupervised by having the DISCOVER model form its own hypothesized role schemes, and indeed \citet{soulos2020discovering} applied such an approach successfully. 
We have deliberately chosen to stick with the supervised version of DISCOVER. To motivate why, it is useful to note that understanding a network's representations encompasses (at least) two goals: determining \textit{what} features are used in the representations, and determining \textit{how} those features are represented. Our goal in this work is providing an answer to the \textit{how} question; to isolate this question, it is beneficial to minimize uncertainty about the \textit{what} question, which we do by using settings where our understanding of the domain is sufficient for us to make clear hypotheses about the role scheme, making it possible to use supervised DISCOVER and thereby sidestep the substantial complications that arise with unsupervised feature discovery in the context of neural network interpretability \citep{conerly2024update,gao2025scaling}. 

Finally, our goal with DISCOVER is to understand the structure of a neural network's representations. We are not attempting to explain how the network produces vectors with this structure (e.g., we are not claiming that neural networks actually compute the tensor product of filler and role vectors), though this is an important direction for future work. As a rough analogy, our goal is similar to that of a biologist characterizing the structure of a complete bird feather, not that of a biologist studying how feathers grow.

\section{Letter sequence models}\label{sec:letter_sequence_models}

The case study in the previous section showed that TPR-structured representations arise in at least one type of neural network (the GRU architecture) for at least one task (reversing a list). 
Is TPR structure a quirk of this one particular setting, or is it a more general property of how neural networks represent information?
Motivated by this question, we extend our DISCOVER analyses to four neural architectures trained on three different tasks.

\subsection{Tasks}

Each neural network that we analyze is trained to perform some operation on an input list of letters. Specifically, these sequence-manipulation tasks are the following:
\begin{itemize}[itemsep=0pt]
    \item Copying: Returning the input list unchanged. E.g., \texttt{Q M Z V R} $\rightarrow$ \texttt{Q M Z V R} 
    \item Reversing: Reversing the input list. E.g., \texttt{Q M Z V R} $\rightarrow$ \texttt{R V Z M Q} 
    \item Interleaving: alternating letters from the start and end of the list until all letters are used up. \\ E.g., \texttt{Q M Z V R} $\rightarrow$ \texttt{Q R M V Z} 
\end{itemize}
We chose these tasks because all of them require representing which letters appear in the input and where they appear. Thus, if a neural network succeeds at learning one of these tasks, it must be representing the input's list structure, making it natural to then use DISCOVER to test hypotheses about \textit{how} it encodes that structure.

For these tasks, each letter sequence could have a length from 1 to 6 (inclusive). Each list element could be any one of the 26 capital letters in the Roman alphabet; letters were allowed to repeat within a list. For each task, we generated 57,000 unique input-output pairs, randomly divided into three subsets: the training set (50,000 examples), the validation set (2,000 examples), and the test set (5,000 examples).

\subsection{Neural network architectures}

We trained four different types of neural networks on these tasks: multi-layer perceptrons \cite[MLPs;][]{rosenblatt1962principles}; GRUs \citep{cho2014learning}, which are a type of gated \citep{hochreiter1997long} recurrent neural network \citep{elman1990finding}; Transformers \citep{vaswani2017attention}; and bottleneck Transformers.
For all four network types, we used a sequence-to-sequence set-up \citep{botvinick2006short,sutskever2014sequence}, such that the network was made of two subnetworks: the encoder (which converts the input sequence to a vector representation) and the decoder (which takes in the vector representation and produces the output sequence). 

Though all architectures share this high-level encoder-decoder structure, they otherwise differ substantially in how they process sequences (feedforward processing in MLPs, sequential processing in GRUs, and attention-based processing in the two types of Transformers). See Appendix~\ref{app:letter_sequence_models} for full technical details of these architectures; here we provide just the most salient points. The MLP and the GRU both encode each input sequence with a single fixed-size vector. Transformers---the architecture underlying most LLMs---are different in that their encoders produce one vector per input token, and their decoders can then access all of these encoding vectors. 
Our tasks are simple enough that each encoding vector in a Transformer likely only needs to encode a single letter (whichever letter appears at that encoding's position), raising the concern that studying these models might not be very illuminating about what strategies are used by Transformers when they need to pack multiple pieces of information into a single vector.
This consideration motivates our final architecture, the bottleneck Transformer.

While MLPs, GRUs, and Transformers are standard types of neural networks, the bottleneck Transformer is a new architecture that we introduce for the purpose of analyzing Transformer representations that we know must be combining multiple pieces of information.
The bottleneck Transformer is the same as a standard Transformer except that its decoder can only access the vector at the first input position (unlike standard Transformer decoders, which can access all input vectors).\footnote{The Transformer encoder is bidirectional, meaning that the representation at the first input position is able to be informed by information from the entire sequence, which gives this representation the opportunity to capture the whole sequence.} As a result, this vector acts as a bottleneck that must encode the entire sequence, and we can therefore study this bottleneck vector to gain a window into how Transformers combine multiple pieces of information when they need to do so.

For each combination of an architecture and a task, we train 10 different networks (varying from each other only in the random values that their parameters are initialized with); see Appendix~\ref{app:letter_sequence_target_training} for training details. All runs of all models attained near-perfect scores on their training tasks (see Figure~\ref{fig:letter_seq_target_acc} in the Appendix), showing that the encoders were successful at encoding the information needed by the decoder to produce the output. We therefore turn to DISCOVER to analyze how the encoders capture this information.

\subsection{DISCOVER results}

We apply DISCOVER to each sequence-manipulation model that we have trained. We consider the same 5 role scheme hypotheses used in Section~\ref{sec:running_example_role_schemes}: \texttt{left-to-right}, \texttt{right-to-left}, \texttt{bidirectional}, \texttt{Wickelroles}, and \texttt{bag-of-words}. Thus, overall we train 600 DISCOVER models (4 target model architectures times 3 tasks times 10 reruns per architecture/task combination times 5 role schemes). For the MLPs, GRUs, and bottleneck Transformers, the encoder produces a single vector that encodes the input; these vectors are the ones that we approximate with DISCOVER. 
The standard Transformer instead produces one encoding vector for each input letter, and the decoder can access all of these encodings. To apply DISCOVER to the Transformer, we therefore have DISCOVER approximate all encoding vectors for each input. 
We hypothesize that each vector captures one role-filler pair: a filler equal to the letter at that vector's position, paired with that letter's hypothesized role.

\begin{figure}
    \centering
    \includegraphics[width=0.8\textwidth]{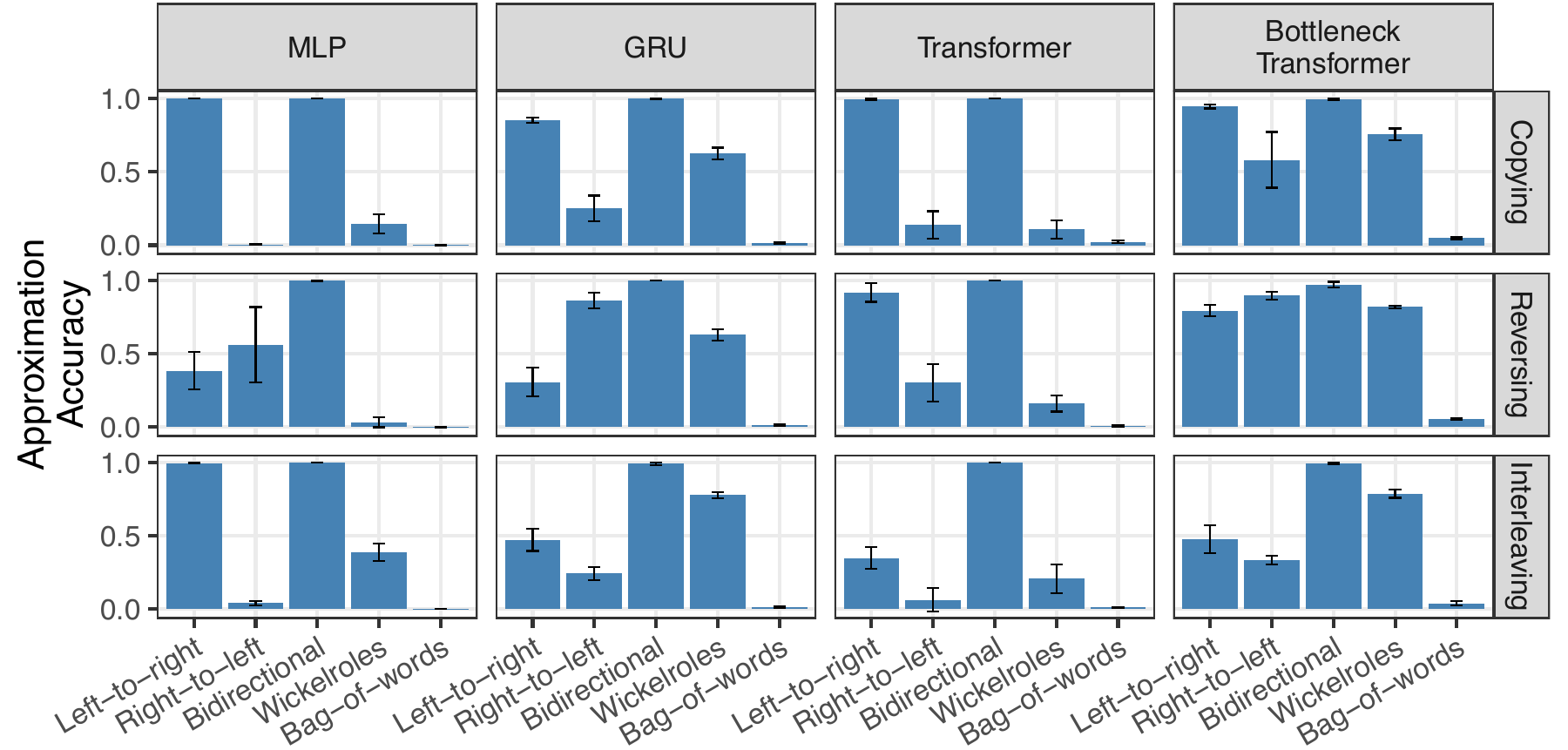}
    \caption{DISCOVER performance for models trained on letter-sequence-manipulation tasks. The bars show the mean across 10 reruns, with error bars showing two standard deviations above and below the mean.}
    \label{fig:letter_seq_tpe_acc}
\end{figure}

Across all architecture-task combinations, the \texttt{bidirectional} role scheme provides a strong DISCOVER approximation to the target models (Figure~\ref{fig:letter_seq_tpe_acc}; the ``reversing GRU'' panel repeats the results from Section~\ref{sec:running_example_results}). If we consider the \texttt{bidirectional} DISCOVER approximations for all 12 architecture-task combinations, the reversing bottleneck Transformer is the setting with the lowest average approximation accuracy (0.973); all other settings have an average accuracy greater than 0.99. It is worth noting that the \texttt{bidirectional} role scheme combines both of the unidirectional role schemes (\texttt{left-to-right} and \texttt{right-to-left}), such that \texttt{bidirectional} is guaranteed to do at least as well as each of these unidirectional role schemes. 
Though all settings are similar in that they show strong \texttt{bidirectional} approximations, they differ in the performance of the unidirectional role schemes. Broadly speaking, \texttt{left-to-right} outperforms \texttt{right-to-left} in the copying models while the opposite is true in the reversing models, which makes sense given the nature of these tasks.
Meanwhile, in most interleaving models, neither unidirectional role scheme gives an effective approximation, which makes sense given that the interleaving task combines left-to-right and right-to-left processing. There are exceptions to these trends; see Chapter~7 of \citet{mccoy2022implicit} for discussion of why we see the differences across architectures that we do.

In all cases, the \texttt{bag-of-words} role scheme gives a poor approximation. This result is expected because \texttt{bag-of-words} instantiates the null hypothesis that the target models do not encode structure (i.e., that they encode which letters are present but not how they are arranged); given that the target models show high accuracy on their tasks (which are structure-dependent), they could not be using such a structureless encoding. One might wonder whether models would use a structureless encoding if their task did not require structure. Indeed, if we train these models to sort lists into alphabetical order---a task for which the order of the input letters is irrelevant---we now see strong approximations with the \texttt{bag-of-words} role scheme (Appendix~\ref{app:sorting}).

In all cases, \texttt{Wickelroles} perform worse than \texttt{bidirectional} roles even though \texttt{Wickelroles} provide more possible roles---and thus more trainable parameters---than \texttt{bidirectional} roles (729 roles vs.\ 21 roles).
This result shows that DISCOVER success is not merely a matter of how many parameters are available to be trained. As further evidence for this point, \texttt{left-to-right} and \texttt{right-to-left} roles have equal numbers of parameters yet often show very different approximation accuracies (see Appendix~\ref{app:rolecount} for more discussion).

Overall, DISCOVER provides strong approximations across all tasks and architectures that we consider. Changing the task or architecture can change which role scheme is instantiated by the target model (i.e., which type of TPR it uses), but all of the settings are united by displaying TPR structure. These results therefore provide evidence that TPR structure is indeed adopted fairly broadly as the approach for encoding structured information in neural networks.

\section{LLM representations}\label{sec:llm_period_encodings}

We now scale up our experiments to LLMs. 
We will study how LLMs encode lists and sentences; we consider lists in order to build from the experiments in the previous section, and we consider sentences because LLMs are first and foremost models of language (the second \textit{L} of \textit{LLM} stands for \textit{language}).
A challenge for studying LLMs is that DISCOVER is designed to analyze individual vectors that encode structures (such as lists and sentences), yet Transformer LLMs do not produce single vectors as their representations; instead, LLMs use many layers of processing, and each layer produces one vector representation for each token in the input (where tokens are typically words, parts of words, or punctuation marks). For instance, if an LLM is given the input \textit{Dogs jump.}, each layer would typically produce three vectors---one encoding \textit{Dogs}, one encoding \textit{jump}, and one encoding the period. What vectors, then, should we apply DISCOVER to?

For this experiment, the hypothesis we make is that an LLM's representation for a period acts as an encoding of the entire sentence that precedes the period. There are two reasons to make this hypothesis. First, in principle, it seems plausible that LLMs would benefit from having a single location where information about entire sentences accumulates, since some scenarios faced by LLMs might require holistic information about a sentence. 
In the LLMs that we study, the representation for a token can be informed by that token itself and all previous tokens, but not by following tokens (that is, these LLMs are unidirectional, in contrast to the bidirectional Transformer encoders studied above); thus, the period would be a natural place to encode whole-sentence information since it is the earliest position at which information about the whole sentence is available.
Second, empirically, prior work has indeed found evidence that punctuation marks act as whole-sentence encodings: \citet{razzhigaev2025llm} found evidence that, given a  punctuation mark's representation from an LLM, the preceding context can be reconstructed reasonably well; further, \citet{marks2023geometry} and \citet{tigges2024language} found that punctuation marks capture sentence-level information (truth judgments and sentiment, respectively).

Therefore, as a first step toward analyzing LLM representations, the analyses in this section will apply DISCOVER to LLM encodings of the periods that appear after lists and sentences. In the following section, we will further expand the scope of the analysis to approximate LLM encodings for all tokens, not just periods.

\subsection{Conditions}

We analyze LLM period encodings in three conditions:
\begin{enumerate}[itemsep=0pt]
    \item \textbf{Lists:} We consider lists of nouns. The lists can be of length 3 to 5 inclusive---e.g., \textit{lion, trouble, kindness, pepper, rubber}. The elements of the list are drawn from the noun portion of the Toronto word pool \citep{friendly1982toronto}, following \citeauthor{armeni2022characterizing}'s (\citeyear{armeni2022characterizing}) use of this resource for studying list repetition in neural networks; our total noun vocabulary size is 300, and all words in a list have to be unique in order to facilitate the \texttt{predecessor} role scheme (see Section~\ref{sec:llm_period_discover}). We place each list after the preamble \textit{Here is a list of words:}, the list items are separated by commas, and a period is placed at the end of the list. An example of a complete model input is \textit{Here is a list of words: lion, trouble, kindness, pepper, rubber.} We then analyze the encoding of the period at the end of the list.
    \item \textbf{Subject-verb-object sentences:} We consider sentences of the form \textit{The SUBJECT VERBED the OBJECT.}---e.g., \textit{The spy helped the poet}. The subject and object were drawn from a list of 80 singular occupation terms, and the verb was drawn from a list of 16 verbs for which these occupation terms are semantically reasonable as either a subject or object. We analyze the period at the end of such sentences.
    \item \textbf{Complex sentences:} We scale up from simple subject-verb-object sentences to sentences that have varying syntactic structures---e.g., they can involve two complete clauses joined by a conjunction, or a subordinate clause, or can be in the passive voice. An example of one of these sentences is \textit{The polite runner avoided the magician after the sailor hired the photographer.} See Appendix~\ref{app:complex_cfg} for more details about the stimuli. We analyze the encodings of the periods at the ends of the sentences.
\end{enumerate}

\subsection{Models}

We analyze 7 LLMs, all accessed via Hugging Face \citep{wolf2020transformers}: Gemma-3-27b \citep{team2025gemma}, GPT-2-XL \citep{radford2019language}, GPT-OSS-20b \citep{agarwal2025gpt}, Pythia-12b \citep{biderman2023pythia}, Qwen3-14b \citep{yang2025qwen3}, OLMo-2-13B \citep{walsh2025}, and Llama-3.1-8b \citep{grattafiori2024llama}. 
The LLMs that we analyze must be open-weights (meaning that their parameters are publicly available) because we are studying internal representations.
We chose these specific LLMs by selecting several prominent open-weights model families; for each family, we then chose the most recent release available on Hugging Face at the time of running the experiments, and then among the models included in that release, we chose the largest one that could run on a single A100 GPU (the largest GPU available to us when we ran the experiments).

These models all have many layers. For each model, we consider the period encodings at 5 different layers: the first non-embedding layer, the layers that are approximately 25\%, 50\%, and 75\% of the way into the model, and the last layer. The vector representations inside these models range in size from 1600 to 5376; see Appendix~\ref{app:llm_period_info} for more details about each system.

\subsection{Do the periods actually encode the preceding sentence or list?}\label{sec:period_decoders}

\begin{figure}[t]
    \centering
    \includegraphics[width=0.98\linewidth]{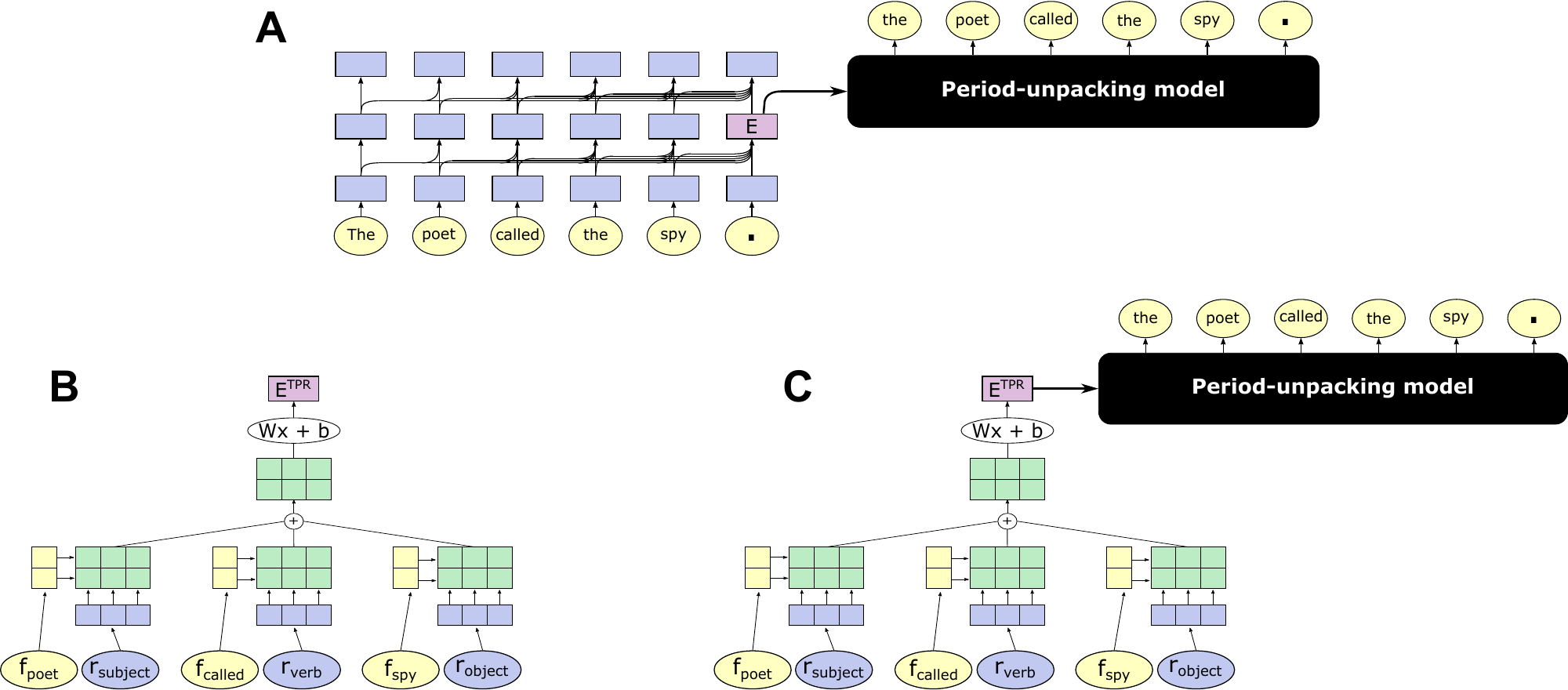}
    \caption{Analyzing LLM representations of periods. \textbf{A.} Given some input, an LLM generates one vector per input token per layer. We analyze the vector $E$ that the LLM uses to represent a period at a particular layer (here, the middle layer). We hypothesize that $E$ encodes the full sentence preceding it; we test this hypothesis by training a period-unpacking model to take in $E$ and reconstruct the sentence preceding $E$. \textbf{B.} To understand how $E$ encodes the information that the period-unpacking model extracts, we train a DISCOVER model to approximate $E$ with a TPR ($E^{TPR}$). \textbf{C.} To evaluate how successful $E^{TPR}$ has been at capturing the structure of $E$, we evaluate whether feeding $E^{TPR}$ into the period-unpacking model yields the correct output sequence. ($E^{TPR}$ and the period-unpacking model were never trained together---they are paired up solely for evaluation.)}
    \label{fig:period_encoding_methods}
\end{figure}

Our aim in this experiment is to see whether DISCOVER can reveal how LLM period representations encode the lists or sentences that precede them. 
This framing only makes sense under the assumption that the period encodings do indeed encode the preceding context, so we test this assumption before we apply DISCOVER.
For each LLM layer that we consider and for each condition, we train a decoding model that takes in a period encoding and attempts to reconstruct the sentence or list that appeared before the period (Figure~\ref{fig:period_encoding_methods}A). Each decoding model---which we refer to as a \textbf{period-unpacking model}---is a decoder-only Transformer with 6 layers, a hidden size of 1024, and 16 attention heads; it first passes the LLM period encoding through a linear layer that maps the LLM encoding (whose size varies across LLMs) into a vector of size 1024 which then acts as the input to the decoder. To prevent rogue dimensions \citep{timkey2021rogue} from having excessive influence on the decoding process, we z-score the period encodings before feeding them into the linear layer that then feeds into the period-unpacking model. We train 5 re-runs of each period-unpacking model in order to get a sense of variability across re-runs. See Appendix~\ref{app:rogue} for discussion of rogue dimensions and z-scoring, and see Appendix~\ref{app:training_period_decoders} for more details about training the period-unpacking models.

In general, the period-unpacking models do a strong job at reading out the list or sentence that preceded a period (see Figure~\ref{fig:period_decoding_acc} in the Appendix for the full results).
In the subject-verb-object setting, all period-unpacking models for all layers achieve a perfect accuracy. In the other settings, the accuracy is not perfect, but it is still reasonably high (typically above 85\% for the lists and above 50\% for the complex sentences); note that we only count the period-unpacking model as correct if it produces exactly the intended output, word-for-word, such that even a score of 50\% is reasonably strong. 
In all conditions, successful period unpacking depends on structure in order to distinguish the correct output from other potential ways of ordering the words---e.g., differentiating \textit{lion, trouble, kindness} from \textit{trouble, lion, kindness} in the list condition, or distinguishing \textit{The captain helped the baker} from \textit{The baker helped the captain} in the subject-verb-object condition.
Therefore, we conclude that the period encodings indeed encode a substantial amount of information about the structure of the preceding sentence or list, and now we use DISCOVER to investigate how they represent this information.

\subsection{DISCOVER results}\label{sec:llm_period_discover}

\begin{figure}[t]
    \centering
    \includegraphics[width=0.93\linewidth]{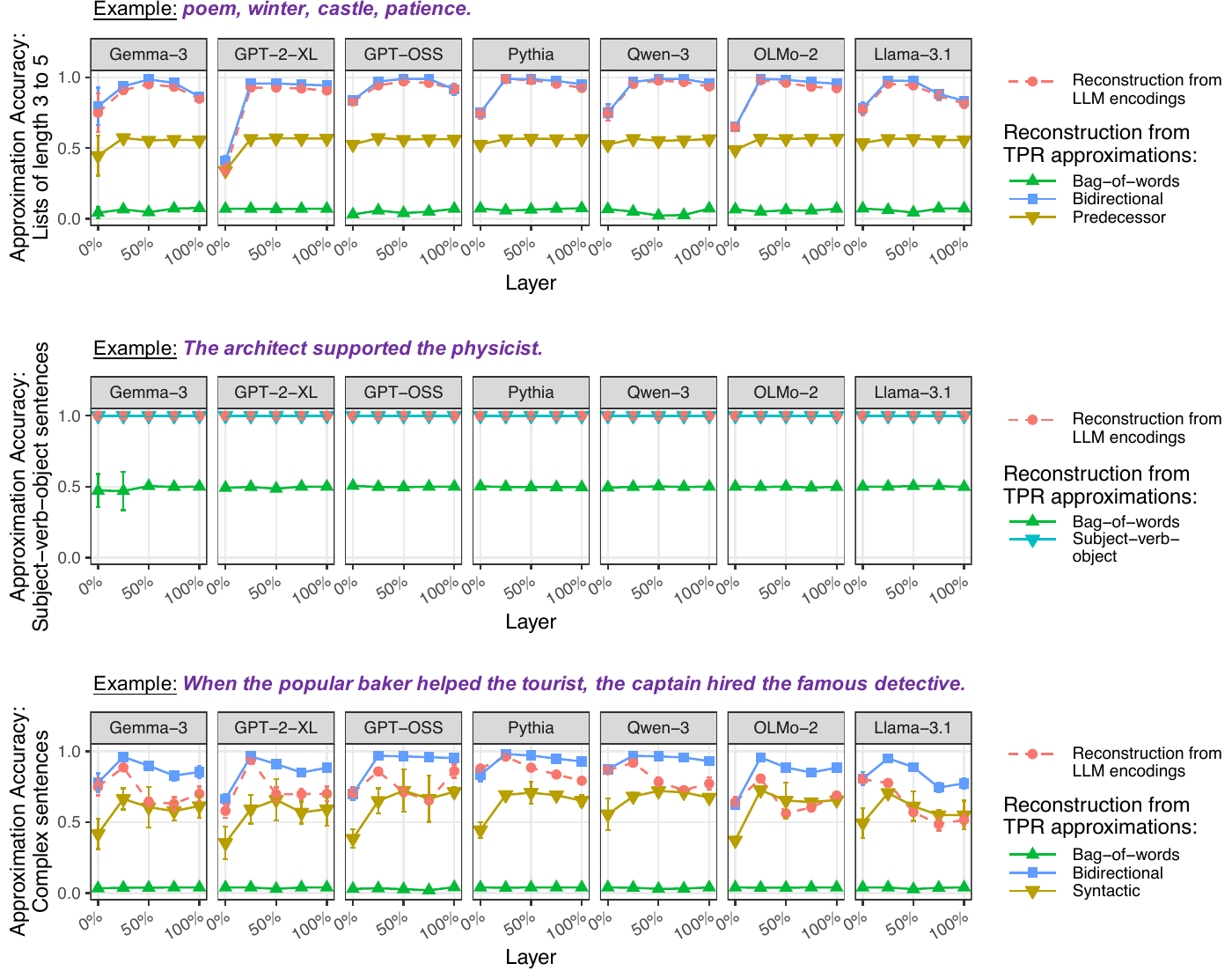}
    \caption{DISCOVER performance at approximating the encodings of periods in LLMs for lists of length 3 to 5 (top), subject-verb-object sentences (middle), and sentences with varying syntax (bottom). The points show the mean across 5 reruns, with error bars showing two standard deviations above and below the mean (the error bars are often too small to be visible). The dashed red lines show the performance of the trained period decoders when they are fed LLM period encodings; the remaining lines show the performance of the same period decoders when they are fed DISCOVER approximations. See Section~\ref{sec:llm_period_discover} for descriptions of the role schemes.}
    \label{fig:tpe_accs_period}
\end{figure}

Figure~\ref{fig:tpe_accs_period} shows the performance of DISCOVER at approximating the LLM period encodings. We measure approximation accuracy by feeding the DISCOVER approximation into the period-unpacking models that we have trained to read from the period encodings (Figure~\ref{fig:period_encoding_methods}, B and C). That is, in Section~\ref{sec:period_decoders}, the period-unpacking models were used to test what information was present in the LLM period encodings; we now see if DISCOVER captures the information that these period-unpacking models rely on. With appropriate role schemes, DISCOVER achieves high accuracy at approximating the period encodings. 

In the ``lists'' condition, the fillers are the nouns in the list. We consider three role schemes: \texttt{bidirectional}, \texttt{bag-of-words}, and \texttt{predecessor}. The \texttt{bidirectional} and \texttt{bag-of-words} schemes are the same as in Section~\ref{sec:letter_sequence_models}. In the \texttt{predecessor} role scheme, each filler's role is equal to the word before it; e.g., if the list is \textit{lion, trouble, kindness}, the set of \textit{role}-\texttt{filler} pairs would be \{\textit{\#}:\texttt{lion}, \textit{lion}:\texttt{trouble}, \textit{trouble}:\texttt{kindness}, \textit{kindness}:\texttt{\#}\} (using \# to indicate list boundaries). Our lists never contain repeated elements, which ensures that the set of \texttt{predecessor} role-filler pairs is always sufficient to reconstruct the list. We find that DISCOVER with a \texttt{bidirectional} role scheme approximates the encodings well, performing similarly to using the actual LLM encodings (labeled \texttt{target model} in the figure). The \texttt{predecessor} role scheme performs substantially worse than \texttt{bidirectional} despite having a much larger role vocabulary (301 vs.\ 12 roles). 

In the subject-verb-object case, DISCOVER achieves perfect approximation accuracy with the \texttt{subject-\allowbreak verb-\allowbreak object} role scheme, in which each sentence has three roles: \textit{subject}, \textit{verb}, and \textit{object}. E.g., the sentence \textit{The photographer called the teacher} would be analyzed with the following set of role-filler pairs: \{\textit{subject}:\texttt{photographer}, \textit{verb}:\texttt{called}, \textit{object}:\texttt{teacher}\}. In the sentence dataset that we use, any noun can be the subject or object, so having distinct roles for \textit{subject} and \textit{object} is important for distinguishing pairs such as \textit{The photographer called the teacher} and \textit{The teacher called the photographer}.\footnote{Although we use \texttt{bidirectional} roles in most of our experiments, we do not include them here because the fixed structure of the sentences ensures that \texttt{bidirectional} roles would capture effectively the same information as the \texttt{subject-\allowbreak verb-\allowbreak object} role scheme that is already included. E.g., the sentence's subject is always the same as its second-from-left-and-fourth-from-right word. It is true that the \texttt{subject-\allowbreak verb-\allowbreak object} roles omit two words that would be included with \texttt{bidirectional} roles (the two instances of \textit{the}), but this difference is not meaningful because those words are invariant across stimuli, meaning that their inclusion in the role scheme would be no different from modifying the fixed bias term that is part of the DISCOVER model.
The ``complex sentences'' condition disentangles linear position from syntactic position by using variable sentence structures.}

For complex sentences, we consider \texttt{bidirectional} roles, \texttt{bag-of-words} roles, and \texttt{syntactic} roles; in the \texttt{syntactic} role scheme, each word's role is its syntactic position, which we define as the path of dependency arc labels needed to get from the word to the root of the sentence's dependency parse (see Appendix~\ref{app:period_discover} for an example). 
The \texttt{bidirectional} role scheme consistently provides a strong approximation. It substantially outperforms the \texttt{syntactic} roles, suggesting that the period encodings mainly capture the sentence's linear order rather than its syntactic structure (a conclusion further corroborated by follow-up analyses in Appendix~\ref{app:period_syntax}). An additional striking property of the results is that the period-unpacking models achieve a higher accuracy when they are given \texttt{bidirectional} DISCOVER approximations than when they are given the actual period encodings from the target LLMs, often by a large margin---a fact that we return to in the Discussion because it has potential implications for the relationship between vectors and symbols.
To briefly preview what we will argue, we take this result as evidence that the LLM period encodings have a \texttt{bidirectional} TPR structure that is realized imperfectly. 
This structure is central enough that it is what the period-unpacking models learn to rely on, but it is noisy enough that the period-unpacking models often then make mistakes. The DISCOVER approximation realizes this same structure in a precise, noise-free way, enabling more robust behavior in the period-unpacking models. See Section~\ref{sec:discussionsymbolic} for elaboration on this argument.

In sum, across the 7 LLMs that we have studied, the representations of periods capture most of the structure of the sentence or list preceding the period, and DISCOVER can closely approximate the encodings of these periods. This result is evidence that the representational spaces of these LLMs have TPR structure.

\section{Arithmetic, logic, coding, and language in an LLM}\label{sec:gptoss}

We now select one specific LLM---GPT-OSS---and analyze the representations that it uses for 4 domains that are especially prominent in symbolic traditions: math, logic, computer code, and language.
For each domain, we analyze the  representations produced by this LLM when it is prompted to perform tasks in this domain, and we see if we can reproduce the LLM's behavior using DISCOVER approximations to these representations.
We select these four domains in order to get at our core puzzle of how LLMs---despite being vector-based---can represent the information needed to perform seemingly symbolic computations.

An additional motivation for this experiment is to address two shortcomings of Section~\ref{sec:llm_period_encodings}. First, while the experiments in that section found TPR structure in LLM encodings, those experiments do not tell us whether the LLMs use that structure because the system whose behavior was reproduced was a newly-trained period-unpacking model rather than the LLM itself. Second, those experiments analyzed only period encodings.
The experiments in this section address these limitations by creating DISCOVER approximations for all LLM representations in the input (across all layers and tokens) and then checking whether the LLM produces the correct answer when given the DISCOVER approximation rather than its own representations.

\subsection{Tasks}

Table~\ref{tab:llm_task_examples} gives examples for each task that we prompt the LLM to perform. Below we describe each task in more detail, with further details in Appendix~\ref{app:symbolic_technical}. For all tasks, the LLM's input contained at least one example of the task with the correct answer since LLMs generally benefit from such examples via the mechanism of in-context learning \citep{brown2020language}. E.g., for arithmetic, the prompt includes two examples: \texttt{-9 + 8 * 7} (with the answer of \texttt{47}) and \texttt{5 * 7 + -2} (with the answer of \texttt{33}). The LLM prompts were in GPT-OSS's standard prompt format,\footnote{https://github.com/openai/harmony} which uses special tokens to demarcate input components (e.g., separating user queries from LLM responses). All of the words in our stimuli were processed as a single token by GPT-OSS (there are some words that GPT-OSS internally splits up into multiple pieces, but we avoided such words).

\renewcommand{\arraystretch}{1.7}
\begin{table}[]
    \centering
    \begin{tabular}{cp{6.5cm}p{6cm}} \toprule
        Task & Example & Answer \\ \midrule
        Arithmetic & 7 + -6 * 3 & -11 \\
        Syllogisms & We are given two premises:\newline No polite agents are painters.\newline All gamblers are painters.\newline \newline Please return the first conclusion listed below that follows from these premises:\newline All gamblers are polite agents.\newline Some gamblers are polite agents.\newline No gamblers are polite agents.\newline Some gamblers are not polite agents. & No gamblers are polite agents. \\ 
        Code execution & def change(sequence):\newline \newline \hspace*{1em} return [``Q''] + sequence \newline \newline def alter(sequence):\newline \newline \hspace*{1em} return sequence[::-1] \newline \newline y = [``S'', ``M'', ``V''] \newline x = [``U'', ``C''] \newline \newline change(y) & [``Q'', ``S'', ``M'', ``V''] \\
        Passivization & Passivize the following sentence: The swimmer who the polite politician advised avoided the accountant near the clever gardener. & The accountant near the clever gardener was avoided by the swimmer who the polite politician advised. \\
        Tense reinflection & Convert the following sentence to the present tense: The important detective who recommended the astronauts recognized the tall detective. & The important detective who recommends the astronauts recognizes the tall detective.\\ 
        Question formation & Convert the following sentence to a yes/no question: The professor who the painter will encourage would visit the musician near the famous runner. & Would the professor who the painter will encourage visit the musician near the famous runner? \\
        \bottomrule
    \end{tabular}
    \caption{Symbolic tasks that we use with GPT-OSS, along with examples and their answers. These tasks span 4 domains: mathematics (arithmetic), logic (syllogisms), computer code (code execution), and language (passivization, tense reinflection, and question formation).}
    \label{tab:llm_task_examples}
\end{table}
\renewcommand{\arraystretch}{1.0}

\paragraph{Arithmetic:} Each arithmetic problem takes the form of either \texttt{a * b + c} or \texttt{a + b * c}, where \texttt{a}, \texttt{b}, and \texttt{c} are integers from -9 to 9 inclusive. Processing these inputs requires correctly applying the order of operations (i.e., performing the multiplication first and then the addition).

\paragraph{Syllogisms:} Syllogisms are a classic form of logical deduction. Two premises are provided, followed by 4 possible conclusions. The LLM must return the first of these conclusions that follows from the premises. 

\paragraph{Code execution:} This task requires predicting the output of a Python function call. All functions in question are manipulations of Python lists. The query first provides the definitions of two functions named \texttt{change} and \texttt{alter}, whose definitions vary from query to query; some example functions are ``reverse the list'' or ``insert a Z at the start of the list.'' Two variables named \texttt{x} and \texttt{y} are then defined; each is a list of letters of length 1, 2, or 3. Finally, a function call is given pairing one of the two functions with one of the two lists as its argument---e.g., \texttt{alter(x)}. The intended output is the result of applying this function call. The two functions can appear in either order, and similarly for the two list variables \texttt{x} and \texttt{y}; thus, one major component of this task is variable binding (associating each variable name with its value).

\paragraph{Passivization, tense reinflection, and question formation:} These tasks are three ways of manipulating the syntax of an English sentence. In passivization, the LLM must convert an active English sentence into its passive form. 
In tense reinflection, the LLM must convert a past-tense English sentence into the present tense, which requires the LLM to identify each verb's subject in order to know whether the verb should take its singular or plural form (since English past tense verbs are not inflected for number while English present tense verbs are).
In question formation, the LLM must convert a declarative English sentence into a yes/no question.
For all three tasks, the input sentences were generated from a context-free grammar in which sentences always took the basic form of subject-verb-object, but with substantial variability in the syntactic structure of the subject and object: they could be just \textit{the} plus a noun (e.g., \textit{the doctor}), or they could optionally include adjectives, prepositional phrases, and relative clauses (e.g., \textit{the illustrator who surprised the gardener next to the tall editor}).

\subsection{Model}

We analyze GPT-OSS \citep{agarwal2025gpt}. We use only one LLM because these experiments are computationally intensive; analyzing even just this one LLM required 
a little over 3,000 GPU hours on h100 and h200 GPUs on a high-performance computing cluster.
GPT-OSS performs well on all 6 tasks that we consider here. Its worst task is syllogisms (with an accuracy of 0.76), followed by passivization (with an accuracy of 0.94); for all other tasks, the accuracy is above 0.96. See Figure~\ref{fig:llm_target_acc} in the Appendix for its performance on all tasks.

\begin{figure}[t]
    \centering
    \includegraphics[width=\linewidth]{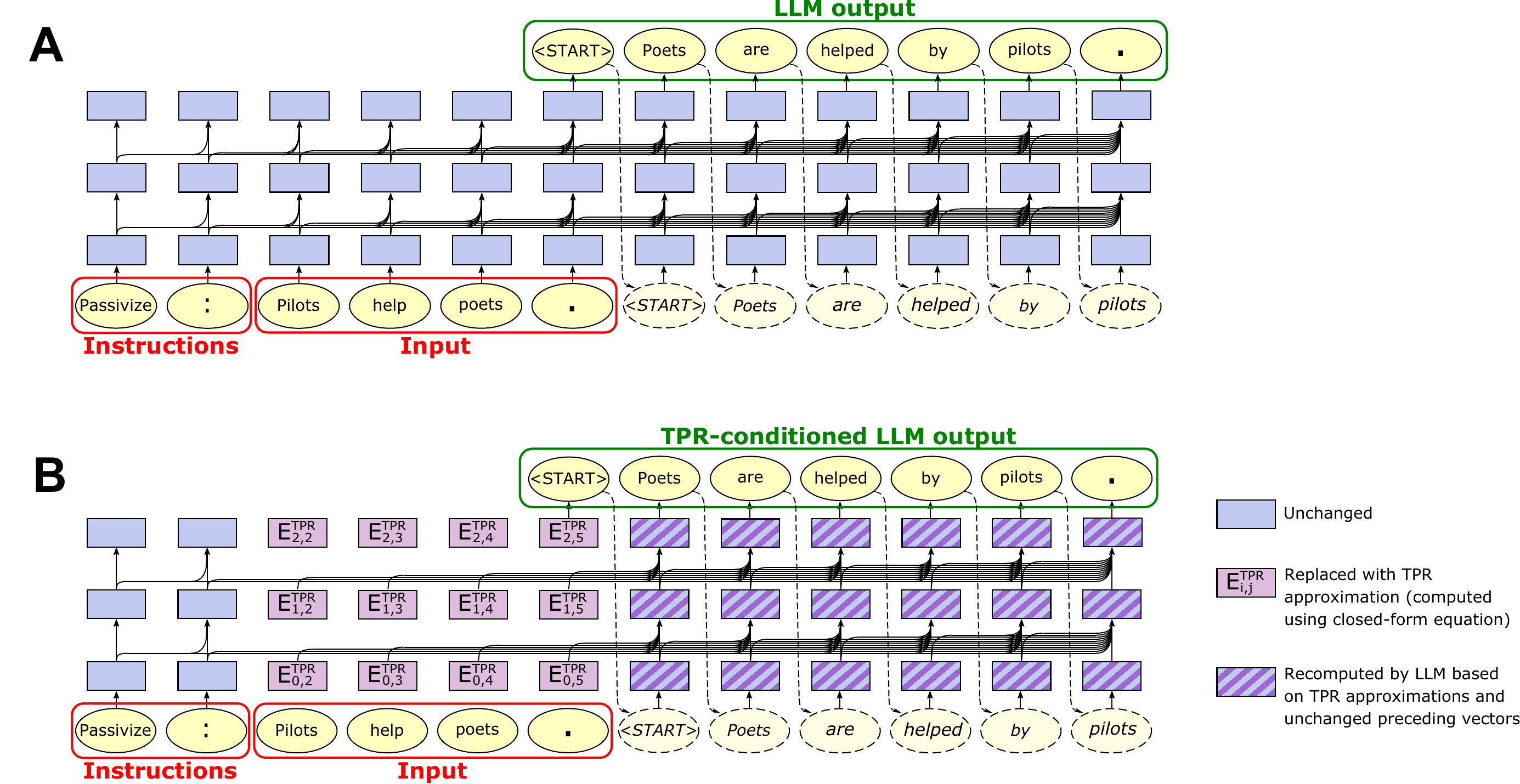}
    \caption{Analyzing the complete input representation of an LLM (GPT-OSS). \textbf{A:} Standard LLM processing. The LLM is given a prompt made of some instructions and an input. The LLM uses many layers of processing (we show 3 layers here, but GPT-OSS in fact has 25 layers); it generates one vector per layer for each token in the prompt. It then generates the output, one token at a time. As each output token is generated, that token is appended to the prompt, and the LLM generates vector representations for it (one vector per layer), until the output is finished. 
    \textbf{B:} Approximating the LLM with DISCOVER. We leave the representations of the instructions unchanged because they are identical across all inputs for a given task, but we replace all vectors that encoded the input---across layers and input tokens---with a TPR approximation (which is produced by an interpretable, closed-form equation). The LLM then generates its output based on the TPR approximations rather than based on its own input representations. We assess whether the resulting LLM output is correct. The instructions and input shown here are simplified for the sake of space.}
    \label{fig:gptossmethods}
\end{figure}

\subsection{DISCOVER results}

We now apply DISCOVER to GPT-OSS's encodings. For each task, we train one DISCOVER model for each of GPT-OSS's 25 layers. The DISCOVER model for a given layer is trained to produce an approximation for every token representation at that layer except for templatic text that is uniform across stimuli within a task; see Figure~\ref{fig:gptossmethods} for a depiction of the approach. The role schemes that we investigate are motivated by two questions:
\begin{enumerate}[itemsep=0pt]
    \item Does GPT-OSS's representation for each token encode just that token (i.e., just one role-filler pair), or does it combine multiple pieces of information (multiple role-filler pairs)?
    \item When performing these tasks, does GPT-OSS use task-specific structure to represent information, or a more general structure?
\end{enumerate}
Motivated by these questions, we consider 5 types of role schemes, with examples in Table~\ref{tab:gpt_oss_roles}:
\begin{itemize}[itemsep=0pt]
    \item \textbf{Bag-of-words:} Each token is assumed to encode itself and all preceding tokens, but with all of them given the same role as each other, making it effectively a bag of words.
    \item \textbf{Bidirectional (self):} Each token is assumed to encode only itself (as a filler), paired with a role denoting this token's bidirectional position in the sequence so far. 
    \item \textbf{Bidirectional (all):} Each token is assumed to encode itself and all preceding tokens; each of these tokens gets a role equivalent to the token's bidirectional position in the sequence so far.
    \item \textbf{Task-specific (self):} For each task, we define a task-specific parsing algorithm that assigns each token a role that denotes its position in the context of the task. E.g., for coding, one such position is \texttt{second letter in the list that is the value of the variable x}; for language tasks, one such position is \texttt{adjective modifying the direct object}. Under this role scheme, each token encodes only itself (as a filler), with a role that is its position as given by this task-specific parsing algorithm.
    \item \textbf{Task-specific (all):} For this scheme, we now assume that each token $t_j$ encodes itself and all preceding tokens---that is, each $t_i$ for which $0 \leq i \leq j$. Inside the representation of $t_j$, each of these $t_i$ tokens is taken to be one filler, with a role that is the concatenation of the task-specific position for $t_i$ as well as the task-specific position for $t_j$.
    E.g., one such role might be \texttt{subject\_noun-object\_adj}, which is the role attached to the representation of the subject noun inside the representation of the adjective modifying the object. 
    Such pair-based roles are motivated by the fact that they can capture the sorts of interactions that we hypothesize are important for how one token might influence the representation of another. Consider the sentence \textit{The tall poet helped the spy.} The token \textit{tall} is likely to have a different influence on the representation of \textit{poet} (which it directly describes) than on the representation of \textit{spy} (which it has no close relationship to). These differing relationships can be captured by having different roles for \textit{subject\_adj-subject\_noun} and \textit{subject\_adj-object\_noun}.\footnote{An additional motivation for the pair-based roles is based on the architecture of GPT-OSS. Like most LLMs, GPT-OSS uses attention, which enables the representation of one token $t_j$ to be informed by the representation of another token $t_i$. The way in which $t_j$ should be influenced by $t_i$ is determined by a representation generated from $t_j$ (called the query) and a representation generated from $t_i$ (called the key). At a high-level, our pair-based roles---which combine a basic role from one token with a basic role from another---act somewhat like the keys and queries in that they enable the interactions between two tokens to be informed by abstract information about both tokens. Thus, we hypothesize that these pair-based roles capture some types of outcomes that could arise from building representations in a way that leverages attention based on queries and keys.}
\end{itemize}

\begin{table}[h!t]
    \centering
    \small
    \begin{tabular}{cp{2.5cm}p{2.0cm}p{2.0cm}p{2.5cm}p{3.8cm}} \toprule
        & Bag-of-words & \multicolumn{1}{c}{Bidirectional} & \multicolumn{1}{c}{Bidirectional}  & \multicolumn{1}{c}{Task-specific}  & \multicolumn{1}{c}{Task-specific}  \\ 
        & & \multicolumn{1}{c}{(self)} & \multicolumn{1}{c}{(all)}  & \multicolumn{1}{c}{(self)}  & \multicolumn{1}{c}{(all)}  \\ \midrule
        The & \{present:The\} & \{(0,0):The\} & \{(0,0):The\} & \{subj\_det:The\} & \{subj\_det-subj\_det:The\} \\ \midrule
        spy & \{present:The, present:spy\} & \{(1,0):spy\} & \{(0,1):The, (1,0):spy\} & \{subj\_n:spy\} & \{subj\_det-subj\_n:The, subj\_n-subj\_n:spy\} \\ \midrule
        helped & \{present:The, present:spy, present:helped\} & \{(2,0):helped\} & \{(0,2):The, (1,1):spy, (2,0):helped\} & \{verb:helped\} & \{subj\_det-verb:The, subj\_n-verb:spy, \newline verb-verb:helped\} \\ \midrule
        the & \{present:The, present:spy, present:helped, present:the\} & \{(3,0):the\} & \{(0,3):The, (1,2):spy, (2,1):helped, (3,0):the\} & \{obj\_det:the\} & \{subj\_det-obj\_det:The, subj\_n-obj\_det:spy, verb-obj\_det:helped, obj\_det-obj\_det:the\} \\ \midrule
        poet & \{present:The, present:spy, present:helped, present:the, present:poet\} & \{(4,0):poet\}  & \{(0,4):The, (1,3):spy, (2,2):helped, (3,1):the, (4,0):poet\} & \{obj\_n:poet\}  & \{subj\_det-obj\_n:The, subj\_n-obj\_n:spy, \newline verb-obj\_n:helped, obj\_det-obj\_n:the, obj\_n-obj\_n:poet\}\\
        \bottomrule
    \end{tabular}
    \normalsize
    \caption{The five role schemes used to analyze GPT-OSS performing passivization, illustrated through the input \textit{The spy helped the poet} (an actual stimulus would also contain formatting tokens, such as a period). Each row corresponds to one input token. Each table entry shows the role-filler pairs that the relevant token's representation is hypothesized to contain under the indicated role scheme. For instance, under the \texttt{bidirectional (all)} role scheme, the token \textit{helped} is hypothesized to contain three role-filler pairs: \{(0,2):The, (1,1):spy, (2,0):helped\}.}
    \label{tab:gpt_oss_roles}
\end{table}

These role schemes are elaborated on in Appendix~\ref{app:gpt_oss_role_schemes}. The two \texttt{self} role schemes allow us to test whether GPT-OSS can be approximated with just one role-filler pair in each token's hidden state (question 1 above); if the other role schemes meaningfully outperform the \texttt{self} ones, this is evidence that each token's representation combines multiple pieces of information. 
Comparing the task-agnostic vs.\ task-specific roles then allows us to test if task-specific structure is necessary to give a strong approximation (question 2 above); \texttt{bag-of-words} and \texttt{bidirectional (all)} give two versions of task-agnostic structure (degenerate and non-degenerate), while \texttt{task-specific (all)} gives task-specific structure.
For each role scheme and task, we perform 5 runs of DISCOVER training to get a sense of variability across re-runs.
See Appendix~\ref{app:gpt_oss_discover} for DISCOVER training details.

\begin{figure}[h!t]
    \centering
    \includegraphics[width=1.0\textwidth]{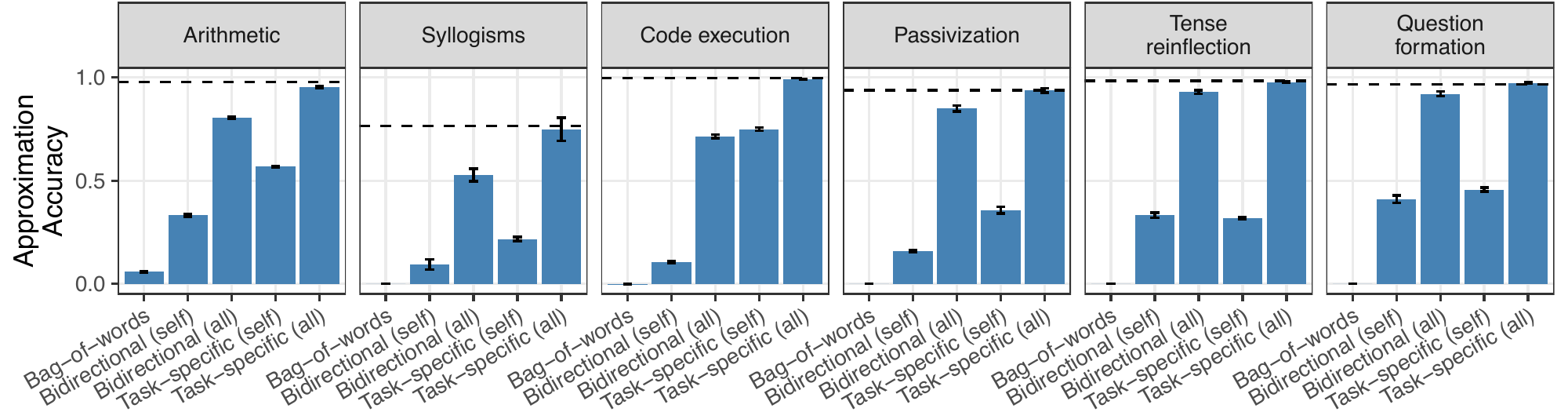}
    \caption{DISCOVER results at approximating GPT-OSS performing various symbolic tasks. The dashed lines indicate GPT-OSS's performance on each task. Each bar is the mean across 5 reruns, with error bars showing two standard deviations.}
    \label{fig:llm_tpe_acc}
\end{figure}

The results are in Figure~\ref{fig:llm_tpe_acc}. For all six tasks, \texttt{task-specific (all)} outperforms the other role schemes,\footnote{Note that, in five of the six tasks, \texttt{task-specific (all)} also uses a larger role vocabulary than any other role scheme. However, there is evidence that this factor is not what drives the success of \texttt{task-specific (all)}; see Appendix~\ref{app:rolecount}.} giving an approximation accuracy that is extremely close to GPT-OSS's own accuracy at the task. (The largest gap between GPT-OSS's accuracy and the average accuracy of the DISCOVER approximation is 2.36\%, in the arithmetic task). 
These results provide evidence that GPT-OSS encodes multiple role-filler pairs at each position and that these role-filler pairs have semantics corresponding to elements of task-specific structure. 
Zooming out, the findings support the view that the way GPT-OSS performs these symbolic tasks is by constructing a TPR for each input token that encodes task-specific information about itself and the preceding context.

\section{Causal interventions}\label{sec:causal_interventions}

An important property of symbolic structures is that they are made of multiple components.
We have shown so far that we can replicate the behavior of various neural networks by replacing their \textit{entire} representations with TPRs. We now investigate whether these same networks are sensitive to the compositional structure within the TPR: 
if we edit just \textit{part} of a network's representation, does the network's behavior change in the expected ways?

The TPR formalism makes it straightforward to perform such partial edits; the procedure is shown in Figure~\ref{fig:letter_seq_interventions} (left). 
In a TPR, the representation of a complete structure is the sum of the representations of the role-filler pairs within it. 
For instance, the representation of the list \texttt{Q,M,Z} using \texttt{left-to-right} roles would be the sum of the representations of the three role-filler pairs (where $TPR(a:b)$ is realized as $W(r_a \otimes f_b)$):\footnote{In our framing earlier in this paper, a linearly-transformed TPR is defined as $W(\sum_i f_i \otimes r_i) + b$. In this section we assume an equivalent expression in which $W$ has been distributed inside the sum: $(\sum_i W(f_i \otimes r_i)) + b$. We omit the bias term $b$ in the discussion here because it cancels across the terms in Equation~\ref{eq:edit}.}
\begin{align}
    TPR(\texttt{Q,M,Z}) &= TPR(1st:\texttt{Q}) + TPR(2nd:\texttt{M}) + TPR(3rd:\texttt{Z})
\end{align}
That is, ``the whole is literally the sum of the parts'' \citep{smolensky2006formalizing}. As a result, we can change one TPR into another by subtracting out the representation of one role-filler pair and adding in the representation of another. For instance,  we could change $TPR(\texttt{Q,M,Z})$ into $TPR(\texttt{Q,M,U})$ as follows:
\begin{align}
    TPR(\texttt{Q,M,Z}) - TPR(3rd:\texttt{Z}) + TPR(3rd:\texttt{U}) = TPR(\texttt{Q,M,U}) \label{eq:edit}
\end{align}
If our target model's representations are indeed implicit TPRs, then we should be able to modify those representations as well using the same edits, where $target(s)$ is a representation obtained from our target model and $TPR(r:f)$ is a representation obtained from our DISCOVER model:
\begin{align}
    target(\texttt{Q,M,Z}) - TPR(3rd:\texttt{Z}) + TPR(3rd:\texttt{U}) = target(\texttt{Q,M,U}) \label{eq:surgery}
\end{align}
To test whether our target models are indeed sensitive to the compositional structure of TPRs, we run tests based on equations like \ref{eq:surgery}: we start with a representation from the target model, then subtract out the representation predicted by DISCOVER for one or several role-filler pair(s),  then add in the representation predicted by DISCOVER for one or several new role-filler pair(s).
Note that the revision we make 
changes \textit{every} number in the representational vector by a precise amount; the method does not assume that individual pieces of information are localized inside individual numbers within the vector. 
We then plug the edited representation into the target model to see if its behavior appropriately reflects the targeted edit that we have made.

TPR-based representational editing was introduced by \citet{soulos2020discovering} under the name of \textit{constituent surgery}. It is a type of causal intervention---an analysis that involves changing a model's representations to see if its behavior changes accordingly, as a test of whether structure that has been identified in the representations is also causally implicated in model behavior \citep{giulianelli2018hood,bau2019visualizing,Besserve2020Counterfactuals,geiger2025causal}. 
It is worth noting some differences between constituent surgery and two other prominent methods for intervening on neural network representations, namely patching \citep{vig2020investigating,meng2022locating,wang2023interpretability,zhang2024towards} and steering \citep{subramani2022extracting,li2023inference,turner2023steering,rimsky2024steering}.
Patching and steering both involve editing a network's representation of one input based on its representation(s) of some other input(s); in patching, entire parts of the other representation are swapped into the current representation, while in steering a targeted vector (created by, e.g., averaging over other representations that show some desired property) is added to the representations in order to encourage a particular behavior. In contrast, the edits made in constituent surgery are derived from an interpretable, closed-form analysis of the network's representational space rather than from the network's representations of particular examples, Constituent surgery shares this property with distributed alignment search \citep[DAS;][]{geiger2024finding}, but DAS differs from constituent surgery in that it operates via directly training an analysis model to facilitate causal interventions, whereas in DISCOVER the training is based on a representational hypothesis that enables causal interventions as a downstream consequence.

\subsection{Letter sequence models}

\begin{figure}[t]
    \centering
    \includegraphics[width=0.38\linewidth]{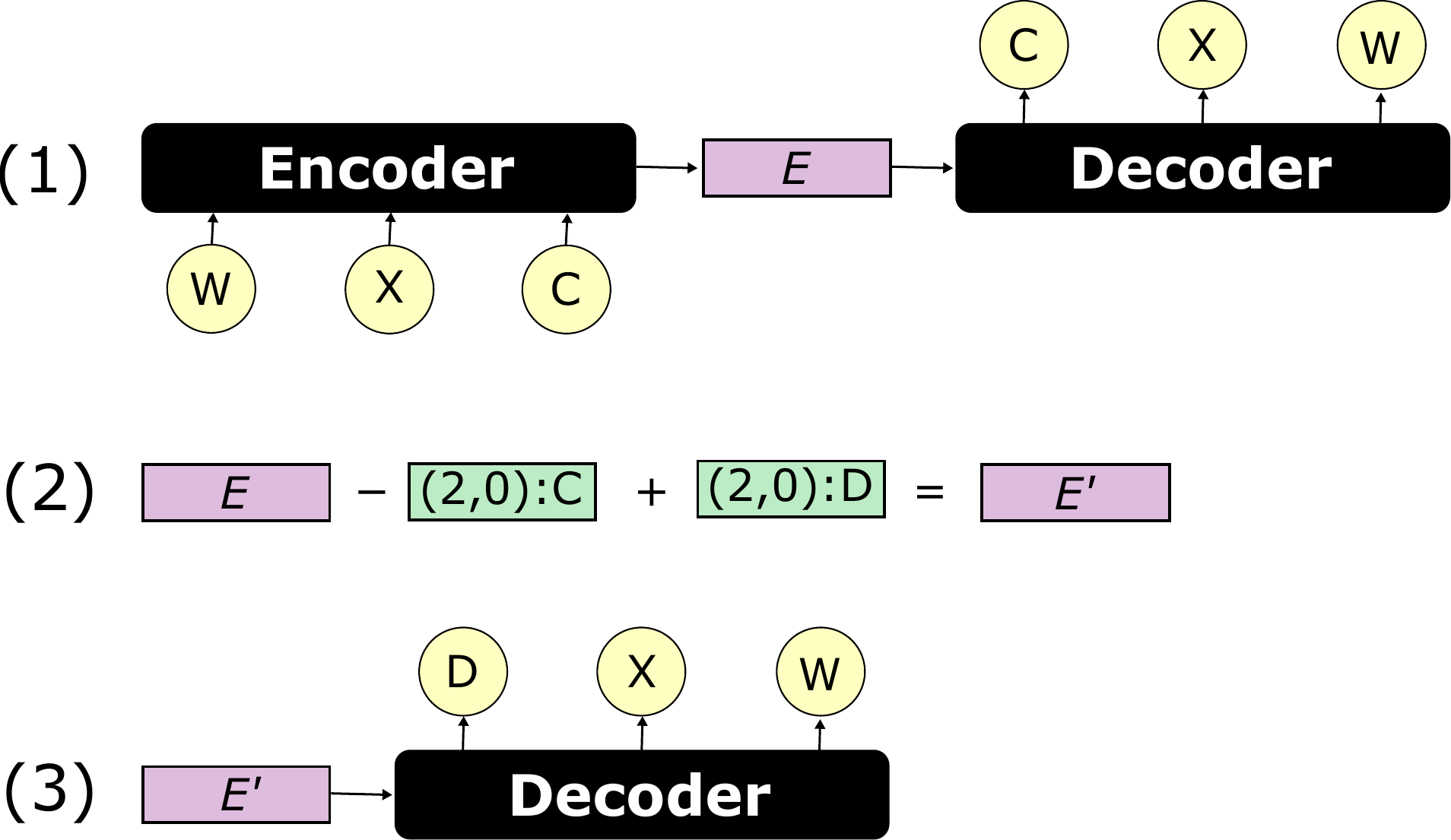}
    \hspace{1cm}
    \includegraphics[width=0.36\linewidth]{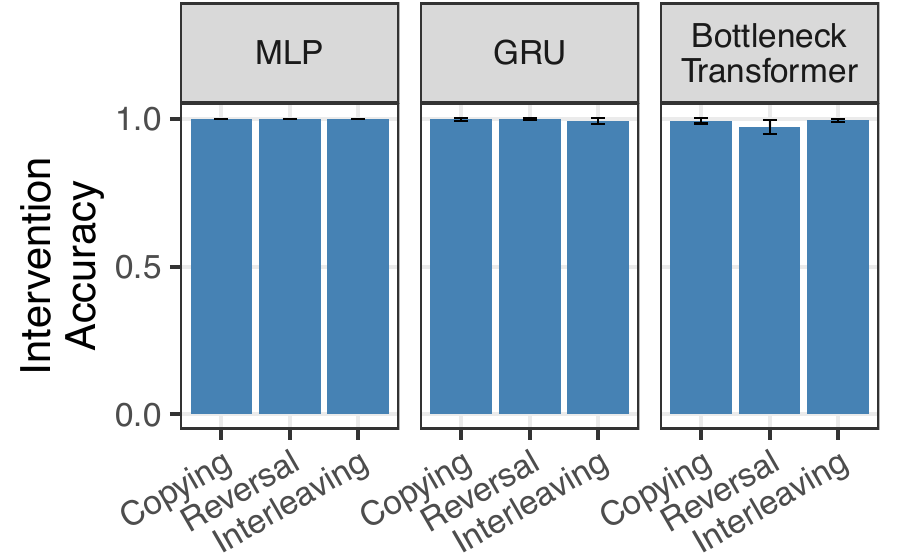}
    \caption{\textbf{Left:} The constituent surgery process used to test if we can edit a model's behavior via targeted edits to its representations \citep{soulos2020discovering}. \textit{(1)} We obtain the target model's encoding $E$ for an input (here, the input \texttt{W,X,C} is being provided to a target model that performs reversal). \textit{(2)} We edit $E$ by subtracting the predicted representation for one role-filler pair and adding the predicted representation for another role-filler pair, producing an edited encoding $E'$. \textit{(3)} We feed $E'$ into the target model's decoder and see if the decoder shows the intended change in behavior (here, producing \texttt{D,X,W} instead of \texttt{C,X,W}). 
    \textbf{Right:} Results of applying constituent surgery on models trained to perform transformations on sequences of letters.}
    \label{fig:letter_seq_interventions}
\end{figure}

We first perform constituent surgery on the models from Section~\ref{sec:letter_sequence_models} that were trained to manipulate sequences of letters. 
The role-filler representations that we used to make the edits were taken from DISCOVER models with \texttt{bidirectional} roles.
We only analyzed the architectures that use a single-vector encoding (MLPs, GRUs, and bottleneck Transformers, but not standard Transformers) because the multi-vector models in Section~\ref{sec:letter_sequence_models} were well-approximated with only one role-filler pair per vector; thus, interventions on those models would be overwriting whole vectors, meaning they would not achieve our goal of editing only part of a representation.
Across architectures and tasks, these causal interventions achieve near-perfect accuracy (Figure~\ref{fig:letter_seq_interventions}, right).

\subsection{Period encodings}

\begin{figure}[t]
    \centering
    \includegraphics[width=0.9\linewidth]{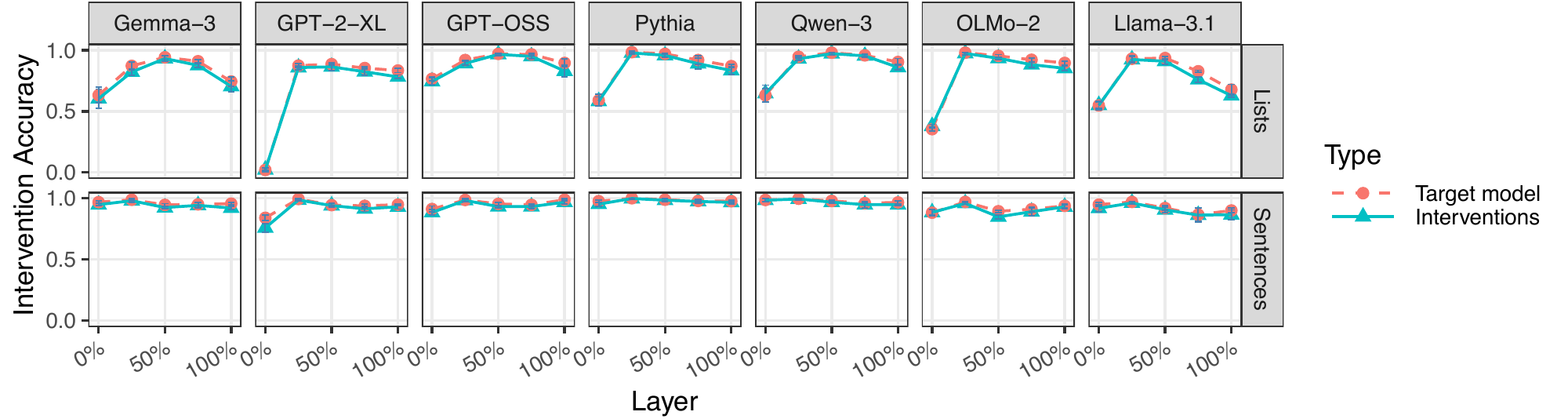}
    \caption{Accuracy of causal interventions carried out on LLM period encodings of lists and sentences.}
    \label{fig:period_interventions}
\end{figure}

We now perform causal interventions on LLM encodings of periods following both lists and sentences, building on the experiments in Section~\ref{sec:llm_period_encodings}.
For lists, we do interventions that replace one word in a list of length 5.
For sentences, we use sentences of the form \textit{The NOUN VERBED the NOUN, and the NOUN VERBED the NOUN}, and we run interventions that replace one noun with another. 
See Appendix~\ref{app:period_encoding_interventions} for more details.
The interventions achieve high accuracy, coming close in all cases to the accuracy of the target model (Figure~\ref{fig:period_interventions}).

\subsection{Symbolic tasks in GPT-OSS}

Finally, we also perform causal interventions on GPT-OSS when it is performing tasks in arithmetic, logic, coding, and language. For instance, 
given the input \textit{The spy helped the clever poet}, can we edit the sentence structure by editing the representation of \textit{clever} to change its position from \textit{object adjective} to \textit{subject adjective}, such that GPT-OSS responds as if the input had been \textit{The clever spy helped the poet}?

Implementing these interventions is more complicated than the interventions described in the previous subsections because the GPT-OSS representations that we approximated via DISCOVER do not use a single vector per input sequence but rather include one vector per input token per layer. Further, under our DISCOVER hypotheses, each token encodes information about itself and all previous tokens.
To conduct an intervention on GPT-OSS, we therefore edit all representations (across tokens and layers) that are predicted under our DISCOVER analysis to require revision in order to carry out the intervention. See Appendix~\ref{app:gpt_oss_causal_interventions_details} for details.

Across the 6 tasks described in Section~\ref{sec:gptoss}, we run 31 types of causal interventions, which yield an average intervention accuracy of \textbf{0.903}. 
This result shows that GPT-OSS is indeed sensitive to the constituent structure of the TPRs that we have used to approximate it, such that we can make edits to one part of this structure and have GPT-OSS's behavior change appropriately. For full details of these 31 evaluations, see Appendix~\ref{app:gpt_oss_causal_interventions_details}. In the rest of this subsection, we describe a selection of the results from these experiments. 

\begin{figure}
    \centering
    \includegraphics[width=0.7\linewidth]{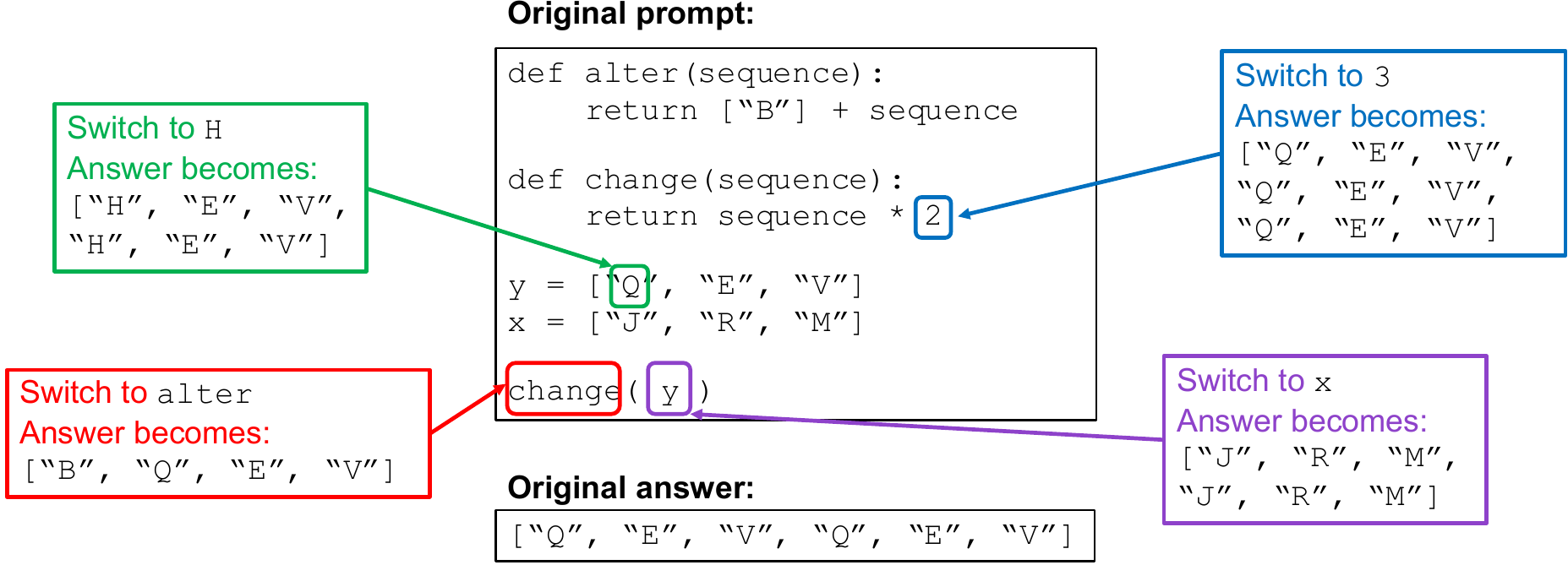}
    \caption{Causal interventions on fillers in GPT-OSS. For each type of intervention shown here, we run 100 interventions on examples of that type, yielding accuracies of 0.99 when changing an element in a list, 1.00 when changing the number of repetitions, 0.95 when changing which function is called, and 0.96 when changing which variable is used as the argument of the function call.}
    \label{fig:coding_causal_interventions}
\end{figure}

\paragraph{Changing fillers:} We first consider interventions in which we change the value of a filler (the same type of intervention performed on the letter sequence models and period encodings). For instance, in the arithmetic setting, we might start with an input of \textit{-2 + 3 * -4} and then edit the filler $3$ to become $8$, which should change GPT-OSS's output from -14 to -34. 
These filler-based interventions generally attain high accuracies; e.g., the aforementioned type of arithmetic intervention yields an accuracy of 0.978. Figure~\ref{fig:coding_causal_interventions} gives some examples of how filler-based interventions (in the code execution task) can cause fairly substantial changes in the output.

\begin{figure}[t]
    \centering
    \includegraphics[width=0.79\linewidth]{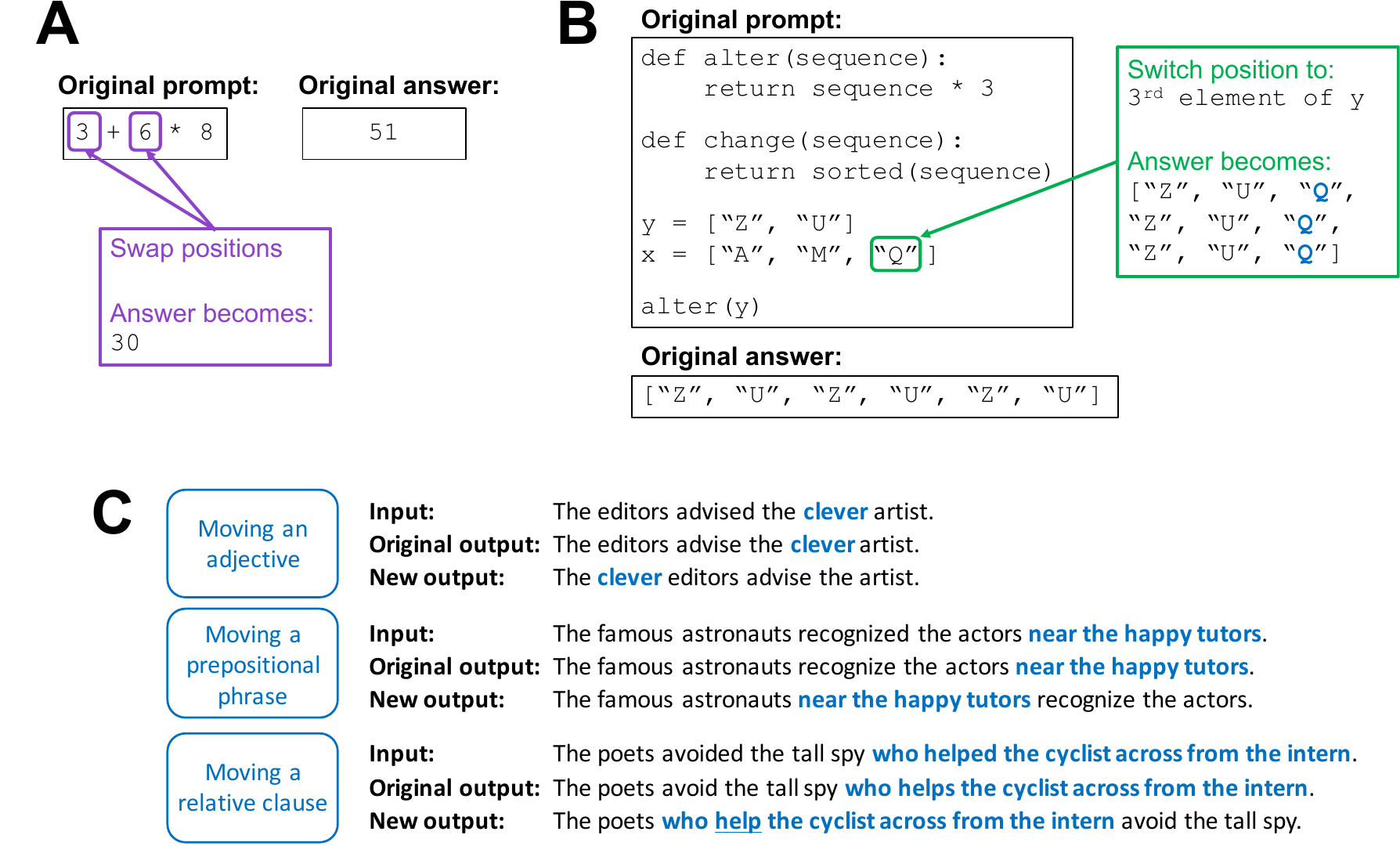}
    \caption{Causal interventions on GPT-OSS that change the structure of the input. Such interventions are conducted by using a DISCOVER approximation to edit the role of an element in the input. \textbf{A.} Swapping the positions of two numbers in an arithmetic expression (average intervention accuracy: 0.978). \textbf{B.} Moving a letter from one list to another in Python code (average intervention accuracy: 0.950). \textbf{C.} Moving modifiers within a sentence in the tense reinflection task (in these examples, the modifier always moves from the object to the subject). Moving a prepositional phrase or relative clause requires editing many role-filler pairs; further, the movement of the relative clause requires one verb to change from singular to plural in the output (underlined). Average intervention accuracy: 0.980 for adjectives, 0.892 for prepositional phrases, 0.958 for relative clauses.}
    \label{fig:gpt_oss_causal_roles}
\end{figure}

\paragraph{Changing roles:} TPRs involve explicit representations of structural positions, via roles. Therefore, our DISCOVER approximations should make it possible to change the position of an element in a structure by editing that element's role.
For example, if we start with the input \textit{The economists stopped the polite tourists}, we should be able to change the position of \textit{polite} by subtracting the representation of \textit{polite} bound to its current position (\textit{object adjective}) and adding in the representation of \textit{polite} bound to a new position (\textit{subject adjective}); after this intervention, GPT-OSS should behave as if its input had been \textit{The polite economists stopped the tourists}.
Such role-modifying interventions achieve high accuracy; see Figure~\ref{fig:gpt_oss_causal_roles} for examples.\footnote{Since the original Transformer \citep{vaswani2017attention} included explicit positional representations that were added to the input, one might wonder if our role-changing interventions merely operate by adjusting explicit positional information from the positional embedding space. However, GPT-OSS---the model we study here---handles positions with a different approach, RoPE \citep{su2024roformer}, that does not involve explicit vector representations of positions (it only modifies the queries and keys during attention rather than directly contributing positional information to the residual stream). Thus, the positional information that we are intervening on must be implicit/emergent rather than a built-in part of the architecture.
}

\paragraph{Complete vs.\ local interventions:} Our DISCOVER analyses involve the hypothesis that each token's representation encodes that token as well as all preceding tokens. Accordingly, our interventions that edit the filler or role of a token have involved modifying the representations of not just that token but also many others. One might wonder, however, if editing only the token of interest would suffice to attain the high intervention accuracies we have seen; perhaps the traces that tokens leave in the representations of other tokens have little impact on model behavior.
To test this possibility, we reran our interventions with only the representations of the target token(s) being modified (see Appendix~\ref{app:gpt_oss_causal_interventions_details} for details). In general these local interventions continued to perform almost as well as the full interventions when a filler was being edited, but when a role was being edited (i.e., when we were performing a structural change) performance dropped substantially in the local case. This result suggests that the representations of token identity are fairly localized to individual tokens in GPT-OSS, while the representations of structure are more distributed across tokens.


\paragraph{Structure-sensitive vs.\ non-structure-sensitive interventions:} Given that a role scheme based on linear order (i.e., \texttt{bidirectional (all)}) gave reasonably strong approximations in Section~\ref{sec:gptoss}, one question that one might have about our structure-changing interventions is whether they truly involve editing abstract structure or if they work purely by manipulating linear position. Appendix~\ref{app:intervention_structure_sensitive} contains an experiment on passivization and tense reinflection motivated by this question. This experiment finds that the interventions are indeed sensitive to sentence structure: interventions that respect sentence structure achieve high accuracies, while interventions that respect linear order but not sentence structure attain low accuracies. 

\subsection{Causal intervention summary} 

Across all of the model types that we have approximated with DISCOVER, we have found that we can make fine-grained edits to a model's behavior by editing components of its internal representations. This finding shows that these models are sensitive to the compositional structure of the TPRs that underlie our DISCOVER analyses.

\section{Testing generalization to novel role-filler pairs}\label{sec:ood}

A critical part of DISCOVER (distinguishing it from much other interpretability work, such as work using sparse autoencoders) is that it hypothesizes that neural network representations capture pairs of roles and fillers bound together, rather than only capturing atomic concepts. Consider the three representations of the sentence \textit{cats chase dogs} shown below. Proposal~\ref{ex:systematicbinding} uses systematic combinations of roles and fillers, while the two proposals in \ref{ex:atomicboth} use only atomic concepts, with each atomic concept capturing either a filler \ref{ex:atomicfillers} or a role-filler pair \ref{ex:atomicrfpairs}: 

\newlength{\topcategorywidth}
\settowidth{\topcategorywidth}{Concepts = role-filler pairs:\hspace{1.5em}}

\newlength{\bottomcategorywidth}
\settowidth{\bottomcategorywidth}{Concepts = role-filler pairs:}

\ex. \makebox[\topcategorywidth][l]{Systematic role-filler binding:} \{\textit{subject}:\texttt{cats}, \textit{verb}:\texttt{chase}, \textit{object}:\texttt{dogs}\}\label{ex:systematicbinding}

\ex. Atomic concepts:\label{ex:atomicboth}
\a. \makebox[\bottomcategorywidth][l]{Concepts = fillers:} \{\texttt{cats}, \texttt{chase}, \texttt{dogs}\}\label{ex:atomicfillers}
\b. \makebox[\bottomcategorywidth][l]{Concepts = role-filler pairs:} \{\texttt{cats-as-subject}, \texttt{chase-as-verb}, \texttt{dogs-as-object}\}\label{ex:atomicrfpairs}

Role-filler binding as in \ref{ex:systematicbinding} is central to our hypothesis about how neural networks encode structure.
We have already discussed why \ref{ex:atomicfillers} cannot be the strategy used by the target models (Section~\ref{sec:bindingproblem}): solely encoding which words are present does not capture word order, yet our target models are clearly sensitive to order. 
Therefore, positional roles must be part of our target models' representations---but are these roles systematically composed with fillers as in \ref{ex:systematicbinding}, or are role-filler pairs merely encoded as atomic units as in \ref{ex:atomicrfpairs}?
The results discussed so far do not answer this question definitively because, even though DISCOVER is intended to capture systematic role-filler binding, it is possible for DISCOVER to arrive at degenerate solutions in which role-filler pairs are encoded as atomic entities (see Section~5.3.3.10 of \citet{mccoy2022implicit}).
To know for sure whether we have found systematic representations of role-filler pairs, we need to test if DISCOVER can generalize to role-filler pairs that did not appear in the training of DISCOVER---a question that the experiments in this section address.

The key point underlying these experiments is that DISCOVER can only generalize to novel role-filler pairs if the target model has systematic role-filler binding. Otherwise, DISCOVER would have no basis for determining how the novel pairings should be represented.
Appendix~\ref{app:ood_whitebox} highlights this point using white-box models (models whose internal structure is known to us); when we apply DISCOVER to white-box models known to have systematic binding, DISCOVER robustly generalizes to novel role-filler pairs; but when we apply DISCOVER to white-box models known to lack systematic binding, then DISCOVER fails to generalize to novel role-filler pairs.

\paragraph{Methods:} We test for generalization to novel role-filler pairs by training DISCOVER in the same way as before, except that during the training phase we withhold certain role-filler pairs. For instance, in the letter sequence experiments, none of the DISCOVER training examples contain \texttt{C} in the 3rd position. We then evaluate how well DISCOVER generalizes to examples that contain one or more of the withheld role-filler pairs. In each experiment, the evaluation set is split into multiple subparts containing examples with differing numbers of withheld role-filler pairs so that we can analyze the extent to which DISCOVER's performance drops off as the evaluation examples contain increasing numbers of unfamiliar combinations.  See Appendix~\ref{app:ood_details} for details about which role-filler pairs are withheld. The withholding of role-filler pairs is only done when training 
DISCOVER, not when training the target model; this is because the withholding is not meant to test whether the target models show systematic generalization behavior (which is often what such withholding is used for in prior work); rather, it is way to assess the extent to which the representations of the target models have systematic structure. 

As a strong baseline, we compute the best accuracy that DISCOVER could achieve if the target model does not combine roles and fillers in a systematic way.
This baseline accuracy is $\frac{1}{n!}$ for examples involving $n$ role-filler pairs that DISCOVER was not trained on. To derive this number, suppose we have applied DISCOVER to a letter-sequence-copying model and that we are testing it on the example \texttt{\underline{A} M \underline{C} \underline{D} Q W}; the underlining indicates which role-filler pairs were not present in DISCOVER's training---e.g., it has never encountered \texttt{A} as the first element. For the role-filler pairs that it has encountered, DISCOVER can confidently place each filler in the correct output position, yielding \texttt{\_ M \_ \_ Q W}. That leaves three leftover fillers (\texttt{A}, \texttt{C}, and \texttt{D}) and three empty slots. This baseline assumes that fillers are not bound to roles in a way that can be systematically leveraged, so no strategy can beat randomly placing the remaining letters into the remaining slots. There are $n!$ ways to put $n$ items into $n$ slots (in this case, $n=3$, so $n! = 6$), but only one of those options is correct, yielding the accuracy mentioned above: $\frac{1}{n!}$. We call this strategy the \textit{strong chance baseline}.\footnote{This name contrasts the strong chance baseline from a weaker baseline in which each novel role-filler pair is decoded by producing a random filler from the vocabulary, yielding an accuracy of $(1/|V_f|)^n$ where $|V_f|$ is the number of possible fillers. We do not include this weaker baseline in our plots because it usually yields an accuracy very close to 0.} 
Note that DISCOVER does not need to be above the baseline for \textit{all} values of $n$ in order for us to conclude that there is systematic binding. As long as it exceeds the baseline for \textit{some} values of $n$, that is enough to be convincing evidence; the cases where it fails to exceed the baseline could be due to extraneous factors such as noise.

As a final methodological note, the experiments in this section used $L_{2,1}$ regularization, which we found to be helpful for DISCOVER models to generalize to novel role-filler pairs. See Appendix~\ref{app:l21} for discussion.

\begin{figure}[t]
    \centering
    \includegraphics[width=0.83\textwidth]{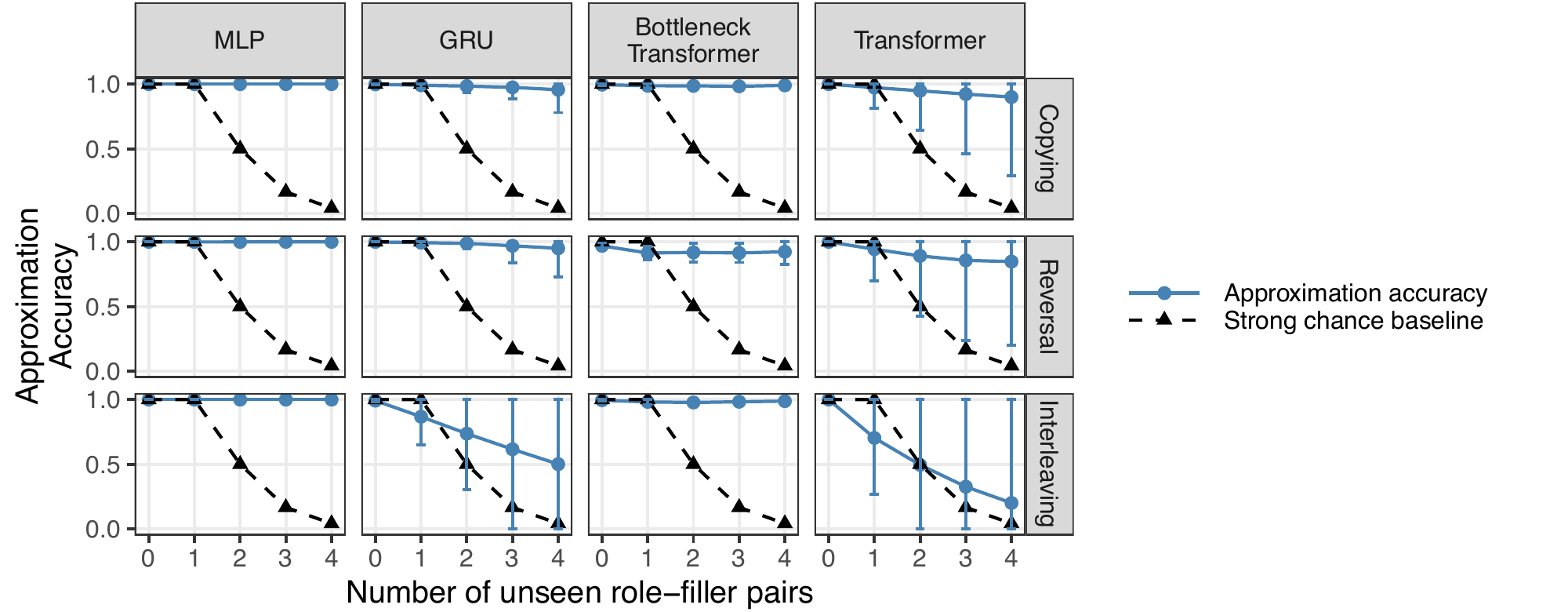}
    \caption{Testing how DISCOVER generalizes to novel role-filler pairs when it is trained to approximate the encodings of models trained to manipulate sequences of letters. The dots show the mean across 10 reruns, while the error bars show two standard deviations (truncated at 0.0 and 1.0).}
    \label{fig:tpe_acc_ood_letter_seq}
\end{figure}

\begin{figure}
    \centering
    \includegraphics[scale=0.51]{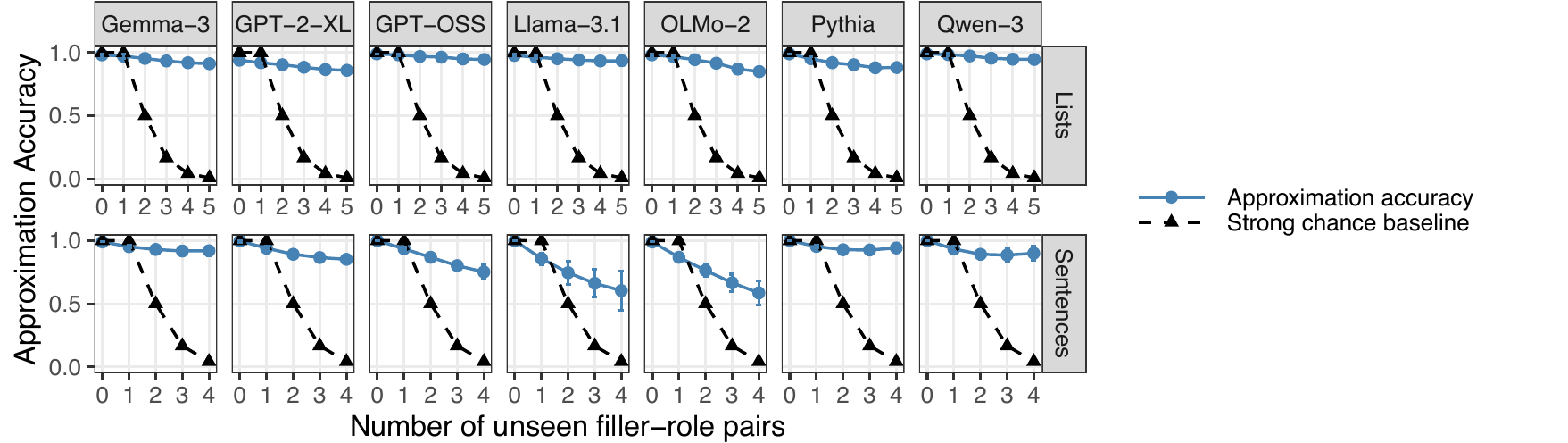}
    \caption{Testing how DISCOVER generalizes to novel role-filler pairs when it is trained to approximate LLM period encodings after lists of length 5 (top) or sentences of the form ``the NOUN VERBED the NOUN, and the NOUN VERBED the NOUN.'' (bottom). This plot shows the results for the middle layer of each LLM (the one labeled ``50\%'' elsewhere in this paper); see Figure~\ref{fig:tpe_acc_ood_period_full} in the Appendix for other layers. The dots show the mean across 10 reruns, while the error bars show two standard deviations (truncated at 0.0 and 1.0).}
    \label{fig:tpe_acc_ood_period}
\end{figure}

\begin{figure}[t]
    \centering
    \includegraphics[scale=0.51]{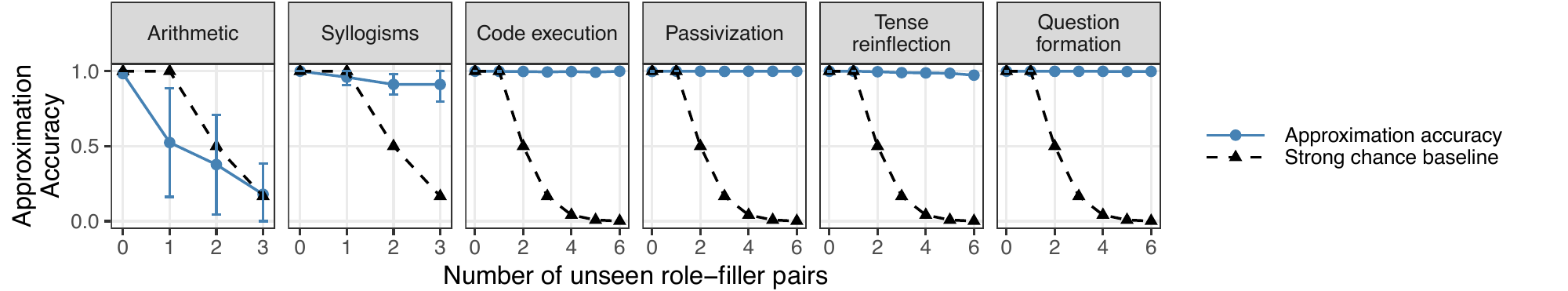}
    \caption{Testing how DISCOVER generalizes to novel role-filler pairs when it is trained to approximate the encodings of GPT-OSS performing symbolic tasks. The dots show the mean across 5 reruns, while the error bars show two standard deviations (truncated at 0.0 and 1.0).}
    \label{fig:llm_tpe_acc_ood}
\end{figure}

\paragraph{Results:} 
For the DISCOVER approximations of letter sequence models (Figure~\ref{fig:tpe_acc_ood_letter_seq}), all cases except for the interleaving Transformer show average performance that is well above the baseline for larger numbers of withheld role-filler pairs; some cases show large error bars indicating variability across re-runs of the experiment, but others show DISCOVER retaining near-ceiling accuracy across reruns and regardless of how many withheld role-filler pairs there are.
The period encoding cases also show robust generalization that is substantially above the baseline across all models (Figure~\ref{fig:tpe_acc_ood_period}).
Finally, for all tasks except arithmetic, the instances of DISCOVER applied to GPT-OSS substantially outperform the baseline for larger numbers of unseen role-filler pairs (Figure~\ref{fig:llm_tpe_acc_ood}).

In sum, for almost all target models we have considered, DISCOVER robustly generalizes to novel role-filler pairs when trained to approximate those target models. Therefore, the target models' representations must incorporate systematic binding between roles and fillers---otherwise DISCOVER could not generalize to novel combinations of roles and fillers. 
This result provides crucial evidence for this paper's core representational hypothesis: the learned representations of neural networks do not solely encode atomic concepts but rather encode systematic role-filler pairings (perhaps instead of, or perhaps in addition to, encoding atomic concepts).

\section{Discussion}

Across a wide range of models, we have found implicit, emergent symbolic structure in the representations of neural networks.
This implicit structure takes the form of linearly-transformed Tensor Product Representations. 
The fact that this structure emerged across many different types of models and training tasks is evidence that it is not simply a quirk of one type of model but rather is at least a fairly general property of how neural networks encode symbolic structure.
Further, these approximations enable us to modify model behavior in targeted ways through controlled interventions on internal representations, and these approximations generalize out-of-distribution to vocabulary elements in novel positions; these two latter points together show that the structure we have identified is systematic and that models rely on this structure when producing their behavior.

\subsection{The relationship between symbols and neural networks: A symbolic victory?}\label{sec:discussionsymbolic}

In cognitive science, there is a longstanding debate about the relationship between neural networks and symbolic systems \citep{fodor1988connectionism,pater2019generative,griffiths2025whither}. The two most prominent positions in this debate are \textbf{eliminativism} and \textbf{implementationalism}. 
The eliminative position is that the successes of neural networks allow us to eliminate symbols from cognitive theories, in favor of theories that are stated in terms of neural network computation \citep{rumelhart1986general}. 
Under this view, symbolic theories are ``occasionally useful but often misleading'' \citep{mcclelland2010letting}: they might give a rough characterization of aspects of cognition, but they do not faithfully reflect the actual cognitive processes that occur in the brain.
In contrast, the implementational position is that the only way a neural network could capture symbol-system-like behaviors is by internally implementing a symbol system \citep{fodor1988connectionism,marcus1998rethinking}.

Recent advances in AI might seem to strongly support the eliminative view because systems without explicit symbolic structure are now the state of the art in domains long viewed as symbolic. However, the results of the current paper go against this view because we have shown that, although such systems lack \textit{explicit} symbolic structure, they nonetheless have \textit{implicit} symbolic structure. Thus, such systems should not be taken as evidence that symbols can be eliminated from theories of advanced cognition; symbols are not only important in the training of such systems \citep{griffiths2025whither} but also in their internal processing.

Our results might be taken to support the implementationalist view that successful neural networks must implement symbol systems, but we argue against this conclusion.
Rather, we believe that the view that is most likely to be correct is the in-between perspective of \textbf{limitivism} \citep{smolensky1988proper}, 
which is the proposal that neural networks approach symbol systems in the limit but in practice only realize symbol systems approximately rather than precisely.\footnote{We remain agnostic about which type of limit is relevant here. Some candidates are the limit as the amount of training approaches infinity, the limit as computation (e.g., network size) approaches infinity, the limit as numerical precision approaches infinity, the limit as soft versions of discretization approach true discretization \citep{merrill2019sequential}, and the limit as noise approaches zero.}
This proposal is not eliminativist because it has symbols---neural symbols---playing a central role in defining the algorithmic and representational structure of neural networks, but it is also not implementational because networks only implement symbol systems approximately rather than precisely.

One reason to view our work as supporting limitivism is that DISCOVER usually achieved high accuracy but not perfect accuracy at approximating target models, which supports the view that these models are approximately but not exactly symbolic. For instance, the average accuracy of our causal interventions on GPT-OSS was 0.903; the fact that these interventions usually worked shows that GPT-OSS's behavior is causally influenced by some version of the symbols we targeted, but the fact that this accuracy is meaningfully lower than 1.0 shows that it does not treat these internal symbols in precisely the same way as a true symbol system would. 

Another one of our results that more directly supports the ``limit'' aspect of limitivism is the fact that, when we approximated LLM period encodings from complex sentences, the period-unpacking models performed substantially better when decoding from DISCOVER approximations than when decoding from the original LLM encodings (compare ``bidirectional'' to ``reconstruction from LLM encodings'' in Figure~\ref{fig:tpe_accs_period}, bottom; e.g., for GPT-OSS's middle layer, the period-unpacking model's accuracy was 0.71 when given LLM encodings yet 0.96 when given \texttt{bidirectional} DISCOVER approximations). Recall that the period-unpacking models take in an LLM period encoding and aim to reconstruct the sentence that preceded the period.
The DISCOVER model was trained to approximate the LLM period encodings, and it was evaluated by feeding its approximations to the period-unpacking model. What is striking is that, even though the period-unpacking model was trained on actual LLM period encodings, it performs better when fed DISCOVER approximations of those period encodings. 
We believe the best interpretation of this result is the following: The LLM period encodings are characterized by an approximately-but-not-precisely symbolic structure. The period-unpacking model picks up on this structure and uses it to produce its output.
When DISCOVER is trained on period encodings, it precisely realizes the symbolic structure that is only approximate in the actual period encodings. Since this structure is what the period-unpacking model relies on, it performs better with DISCOVER's precise version of this structure than with the approximate version of the structure that is present in the actual period encodings.
This idealized structure must be truly ``present'' in the period encodings in some sense, or else the period-unpacking model trained on the period encodings would not come to rely on this structure, but the discrepancy between the period-unpacking model's accuracy when fed LLM encodings vs.\ DISCOVER approximations shows that it remains only approximately realized in the period encodings.
This overall picture is the sort of scenario predicted by the limitivist view.

To the extent that neural networks deviate from precise symbol structures, an important goal for future work is understanding the nature of these deviations \citep{baroni2022proper,mcgrath2024can}. 
It could be that the deviations are merely noise, but there are important reasons to believe instead that the deviations equip neural networks with abilities that could not be achieved by purely symbolic systems---e.g., by enabling a seamless integration of computation that handles the systematic, compositional aspects of cognition with computation that handles the fuzzy, statistical aspects of cognition \citep{piantadosi2024concepts,lampinen2024language,rodriguez2025characterizing}.
Such integration might be central in enabling LLMs to process language \citep{boleda2025llms,futrell2026linguistics}: it has been frequently argued that much of the crucial structure within natural language is not discrete and not combinatorially productive, and that this aspect of language structure is missed by symbolic theories \citep[e.g.][]{bybee2005alternatives}.

Even in cases where neural networks do precisely implement symbol systems, it could be that having a symbol system implemented in a neural network has consequences that make it meaningfully different from a discrete-symbol-based implementation. For instance, a neural implementation might display graceful degradation \citep{mcclelland1986appeal} that makes it more robust to certain types of noise than a discrete implementation would be; and it might be more readily learnable from data because its differentiability enables powerful gradient-based learning methods that are not available within discrete symbolic computation.

An additional issue in debates about symbols relates to innateness: Is symbolic processing an innate aspect of human cognition? 
Some have argued that aspects of symbolic computation are a central part of what is innate in humans \citep{fodor1975language,hauser2002faculty}, and very young infants display abilities that appear to involve symbol processing \citep{marcus1999rule,gomez1999artificial,cesana2018precursors}. However, others argue that symbolic cognition might emerge from learning rather than being innate \citep{mcclelland2010letting,santoro2021symbolic}.
The structure that DISCOVER reveals is unlikely to be innate in the target models we have studied, so our results are evidence that symbolic structure can emerge over the course of learning inside systems that are not innately provided with symbols.
Note, though, that the emergence of symbolic structure might depend on training data generated by symbolic systems \citep{griffiths2025whither}; further, attaining human-like cognition from the quantity of data that humans experience might require symbolic predispositions, since the systems we have studied likely received substantially more training data than humans need in order to acquire the same abilities; e.g., LLMs receive orders of magnitude more linguistic input than children do \citep{frank2023bridging}.

\subsection{Other types of role-filler representations}\label{sec:other_vsas}

TPRs are just one approach for encoding symbolic structure in vector space; many other approaches exist, often grouped under the term vector symbolic architectures (VSAs).
Some other prominent formalisms include Holographic Reduced Representations \citep{plate1991holographic}, Multiply-Add-Permute \citep{gayler1998multiplicative}, and Binary Spatter Codes \citep{kanerva1994spatter}; for overviews, see \citet{mitchell2010composition} and \citet{kleyko2021survey}. 
It turns out that all major VSAs are special cases of linearly-transformed TPRs (\citealt{smolensky2006symbolic}; \citealt{hiratani2022optimal}; \citealt{mccoy2022implicit}, Section 5.3.3), in that they can be implemented using a linearly-transformed TPR given appropriate choices for the filler vectors, role vectors, and linear transformation.
A positive consequence of this fact is that, as long as the target model uses any member of this broad family to represent structure, then DISCOVER will be able to approximate the target model; a negative consequence is that a successful DISCOVER analysis does not tell us which member of this family is used by the target model. We do not consider this negative consequence to be a  problem because we are not committed to any particular member of the VSA family but rather to the high-level structure that all such approaches share (role-filler representations with multiplicative binding and additive aggregation). One direction for future work would be to more narrowly characterize which member(s) of this representational family are being used by the target models we have analyzed.

\subsection{Extensions of DISCOVER}

This paper only scratches the surface of what must be studied to understand compositional structure in neural networks. Our results have shown that trained networks utilize a particular type of representational structure, and future work should analyze (i) how networks produce this structure using the computational mechanisms that are available to them (which, notably, usually do not include the tensor product); (ii) how downstream parts of the network process this structure; and (iii) how this structure arises during training. 
The connections made by \citet{schlag2021linear} between Transformer memory and TPRs, as well as recent advances in understanding neural network circuits \citep[e.g.,][]{lindsey2025biology}, might help in answering (i) and (ii), while (iii) connects to the growing interest in understanding neural training dynamics \citep{biderman2026position}. 

DISCOVER could also be extended to additional types of fillers and roles. 
We have only used TPRs with a discrete, atomic vocabulary of fillers and roles, but the TPR formalism can also handle fillers and roles that are continuous or that have recursive structure \citep{smolensky1990tensor}.
TPRs can encode structures in which a role has a blend of numerically-weighted fillers, or in which a filler is bound to a blend of numerically weighted roles: such Gradient Symbol Systems have been shown to enable unification of aspects of cognition that are quite discrete (like the tasks analyzed here) and those that mix discrete and continuous dimensions \citep{smolensky2014optimization,smolensky2022neurocompositional}.
Extending DISCOVER to embrace such Gradient Symbol Systems could shed considerable light on how neural computation gives LLMs their extreme power with natural language.

Finally, our analyses have focused on specific tasks and domains (e.g., arithmetic problems with a particular structure). Future work should explore how to scale up DISCOVER to broader types of stimuli, ideally matching the breadth of the stimuli that modern AI systems can handle. Doing so will likely require modifying DISCOVER to make it unsupervised such that we do not need to manually form hypotheses about what fillers and roles to expect; see \citet{soulos2020discovering} for an example of how we can reduce the supervision involved in DISCOVER.

\subsection{Implications for mechanistic interpretability}

Before you can analyze a neural network, you must decide which aspect of it to analyze---e.g., attention heads \citep{clark2019bert}, individual neurons \citep{lakretz2019emergence}, 
directions in vector space \citep{cunningham2023sparseautoencodershighlyinterpretable}, or circuits \citep{wang2023interpretability}.
Even unsupervised methods---which are called \textit{unsupervised} because they make fewer assumptions than other methods---still make assumptions about which network components should be studied.
Therefore, a critical question for mechanistic interpretability is what units of analysis drive model behavior \citep{mueller2026quest}. If we make incorrect assumptions about what a network's basic representational and processing units are, then our analyses will be doomed. 
For instance, analyzing individual neurons will not be illuminating if the network encodes information in subspaces that are not basis-aligned. 
Our work contributes to this discussion by showing that, at least in many cases, some units of analysis that causally influence model behavior are multiplicative combinations of filler and role vectors. We suggest that future interpretability methods should incorporate this finding into the assumptions that they make about how neural networks implicitly operate. 
Indeed, \citet{enyan2026unifying} show that the successes of several existing interpretability methods can be explained by the hypothesis that neural networks implicitly use TPR structure. If TPR structure does indeed underlie existing interpretability results, it is likely that it can also form a useful foundation for future work.  

\subsection{Why do neural networks perform so poorly at compositional generalization?}

One benefit of representing information in a compositional way is that it enables generalization to novel combinations of units. 
For instance, even if you have never heard the phrase \textit{purple walrus} before, you can understand it by knowing the meanings of the two words and how to combine them.
We have found consistent evidence that neural networks encode information in a compositional way. However, an extensive body of prior literature has found that neural networks perform poorly at generalizing to novel combinations of units \citep{marcus1998rethinking,lake2018scan,kim2020cogs,hupkes2020compositionality,mccoy2024embers,lewis2025evaluating}. For example, \citet{kim2020cogs} studied networks trained to take in an English sentence and produce a representation of its meaning. Such networks performed poorly at handling known words in novel positions; e.g., when networks had only ever encountered the word \textit{hedgehog} as the subject of a sentence, they often made errors when \textit{hedgehog} appeared as the object of the sentence, essentially failing to generalize to a novel role-filler combination---a type of generalization that one might hope would result from using TPR-based representations. 

One way to explain how neural networks could display compositional representations without compositional generalization is that they might only develop compositional representations for the role-filler combinations that they have encountered.
Such a scenario would produce networks whose representations are compositional but which fail to generalize compositionally. 
As an illustration of how such a situation could arise, consider the word2vec model \citep{mikolov2013efficient}, which produces a vector representation for each word in its vocabulary. \citet{mikolov2013linguistic} found that these vectors display compositional structure. For instance, the vector for \textit{king} minus the vector for \textit{man} plus the vector for \textit{woman} was approximately equal to the vector for \textit{queen}. Later work has raised concerns with these results \citep{linzen2016issues,rogers2017many}, but for the sake of argument we will assume that these additive relationships hold, as this discussion is an in-principle one. 
The word2vec architecture is simply a lookup table: each word's representation is an arbitrary vector, with nothing in the architecture requiring related words to have related representations---whatever structure exists in the word representations must emerge as a result of top-down pressures from the training process. Therefore, if the training data omitted the word \textit{queen}, word2vec's learned vectors would no longer obey the math described above, because the representation of \textit{queen} would have had no opportunity to be shaped into something reasonable. This word2vec thought experiment illustrates how a system could have compositional representations without being able to behaviorally generalize to novel combinations of features; future work could study if some version of this hypothesis also applies to architectures (such as our target models) that are more complex than lookup tables.

\subsection{Might the brain use TPRs?}

The puzzle that motivated this paper was that artificial neural networks can perform tasks that seem to require symbolic structure. The same puzzle also applies to the brain: the brain is a neural network (albeit a biological one rather than an artificial one), and it excels in seemingly-symbolic domains such as language. Our conclusions could potentially extend from artificial networks to the brain---perhaps the brain also leverages TPR structure to carry out symbolic tasks. 
We have found TPR structure across a broad range of neural architectures, suggesting that TPR structure is adopted fairly generally across types of neural networks; perhaps this generality extends as far as to the brain. However, the brain differs in important ways from artificial neural networks, so the current results certainly do not allow us to make any strong claims about biological cognition without empirical studies that look for TPRs in brain recording data; DISCOVER could directly support such studies.

One reason to be optimistic that DISCOVER could capture structure in brain data is that LLMs and their precursors have been shown to provide strong fits to brain responses \citep{jain2018incorporating,schrimpf2021neural,caucheteux2022brains,tuckute2024driving}, and we have shown that DISCOVER can provide strong fits to LLMs; by transitivity, DISCOVER might then provide strong fits to brain data. However, the stimuli that we used in our LLM analyses were different from those used in the studies that fit LLM encodings to brain data, so we cannot yet draw strong conclusions about whether TPRs can explain patterns in brain activations.

\section{Related work}\label{sec:relatedwork}

\paragraph{Analyzing neural networks with TPRs:} TPRs were introduced by \citet{smolensky1987analysis} at the first NeurIPS conference.
In the longer journal version of that work, \cite{smolensky1990tensor} raised the possibility that neural networks might naturally learn to produce TPRs: ``In the short term at least, our learning rules and network simulators do not seem powerful enough to make network learning of linguistic representation feasible. Even if such learning is feasible at some future point, we will still need to \textit{explain} how the representation is done. There are two empirical reasons to believe that such explanation will require the kind of analysis begun in this paper....'' It seems reasonable to conclude that ``some future point'' has now arrived, and the current paper indeed applies TPR-style analyses toward understanding systems that have learned their own linguistic representations.

To our knowledge, the first paper to empirically look for implicit, learned role-filler representations inside neural networks was \citet{monner2012emergent}, which analyzed a stripped-down version of an LSTM \citep{hochreiter1997long} to reveal how it builds and processes neurosymbolic representations. 
While this analysis was ahead of its time, it required detailed, architecture-specific reasoning, making it challenging to apply to new architectures or to complex role schemes. DISCOVER addresses those limitations through trainable TPRs that are agnostic to the target model architecture, enabling us to analyze a wide range of target models.

DISCOVER was introduced by \citet{mccoy2018rnns} who applied it to recurrent neural networks (RNNs) trained on synthetic data as well as to sentence embedding models.
DISCOVER was subsequently used by \citet{soulos2020discovering}, who applied it to RNNs performing the SCAN task \citep{lake2018scan} and showed that trained DISCOVER models could enable causal interventions on the target model, and \citet{jawahar2019bert}, who applied it to BERT \citep{devlin2019bert}.
The current paper expands beyond these prior papers by extending DISCOVER to a range of neural network architectures, including modern LLMs.
The dissertation of \citet{mccoy2022implicit} conducted preliminary versions of the letter-sequence experiments reported here. 

Several other papers have made connections between black-box neural networks and TPRs or related formalisms. \citet{smolensky2025mechanisms} shows that a TPR-based algorithmic structure can explain the LLM ability of in-context learning; this result serves as a theoretical demonstration of what could be achieved by Transformers, while the current paper empirically studies what strategies are naturally learned by standard systems. 
\citet{wattenberg2024relational} gives a call to action for more work to connect role-filler models and LLMs, a direction that the current paper certainly fits. 
\citet{knittel2024gpt} point out several ways in which the Transformer architecture could relate to role-filler models, and they find evidence that GPT-2 uses aspects of its architecture in the hypothesized ways; our work analyzes representations rather than processing mechanisms. \citet{lee2026tensor} and \citet{bronzini2025hyperdimensional} proposed new types of probes that decode TPR-like information from Transformer representations; while these papers use a decoding-based approach (reading information out of neural representations), DISCOVER uses an encoding-based approach (reconstructing neural representations).

Zooming out, the core novel contributions that the current paper makes beyond the papers discussed above are the following: (i) It identifies TPR structure across a wide range of neural network architectures; (ii) it identifies TPR structure in LLMs performing  tasks in several symbolic domains; (iii) it shows that the TPR structure we have identified is causally implicated in LLM behavior; and (iv) it demonstrates that the TPR structure generalizes to novel role-filler pairs. These findings are central to our core argument that systematic role-filler structure is used broadly across types of neural networks in order to perform symbolic tasks.

\paragraph{Symbolic mechanisms in neural networks:}
Several interpretability papers have identified mechanisms inside neural networks that show aspects of symbolic processing \citep{clark2019bert,olsson2022context,lepori2023break,lindsey2025biology,yang2025emergent}. We focus on representations, not processing, but
one aspect of symbolic processing that is highly relevant to our work is variable binding. \citet{lewis2024clip} uses behavioral tests to analyze whether a vision-language model binds objects and attributes; we focus on representations rather than behavior. Several papers have analyzed the mechanisms that models use to perform the type of variable binding that underlies entity tracking \citep{dai2024representational,feng2024how,gur-arieh2026mixing} or variables in computer code \citep{wu2025how}; we instead analyze how models represent structured inputs. Finally, \citet{huang2026decomposing} introduces a method for finding variable-like subspaces in neural network representations; one direction for future work is investigating whether the sorts of subspaces identified in that work are related to the roles identified in the current paper.

\paragraph{Analyzing structure in neural networks:} An extensive line of literature has investigated what types of structural information can be decoded from neural network representations, such as syntactic information \citep{ettinger2016probing,conneau2018cram,hewitt2019structural,tenney2019bert,jawahar2019bert,lin2019open,pimentel2020information,voita2020information,hall2021syntactic,li2026linear}.
These papers use decoding-based approaches in which information is read out from the target model; successful decoding indicates that the information in question is present in the target model, but
there is no guarantee that the extracted information reflects the global structure of the vector space---it might be only a small part of a representational space that also contains much other information. 
DISCOVER is instead an encoding-based approach that aims to provide a complete reconstruction of target model representations. Thus, one of the most important words in our title is \textit{of}: Decoding-based methods find structural information \textbf{in} vectors, whereas we analyze the structure \textbf{of} vectors.

Some other papers have used the general strategy of mapping the complete representational space to an interpretable symbol structure \citep{kirov2012processing,andreas2019measuring,chrupala2019correlating,lepori2020picking,liu2022representations,murty2023characterizing,pandey2023syntax,geiger2024finding}. Despite this high-level connection, the current paper differs from these other papers in the methods it uses, the models it analyzes, and the representational structure that it is based on (i.e., TPR structure). Like the other approaches mentioned in this paragraph, sparse autoencoders \citep{cunningham2023sparseautoencodershighlyinterpretable,templeton2024scaling} analyze the complete structure of representations, but they do so in a bottom-up way that is usually only partially interpretable.

\paragraph{Integrating TPRs into neural networks:} While the current paper analyzes whether \textit{implicit} TPR structure emerges in standard neural networks, there is also a line of literature aimed at \textit{explicitly} incorporating TPRs and related structures into neural networks, typically as a way to improve the network's compositional abilities \citep{palangi2018question,schlag2018learning,lalisse2019augmentic,schlag2019enhancing,soulos2021structural,jiang2021enriching,soulos2023differentiable,soulos2024compositional}. For an overview of such work, see \citet{smolensky2022neurocompositional}. One possible way in which DISCOVER might be beneficial for such work is that understanding of what sorts of representations naturally emerge in trained networks might inform techniques for explicitly encouraging particular types of structure (e.g., if the emergent structure has shortcomings that can be identified, we might then want to encourage representations that overcome those flaws; or if the emergent representations are robust, we might want to provide inductive biases that can encourage such representations to emerge with less training).

\section{Conclusion}

Neural computation and symbolic computation appear to differ radically and irreconcilably, yet our results reveal that there is not necessarily a tension between the two: A single system can be simultaneously neural and symbolic by using a neural architecture to construct symbolic representations.
Specifically, we have shown via the Tensor Product Representation formalism that the representations of a variety of neural networks can be closely approximated with symbolic structures, providing an interpretable closed-form equation for the representations of these networks.
All the domains that we have analyzed are fully systematic, meaning that one could write a simple symbolic program that would perfectly perform the task. 
However, there are also many domains that are only partially systematic, where neural networks outperform all known symbolic algorithms; most tasks that involve naturalistic language fall into this category because processing natural language requires fuzzy statistical inferences in addition to systematic structure.
An important future direction would be understanding not just the representations used by neural networks in fully systematic domains (which is what we study here) but also the representations used in settings that are only partially systematic. Such an understanding would illuminate how a system can unify systematic compositionality with gradience and idiosyncrasies---a type of unification that large language models seem to have achieved, at least in a partial sense, yet which remains poorly understood.


\section*{Acknowledgments}

We are grateful to the many people and groups who have filled the role of \texttt{helpful discussion partner}: Jacob Andreas, Suhas Arehalli, Isabel Baird, Tyler Brooke-Wilson, Jonathan Cohen, Ewan Dunbar, Roland Fernandez, Robert Frank, Jianfeng Gao, Tom Griffiths, Jan H\r{u}la, Najoung Kim, Philipp Koehn, Géraldine Legendre, Xiaodong Liu, Grusha Prasad,  Alan Prince, Kaya Stechly, Abigail Tenenbaum, Ruben Van Genugten, Marten van Schijndel, Colin Wilson, Aditya Yedetore, Enyan Zhang, Herbert Zhou, the Johns Hopkins Gradient Symbolic Computation Group, the NYU Computation and Psycholinguistics Lab, the Deep Learning Group at Microsoft Research, Redmond, the Computational Linguistics at Yale lab, and the Yale Mechanistic Interpretability discussion group.
Portions of this research were supported by the National Science Foundation Graduate Research Fellowship Program under Grant No.\ 1746891, National Science Foundation grants BCS-2114505, IIS-2239862 and IIS-2504953,  and Microsoft Research Redmond Deep Learning. 
For computing resources, we thank the Yale Center for Research Computing.
Any errors are our own.

\newpage
\bibliographystyle{aaai24}
\bibliography{main}

\appendix
\addtocontents{toc}{\protect\setcounter{tocdepth}{1}}

\section{Linearly-transformed TPRs vs.\ basic TPRs}\label{app:need_for_linear_transformation}

DISCOVER aims to approximate the representations of a neural network using linearly-transformed TPRs. There are two reasons why we use linearly-transformed TPRs rather than basic TPRs.

First, the linear transformation gives us flexibility over the size and shape of the representation that is generated. In a basic TPR, the representation must be a matrix of size $d_f \times d_r$, where $d_f$ is the size of the filler vectors and $d_r$ is the size of the role vectors. 
Our target models always encode information in vectors rather than matrices, which necessitates some transformation that can turn a matrix into a vector. A simple choice for such a transformation would be flatten the matrix into a vector by concatenating its rows; however, the resulting vector would have size equal to $d_f d_r$, but it might be that the size of the target model's representations does not permit a precise factorization into two integers that can serve as reasonable values for $d_f$ and $d_r$. For instance, if the target model uses representations with a prime number as the size---e.g., vectors of size 127---then either the filler or role vectors would need to be vectors of size 1, a situation that is unlikely to be able to capture much richness in the representational space in question. Including the linear transformation removes such strict constraints on size: We can specify $d_f$ and $d_r$ to be whatever we want and then use a linear transformation that converts $d_f \times d_r$ matrices into vectors of the size used by the target model.

The other reason why we include a linear transformation on top of basic TPRs is that it allows DISCOVER to capture degrees of freedom that do not meaningfully change the conceptual structure of the vector space but that might be important for approximating the target model. 
At a high level, the intuition is similar to including a constant multiple when correlating two single-dimension variables: e.g., ``height in meters'' and ``height in feet'' are effectively the same quantities because they are identical \textit{up to a constant multiple}. Our use of a linear transformation is meant to be the analogue of a constant multiple for multi-dimensional vector space: successful DISCOVER results show that the target model's representations are equivalent to TPRs \textit{up to a linear transformation}.
More precisely, without a linear transformation, the target model's representations could only be approximated by TPRs if the elements of the target model's representations and the elements of the TPR are in a one-to-one correspondence and appear in the same order as each other.
However, there are many ways in which the two representational spaces could be effectively the same as each other without having this direct alignment. They might use the same elements as each other but in a shuffled order. They also might use the same conceptual structure as each other but encoded in ways that are stretched and rotated with respect to each other, such that their coordinates are not in a one-to-one correspondence even if reordered (see Figure~\ref{fig:tpr_structure}C). 
In such situations, the target model would still be utilizing representations that follow the same role-filler structure as the TPR in question, but a basic TPR could not approximate the target model's representations due to the reordering, rotation, and stretching that superficially distinguish the target model from the TPR. 
Adding a linear transformation, however, makes it possible to capture such situations.

\section{Technical details for letter sequence models}\label{app:letter_sequence_technical_details}

\subsection{Target models}\label{app:letter_sequence_models}

For all target models, the output that they are trained to produce is the intended output sequence followed by a special end-of-sequence token \texttt{<EOS>} that indicates that the output sequence is finished. E.g., if the task is reversal and the input is the sequence \texttt{Q M Z}, the intended output would be \texttt{Z M Q <EOS>}. The model's output is taken to end at the leftmost position at which an \texttt{<EOS>} token is produced; if it never produced an \texttt{<EOS>} token (instead continuing to generate text until its maximum output length is reached), its output is taken to end at the leftmost token at which a padding token is produced; if it never produced an \texttt{<EOS>} token or a padding token, its output is taken to constitute all tokens that it produces. The \texttt{<EOS>} token was considered part of the intended output, such that a model was only considered correct if it produced precisely the correct target sequence followed by the \texttt{<EOS>} token.
We train all models with a dropout proportion of 0.1 \citep{srivastava2014dropout}; see the description of each model for where the dropout is applied.

\paragraph{MLP:} We use multi-layer perceptrons with an embedding size of 10 and a hidden size of 64. The input sequence is padded to the maximum sequence length $l_{max}$ (i.e., 6), since MLPs require a fixed-size input. 
The sequence elements are embedded and concatenated to form one vector of size $10*l_{max}$. 
This vector is then passed through a dropout layer, a linear layer, a ReLU layer, a linear layer, a ReLU layer, a dropout layer, a linear layer, and then a final linear output layer, producing a vector of size $|V|(l_{max}+1)$, where $|V|$ is the vocabulary size (i.e., 29: the 26 letters, a beginning-of-sequence token that is never used, an end-of-sequence token that should be produced after the rightmost letter in the output, and a padding token). This size uses $l_{max}+1$ rather than $l_{max}$ because the model outputs have the \texttt{EOS} token appended to them, making the output lengths 1 greater than the input lengths.
That output vector is partitioned into $l_{max}+1$ pieces (each having size $|V|$), each of which is passed through a softmax and then treated as the prediction for the token at that position. 
The vector that is approximated by DISCOVER is the one after the first ReLU layer.

\paragraph{GRU:} We use 1-layer sequence-to-sequence GRUs with an embedding size of 10 and a hidden size of 64. Dropout is applied to the token embeddings in the encoder and the decoder. The final hidden state of the encoder (which also acts as the initial hidden state of the decoder) is the vector that is approximated by DISCOVER.

\paragraph{Transformer:} We use 4-layer sequence-to-sequence Transformers with an embedding size of 64, a hidden size of 64, a feedforward dimensionality of 256, and 8 attention heads (note that, in Transformers, the embedding size must be the same as the hidden size). As is standard in sequence-to-sequence Transformers, the encoder used bidirectional attention (no masking), while the decoder was autoregressive, with each decoder token using causal masking when attending to other decoder tokens but not using any masking when attending to encoder tokens. The encoder and decoder used sinusoidal positional encodings, following the original Transformer \citep{vaswani2017attention}. 
Dropout is applied to the token embeddings (after the positional embeddings have been incorporated) in the encoder and the decoder. DISCOVER approximates each final-layer hidden state of the encoder, analyzing each one as a single role-filler pair (i.e., the filler is whichever token is at that position, and the role is the role that is assigned to that token under the role scheme in question). 

\paragraph{Bottleneck Transformer:} The bottleneck Transformer is identical to the standard Transformer (with all of the same hyperparameters) except that the cross-attention in the decoder can only attend to the first token in the input sequence, rather than all input tokens as is the case with the standard Transformer.
This constraint forces the first token's representation to act as a single-vector encoding of the entire input sequence. Note that our Transformer encoders use bidirectional attention, which is what makes it possible for the first token's representation to capture information about all input letters. 

\subsection{Training target models}\label{app:letter_sequence_target_training}

\begin{figure}
    \centering
    \includegraphics[width=0.57\textwidth]{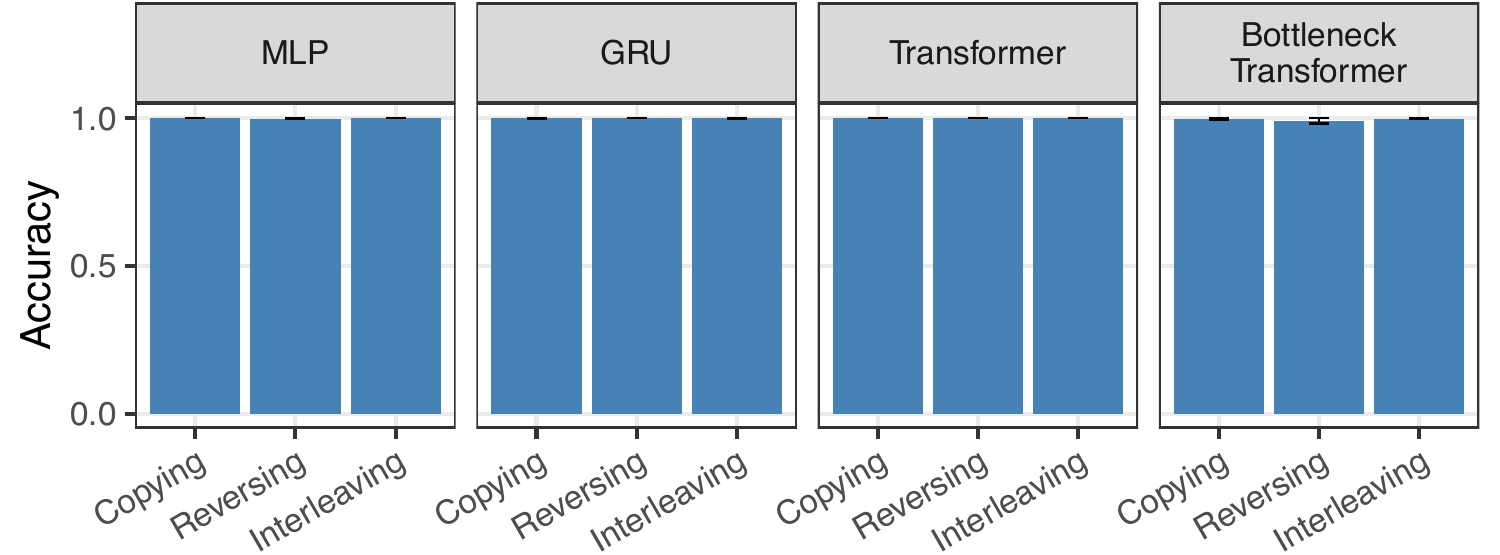}
    \caption{Target model accuracies on tasks that involve manipulating sequences of letters. The bars show the mean across 10 reruns, with error bars showing two standard deviations above and below the mean.}
    \label{fig:letter_seq_target_acc}
\end{figure}

The target models are trained using negative log likelihood loss. For the architectures whose decoders generate autoregressively (i.e., all architectures except the MLP), we use teacher forcing during training, but when evaluating models we turn off teacher forcing.
The initial learning rate is 0.001.
We use a batch size of 32 and evaluate models on the validation set after every 300 batches, saving the model weights whenever it achieves a lower validation loss than it has at any previous checkpoint. If the model has gone through 5 of these evaluations without the loss on the validation set improving, the model weights are reloaded from the last saved checkpoint (i.e., the one with the lowest validation loss), and the learning rate is cut in half. Training is halted at the point when the sixth such learning rate decrease would apply. The model weights are then reloaded from the last saved checkpoint (i.e., the point during training that achieved the lowest validation loss), and the model is evaluated on the withheld test set. Figure~\ref{fig:letter_seq_target_acc} shows the performance of these target models on their training tasks.

\subsection{Training DISCOVER models}\label{app:letter_sequence_discover_training}

For DISCOVER models, we used an embedding size of 20 for both the fillers and the roles. The TPR that results from these filler and role embeddings thus has a size of $20*20 = 400$, but we are aiming to approximate encodings from target models that used a hidden size of 64. To address this inconsistency, we applied a trainable linear layer (which included a bias term) that compressed the 400-dimensional TPRs into 64-dimensional vectors.

The DISCOVER models were trained to minimize the mean squared error between the target model encodings and the encodings produced by DISCOVER. They otherwise followed the same training procedure as the target models (using a batch size of 32, evaluating after every 300 batches, and using the same procedure for decaying the learning rate and determining when to stop training). The dataset splits used for training the DISCOVER models were the same as the ones used to train the target models.

The trained DISCOVER models were then evaluated using approximation accuracy on the test set: When the target model's decoder is fed the DISCOVER model's encodings for the test set inputs, what proportion of the time does the target model produce the correct output sequence?

\section{Analyzing models trained on alphabetizing a list}\label{app:sorting}

\begin{figure}
    \centering
    \includegraphics[scale=0.4]{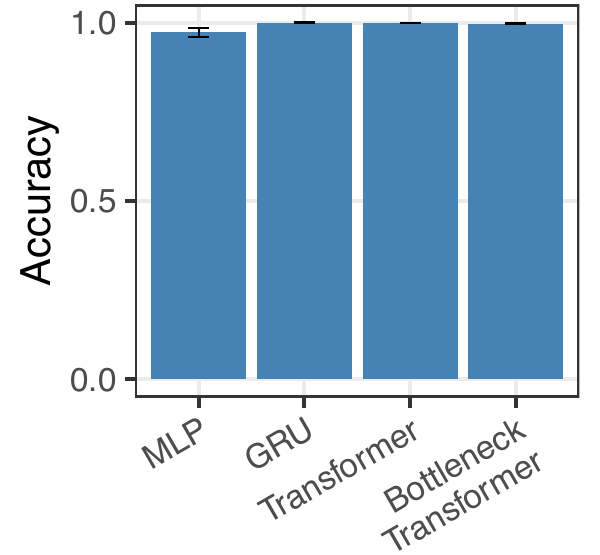}
    \hfill
    \includegraphics[scale=0.4]{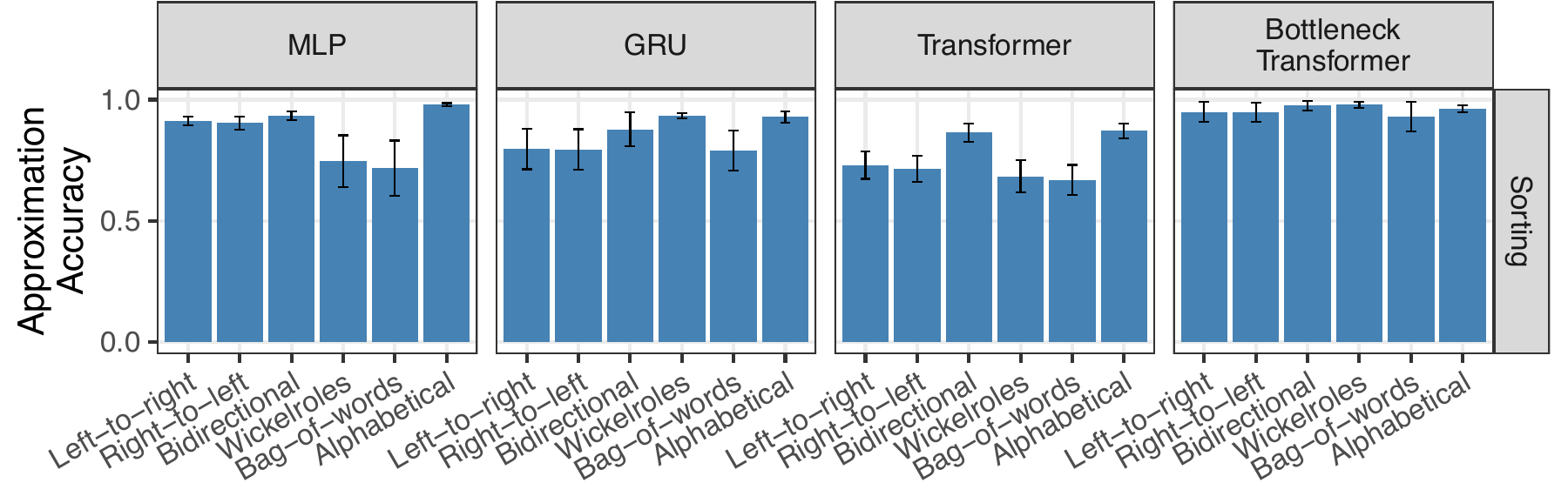}
    \caption{\textbf{Left:} Target model accuracies on the task of sorting a list into alphabetical order. \textbf{Right:} DISCOVER performance for models trained on sorting a list of letters into alphabetical order. The bars show the mean across 10 reruns, with error bars showing two standard deviations above and below the mean.}
    \label{fig:letter_seq_tpe_acc_sort}
\end{figure}

In Section~\ref{sec:letter_sequence_models}, we analyzed models trained on tasks that involve manipulating letter sequences in ways that require the order of the input letters to be encoded in the model's representations (i.e., the tasks of copying, reversing, and interleaving). Here we consider a letter sequence task for which the order of the input elements does not matter: sorting the list into alphabetical order (e.g., \texttt{Q M Z V R} $\rightarrow$ \texttt{M Q R V Z}). We use the same setup for training the target models as described in Appendix~\ref{app:letter_sequence_technical_details} except that, for the MLPs, we used a hidden state size of 128 rather than 64 since we found that a hidden state size of 64 was insufficient for the MLP to perform well on this task. After training, all four architectures scored well on this task (Figure~\ref{fig:letter_seq_tpe_acc_sort}, left); the average accuracy across reruns was 0.973 for the MLPs, and for the other three architectures the average accuracy was over 0.99.

We then applied DISCOVER to these models, with results in Figure~\ref{fig:letter_seq_tpe_acc_sort} (right). 
We have added one more role scheme not considered in the previous experiments, namely the \texttt{alphabetical} role scheme, in which each letter's role is its position in alphabetical order; e.g., in the sequence \texttt{Q M Z C}, the letters would get the roles of \texttt{3rd}, \texttt{2nd}, \texttt{4th}, and \texttt{1st}, respectively.
In stark contrast to the models trained on the order-sensitive tasks in Section~\ref{sec:letter_sequence_models}, the sorting models are all approximated at least reasonably well by the \texttt{bag-of-words} role scheme. Thus, it appears that, when the training task does not require structure to be represented, neural networks will not learn to encode structure, or at least not in a way that they rely on deeply; \citet{mccoy2018rnns} drew a similar conclusion.
Nonetheless, for most architectures, the order-sensitive \texttt{bidirectional} role scheme still outperforms the \texttt{bag-of-words} role scheme, indicating that the encodings of these models still capture order to some extent.

Though the approximation accuracies that we see are fairly high, there are some architectures for which there is still noticeable headroom in the approximation accuracy. The accuracies of the best-performing role schemes (averaged across reruns) are 0.979 for the MLP (\texttt{alphabetical} roles), 0.935 for the GRU (\texttt{Wickelroles}), 0.871 for the Transformer (\texttt{alphabetical} roles), and 0.979 for the bottleneck Transformer (\texttt{Wickelroles}); the accuracies for the GRU and the Transformer are respectable but are noticeably below the well-over-95\% accuracies that we have seen elsewhere. One possible explanation is that the sorting models might not fully use systematic TPR structure but rather only a rough approximation of it. 
Another possibility is that they might use systematic TPRs but with a role scheme we have not considered; e.g., since different types of role schemes show improvements over \texttt{bag-of-words} (e.g., \texttt{bidirectional}, which adds linear information; \texttt{alphabetical}, which adds alphabetical order information; and \texttt{Wickelroles}, which add contextual information), perhaps fully reconstructing these representations would require a role scheme that includes all of these types of information.

\section{Context-free grammar for complex sentences}\label{app:complex_cfg}

\begin{figure}[t]
\begin{center}
\small

\vskip 0.12in
\fbox{\parbox{9.5cm}{
S $\rightarrow$ NP\_RC V NP\_RC

S $\rightarrow$ NP V NP , and NP V NP

S $\rightarrow$ NP V\_emb that NP V NP

S $\rightarrow$ NP\_RC was V by NP\_RC

S $\rightarrow$ TempP , NP V NP

S $\rightarrow$ NP V NP TempP

\bigskip
TempP $\rightarrow$ Temp NP V NP

\bigskip
NP $\rightarrow$ the N

NP $\rightarrow$ the Adj N

NP\_RC $\rightarrow$ the N\_post

NP\_RC $\rightarrow$ the Adj N\_post

N\_post $\rightarrow$ N

N\_post $\rightarrow$ N RC

\bigskip
RC $\rightarrow$ who the N V

RC $\rightarrow$ who the Adj N V

RC $\rightarrow$ who V the N

RC $\rightarrow$ who V the Adj N

\bigskip
\hangindent=0.7cm
{Temp $\rightarrow$ $\{$when $|$ after $|$ before $\}$}

\hangindent=0.7cm
{V\_emb $\rightarrow$ $\{$knew $|$ believed $|$ thought $|$ said $\}$}

\hangindent=0.7cm
{N $\rightarrow$ $\{$senator $|$ journalist $|$ composer  $|$ director $|$ ... $|$ athlete $\}$}

\hangindent=0.7cm
{V $\rightarrow$ $\{$recommended $|$ visited $|$ helped  $|$ supported $|$ ... $|$ admired $\}$}

\hangindent=0.7cm
{Adj $\rightarrow$ $\{$happy $|$ famous $|$ clever  $|$ tall $|$ ... $|$ polite $\}$}

}
}
\caption{Context-free grammar for complex sentences (Section~\ref{sec:llm_period_encodings}). Some of the lists of terminals have been truncated for space; overall there are 80 nouns (\textit{N}), 16 verbs of the category \textit{V} (in addition to the 4 verbs shown in the category \textit{V\_emb}), and 10 adjectives (\textit{Adj}). See the project GitHub for the complete word lists.
} \label{fig:complex_cfg}
\end{center}
\end{figure}

The context-free grammar (CFG) used to generate sentences in the ``complex sentence'' experiments is in Figure~\ref{fig:complex_cfg}. We do not intend for this CFG to capture reasonable syntactic analyses of the English sentences in question; rather, its purpose is simply to serve as a concise way to express variable templates for the stimuli that we generate. 
Though we wanted to have sentences with reasonably complex structures, we also wanted to prevent them from becoming \textit{too} complex, and the CFG implements certain constraints toward that end. For example, relative clauses can only appear in sentences that are otherwise fairly simple, and relative clauses cannot be nested inside each other.
Before being fed into LLMs, the CFG-generated sentences were postprocessed by capitalizing the first letter, removing the space before the comma, and adding a period at the end.

\section{Rogue dimensions and z-scoring}\label{app:rogue}

\citet{timkey2021rogue} found that LLM encodings sometimes contain ``rogue dimensions''---vector positions that are extremely prominent in the representational space (e.g., dominating cosine similarity measures) yet that are not very important in model behavior.
DISCOVER aims to capture the structure of the representational space, so rogue dimensions could potentially distract DISCOVER from capturing the aspects of the representations that are most behaviorally important.
To guard against this possibility, our experiments with LLMs z-score LLM representations before analyzing them (i.e., before approximating these representations with DISCOVER or before training a decoder to read out from them).
Z-scoring is a simple way to give all vector dimensions a similar prominence as each other in the representational space; it is one of the methods for doing so discussed by \citet{timkey2021rogue}. Specifically, if an original LLM encoding is $h$, the z-scored version of it is $\frac{h - \mu}{\sigma}$, where $\mu$ is a vector containing the mean value of each vector position, and $\sigma$ is a vector containing the standard deviation of the values at each vector position. 

Since our goal is to understand LLM representations, one might feel intuitive objections to the use of z-scoring because it means that the vectors we analyze are no longer the exact vectors used by the LLMs. Below we respond to this objection for each of the three model types to which it applies. 
The basic point across all three responses is that z-scoring and its inverse are both linear transformations such that the z-scoring transformation or its inverse could be incorporated into any one of these models purely by modifying its weights, such that z-scoring does not change the expressive capacity of the models but rather only affects how easily training proceeds. 

\paragraph{Period-unpacking models:} When we train Transformer decoders to take in an LLM period encoding and then generate the sentence or list that preceded the period, we first z-score the LLM period encodings. However, given such a decoder, it would be straightforward to adjust its weights so that it produced equivalent decoding behavior when given non-z-scored LLM representations. This would be done by modifying the initial linear layer of the decoder so that it becomes the composition of the z-scoring transformation with this linear layer; this composition would still be a linear layer. 
Thus, the fact that we achieve a certain level of performance with a decoder that takes in z-scored encodings guarantees that we could achieve the same level of performance with a decoder that takes in non-z-scored encodings. 

\paragraph{DISCOVER models trained on LLM period encodings:} These DISCOVER models are trained to approximate z-scored LLM period encodings. They are then evaluated by inputting their approximations into period-unpacking models that have been trained to read out from z-scored LLM period encodings. 
The preceding paragraph describes how the period-unpacking models could be adjusted to produce the same behavior but given non-z-scored encodings (by incorporating z-scoring into the first linear layer of the decoder).
Similarly, once we have a DISCOVER model trained to approximate z-scored LLM representations, we can incorporate the inverse of the z-scoring transformation into the final linear layer of the DISCOVER model.
If we then take this DISCOVER model (modified to perform inverse z-scoring at the end) and this period-unpacking model (modified to perform z-scoring at the start), the results would be the same as if we had used the original DISCOVER model and period-unpacking model, showing that the results we have achieved with z-scoring could also have been produced using models that do not involve z-scoring.

\paragraph{DISCOVER models trained on GPT-OSS performing symbolic tasks:} These DISCOVER models are trained to approximate z-scored GPT-OSS encodings. However, we score the DISCOVER models by feeding their approximations into GPT-OSS, and before doing so we apply the inverse z-score to translate the DISCOVER representation space into the GPT-OSS representation space. Thus, all results that we report for the symbolic tasks in GPT-OSS indicate the success of reproducing the non-z-scored GPT-OSS representations; the z-scoring is used during training but is undone when evaluating the trained DISCOVER model. As described in the previous paragraph, the inverse z-scoring transformation could be incorporated into the DISCOVER model's final linear layer, such that we could achieve precisely the same results that we have achieved using just a DISCOVER model (with appropriately adjusted weights) rather than explicitly needing to apply inverse z-scoring.

\section{Technical details for LLM period encoding experiments}

\subsection{Information about the LLMs we studied}\label{app:llm_period_info}

\begin{table}[]
    \centering
    \small
    \begin{tabular}{cccc} \toprule
        LLM & Layers analyzed & Hidden state size \\ \midrule
        Gemma-3-27b & 1, 16, 31, 46, 62 & 5376 \\
        GPT-2-XL & 1, 12, 24, 36, 48 & 1600 \\
        GPT-OSS-20b & 1, 6, 12, 18, 24 & 2880 \\
        Pythia-12b & 1, 9, 18, 27, 36 & 5120 \\
        Qwen3-14b & 1, 10, 20, 30, 40 & 5120 \\
        OLMo-2-13B  & 1, 10, 20, 30, 40 & 5120 \\
        Llama-3.1-8b & 1, 8, 16, 24, 32 & 4096 \\ \bottomrule
    \end{tabular}
    \caption{Information about the LLMs whose period encodings we analyze. For all models, layer 0 is the embedding layer, so layer 1 (the earliest one we analyze) is the first non-embedding layer. The last layer that we analyze is always the latest layer in the model.}\label{tab:llm_layers_hidden_size}
\end{table}

For each LLM whose period encodings we analyzed, Table~\ref{tab:llm_layers_hidden_size} provides the specific layers that we analyzed as well as the LLM's hidden state size (i.e., the size of the vectors approximated by DISCOVER). The layers were selected to be approximately 0\%, 25\%, 50\%, 75\%, and 100\% of the way into the model's processing, but the model's number of layers was not always divisible by 4, so the layer position is sometimes not exact.

\subsection{Dataset details}\label{app:period_datasets}

For the subject-verb-object sentences, we generated all possible sentences that fit our templates and then split them into 80\% training, 10\% validation, and 10\% testing, yielding 81,920 training sentences, 10,240 validation sentences, and 10,240 testing sentences. For the lists, we used 100,000 training examples, 1,000 validation examples, and 5,000 testing examples. For the complex sentences, we used 200,000 training sentences, 5,000 validation sentences, and 5,000 test sentences.

\subsection{Training period-unpacking models}\label{app:training_period_decoders}

This section provides technical details for the period-unpacking models that we train to take in an LLM period encoding and produce the list or sentence that preceded the period.
The period-unpacking models are decoder-only Transformers with 6 layers, 16 attention heads, a hidden state size of 1024, and a feedforward dimensionality of 4096. The LLM period encoding that is provided as input is first compressed to a vector of size 1024 using a trained linear layer before being fed to the Transformer.

For each combination of a condition (lists, subject-verb-object sentences, or complex sentences), LLM, and layer, we train 5 distinct period-unpacking models on the training set for that condition. The period-unpacking models are trained using negative log likelihood loss, a batch size of 32, an initial learning rate of 0.0001, and dropout with a proportion of 0.1 applied to the embeddings after the positional embedding has been applied. After every 100 batches, the period-unpacking models are evaluated on the validation set, and the weights are saved if they achieve a lower validation loss than any previous checkpoint has; if the loss on the validation set has not improved for 5 of these evaluations, then the learning rate is halved and the model weights are reloaded from the checkpoint with the best validation loss thus far.
Training halts when the sixth of these learning rate decreases would occur, at which point the weights are reloaded from the best saved checkpoint and then used to evaluate the model on the test set.
The dataset splits are the ones described in Appendix~\ref{app:period_datasets}. The period encodings are z-scored before being provided as input to the period-unpacking models (Appendix~\ref{app:rogue}).

\begin{figure}[t]
    \centering
    \includegraphics[width=0.8\linewidth]{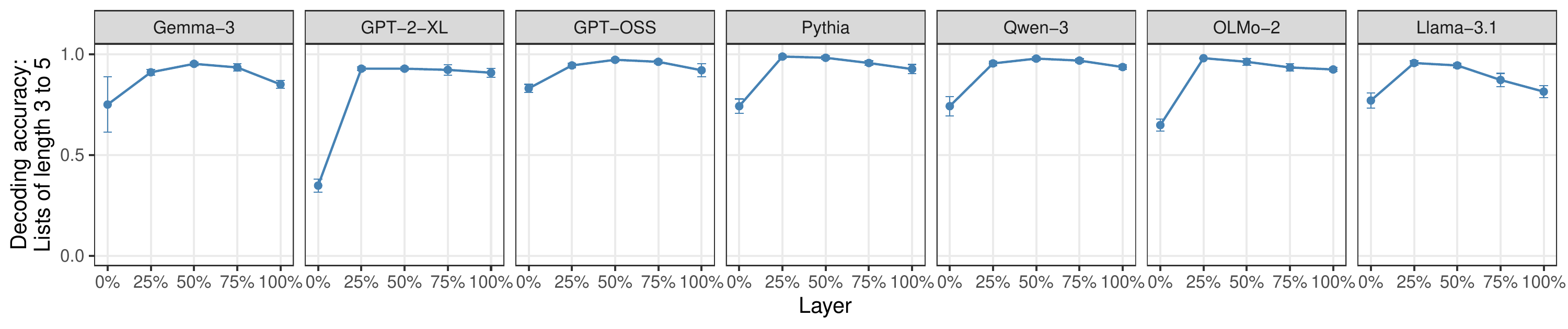}
    
    \includegraphics[width=0.8\linewidth]{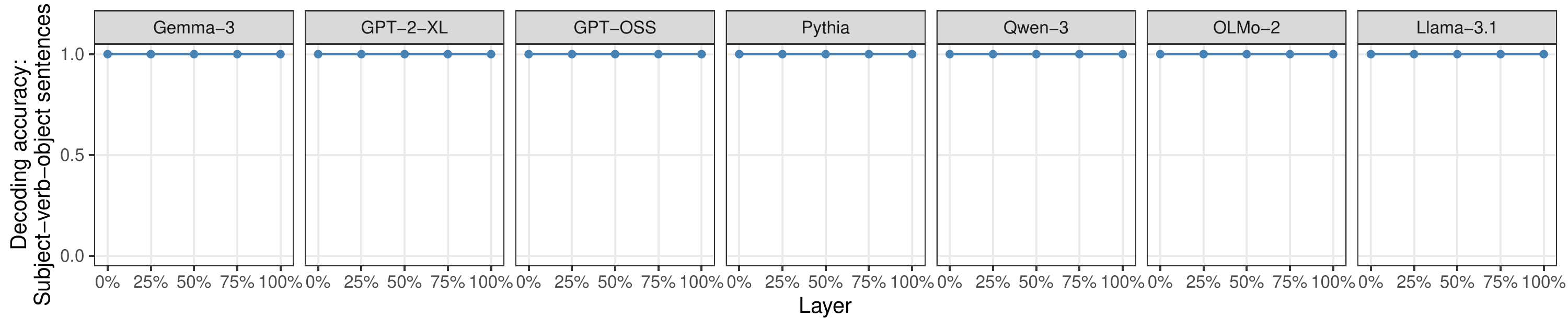}
    \includegraphics[width=0.8\linewidth]{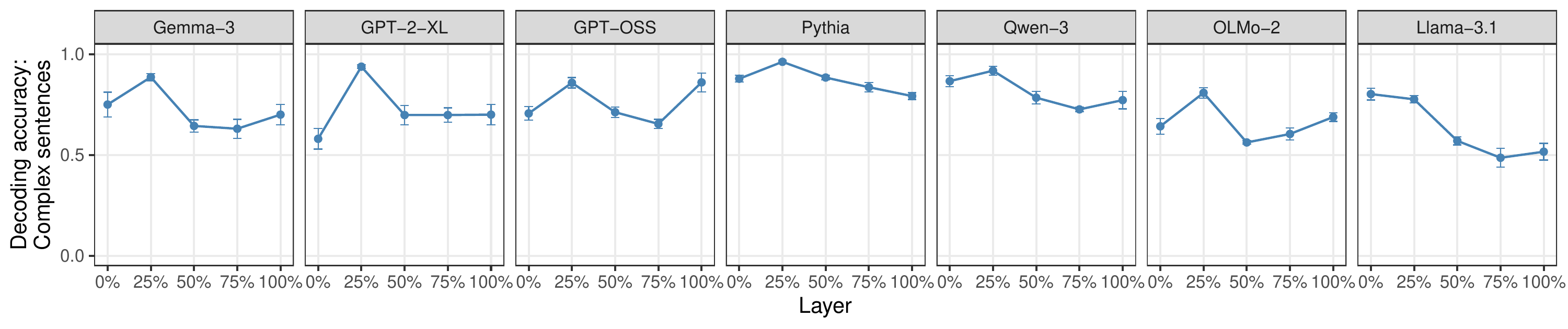}
    \caption{Accuracy of period-unpacking models trained to take in an LLM's representation of a period and reconstruct the sentence that preceded that period.}
    \label{fig:period_decoding_acc}
\end{figure}

For the lists, the period-unpacking model's target output is just the list of words (excluding the preamble and the punctuation); e.g., if the input is the period encoding from the end of \textit{Here is a list of words: lion, trouble, kindness.}, the decoder would be trained to produce \textit{lion trouble kindness}.
For the subject-verb-object sentences and complex sentences, the period-unpacking model is trained to produce the complete sentence but without capitalization and with punctuation treated as a separate token; e.g., if the input is the period encoding from the end of \textit{When the popular photographer surprised the journalist, the teacher helped the swimmer.}, the period-unpacking model's intended output would be \textit{when the popular photographer surprised the journalist , the teacher helped the swimmer .}. In all conditions, a special end-of-sequence token was appended to the end of the period-unpacking model's target output, and during evaluation it was judged to be done generating when it had produced this end-of-sequence token. 
See Figure~\ref{fig:period_decoding_acc} for the accuracies of the trained period-unpacking models.

\subsection{DISCOVER}\label{app:period_discover}

\begin{figure}[t]
    \centering
    \includegraphics[width=0.8\linewidth]{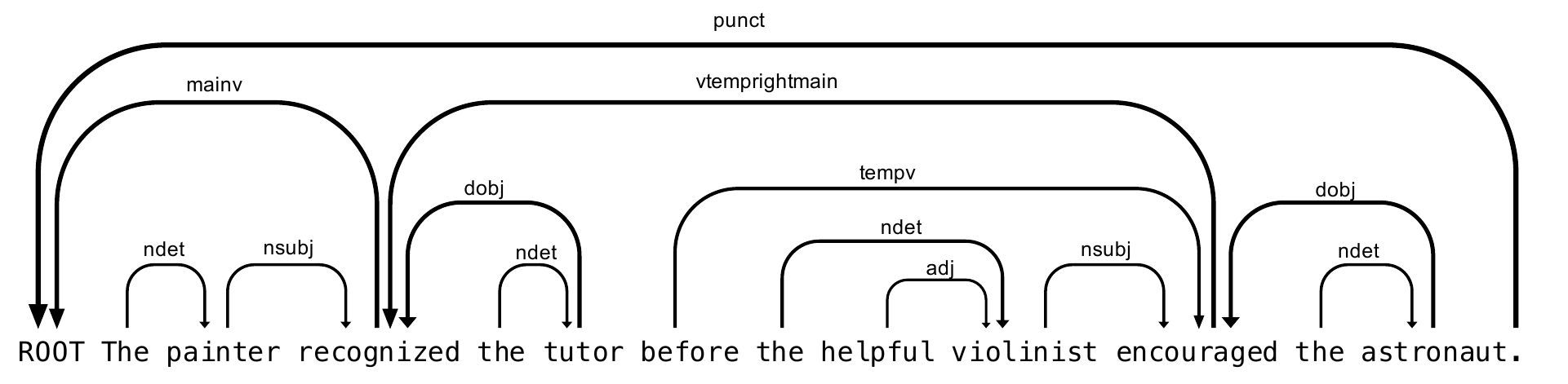}
    \caption{An example of a dependency parse used for the \texttt{syntactic} role scheme when analyzing the LLM period encodings of complex sentences.}
    \label{fig:dep_parse}
\end{figure}

Across the period encoding experiments, we use 5 role schemes:
\begin{enumerate}
    \item \texttt{Bidirectional:} Each filler's role is the concatenation of its left-to-right and right-to-left positions (e.g., (0,3) means ``1st-from-left-and-4th-from-right'').
    \item \texttt{Predecessor:} This role scheme is only used in the ``lists'' condition involving encodings of lists of words. Each filler's role is whatever filler appears immediately before it. For instance, if the list is \textit{lion, trouble, kindness}, the set of role-filler pairs would be  \textit{\{\#:lion, lion:trouble, trouble:kindness, kindness:\#\}} (using \# to indicate list boundaries).
    \item \texttt{Subject-verb-object:} This role scheme is only used for the subject-verb-object sentences. It uses three roles: \textit{subject}, \textit{verb}, and \textit{object}. E.g., the sentence \textit{The photographer called the teacher} would be analyzed with the following set of role-filler pairs: \textit{\{subject:photographer, verb:called, object:teacher\}}.
    \item \texttt{Syntactic:} This role scheme is only used with the complex sentences. In it, each filler is given a role that indicates is syntactic position, which we define as the path of dependency arc labels needed to get from the word to the root of the sentence's dependency parse tree (where we hand-define the dependency parsing algorithm that we use). Figure~\ref{fig:dep_parse} gives an example of a dependency parse; in this sentence, \textit{painter} would be given the role of \textit{nsubj-mainv}, and \textit{helpful} would be given the role of \textit{adj-nsubj-vtemprightmain-mainv}.
    \item \texttt{Bag-of-words:} All fillers are given the same role as each other, making this role scheme a degenerate one that does not capture structure.
\end{enumerate}
For the ``lists'' setting, we used a filler embedding size of 300 and a role embedding size of 100. For the ``subject-verb-object sentences'' and ``complex sentences'' settings, we used a filler embedding size of 100 and a role embedding size of 100. We trained the DISCOVER models with a mean-squared error objective on approximating the target model's representations; the DISCOVER training used the same training, validation, and test splits as the decoders that we trained. We trained the DISCOVER models with a batch size of 32 and an initial learning rate of 0.001. We evaluated them on the validation set after every 100 batches and saved their weights if the validation loss was the lowest seen so far. If the training went through 5 such evaluations with no improvement to the validation loss,  we cut the learning rate in half and reloaded the weights from the last saved checkpoint. Training was halted when the sixth such learning rate decrease would have applied.

\section{Period encodings with syntactic outputs}\label{app:period_syntax}

\begin{figure}[t]
    \centering
    \includegraphics[width=0.9\linewidth]{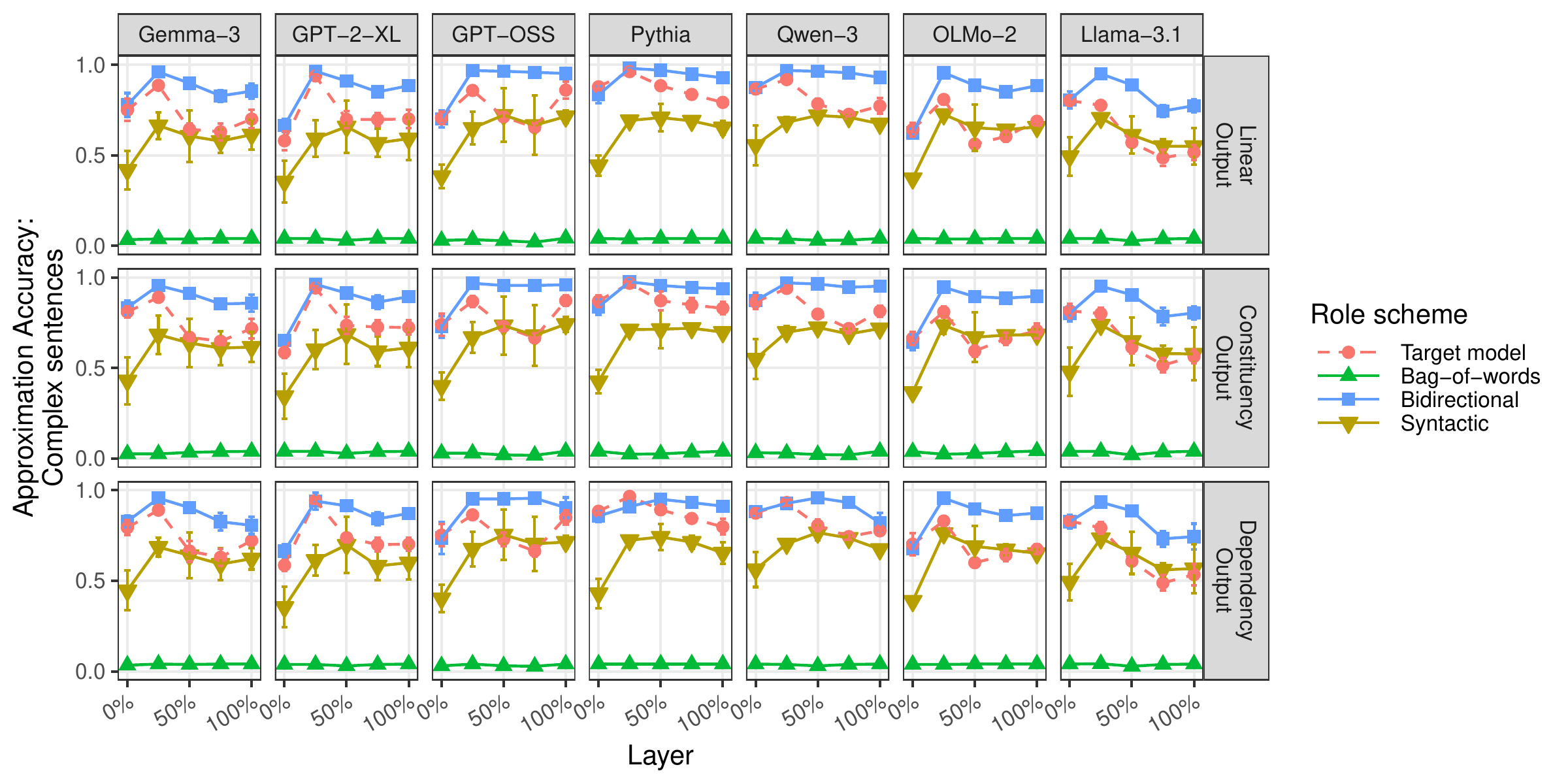}
    \caption{DISCOVER performance at approximating the encodings of periods in LLMs for the ``complex sentences'' condition, when the period-unpacking model into which the DISCOVER approximations are fed is trained to reproduce the original sentence as a linear string (top), a linearized constituency parse (middle), or a linearized dependency parse (bottom). Each point shows the mean over 5 re-runs, with error bars showing two standard deviations above and below the mean. The dashed red lines show the performance of the trained period-unpacking models when they are fed LLM period encodings; the remaining lines show the performance of the same period-unpacking models when they are fed DISCOVER approximations.}
    \label{fig:period_syntax_discover}
\end{figure}

Our DISCOVER analysis of the LLM period encodings for complex sentences found that the \texttt{bidirectional} role scheme substantially outperformed the \texttt{syntactic} role scheme. We found this surprising, as prior work has found that LLMs encode a substantial amount of syntax \citep[e.g.,][]{hewitt2019structural}. One possible explanation for the strong performance of the \texttt{bidirectional} roles is that we fed the DISCOVER approximations into a period-unpacking model that we had trained to reproduce the input sentence in linear order; perhaps if the period-unpacking model had instead been trained on a syntactic task the \texttt{syntactic} DISCOVER models would perform better. To test this hypothesis, we trained new period-unpacking models for which the target task was now to produce either a linearized constituency parse or a linearized dependency parse (we made the parses linearized so that we could still use the same decoding architecture unchanged). Example~\ref{ex:linearized_parses} gives an example of a sentence with its linearized constituency and dependency parses; the period-unpacking models trained to produce such parses were required to produce the space-delimited tokens in the parses as a linear string. The linearized constituency parse follows standard bracket notation; the linearized dependency parse follows a depth-first traversal of the labeled dependency tree.

\ex. \label{ex:linearized_parses}
\a. \textbf{Sentence:} The physicist said that the librarian recommended the important writer.
\b. \textbf{Linearized constituency parse:} [S [S2 [DP [Det the ] [N physicist ] ] [VP [V said ] [CP [Comp that ] [S2 [DP [Det the ] [N librarian ] ] [VP [V recommended ] [DP [Det the ] [N [Adj important ] [N writer ] ] ] ] ] ] ] ] [PUNCT . ] ] 
\c. \textbf{Linearized dependency parse:} mainv said nsubj physicist ndet the <end> <end> compvmainv recommended compclausev that <end> nsubj librarian ndet the <end> <end> dobj writer ndet the <end> adj important <end> <end> <end> <end> punct . <end> 

Because only the period-unpacking model changes in these parsing-based conditions, we do not need to re-train DISCOVER models; instead, we feed the same DISCOVER approximations into the original linear period-unpacking models as well as these newer syntactic period-unpacking models. The results are in Figure~\ref{fig:period_syntax_discover}; across all three decoder types, \texttt{bidirectional} continues to be the best-performing role scheme. Our conclusion is that the period encodings truly do prioritize sequential information over syntactic information; even when the period-unpacking model is trained to perform a syntactic task, it seems to mainly leverage sequential (bidirectional) information (which could be explained, e.g., by the encodings being essentially linear but then the period-unpacking model learning to do parsing). Although the period encodings do not appear to be very syntactic in nature, the possibility remains that other tokens' encodings might be, which would reconcile our findings here with prior work finding extensive syntactic information inside LLMs; indeed, in some later experiments (Section~\ref{sec:gptoss}), we will see a syntactic role scheme outperforming the linear, bidirectional one.

\section{Technical details for experiments on GPT-OSS in symbolic domains}\label{app:symbolic_technical}

\subsection{Datasets and GPT-OSS's performance}\label{app:symbolic_data}

\paragraph{Arithmetic:} Each arithmetic problem takes the form of either \texttt{a * b + c} or \texttt{a + b * c}, where \texttt{a}, \texttt{b}, and \texttt{c} are integers from -9 to 9 inclusive. The prompt shows two examples: \texttt{-9 + 8 * 7} (with the answer of \texttt{47}) and \texttt{5 * 7 + -2} (with the answer of \texttt{33}). These two examples highlight the order of operations (i.e., carry out multiplication first and then addition), to avoid any possible ambiguity regarding order of operations. The training set had 10,718 examples, the validation set had 1,000 examples, and the test set had 2,000 examples (these numbers were obtained by generating all possible expressions, allocating 1,000 to validation and 2,000 to test, and then putting the rest in traniing).

\paragraph{Syllogisms:} Two premises are provided followed by a list of 4 possible conclusions. The LLM must return the first of these 4 conclusions that follows from the premises. Every premise and conclusion takes one of the following 4 forms: \texttt{All X are Y}, \texttt{Some X are Y}, \texttt{No X are Y}, or \texttt{Some X are not Y}, where \texttt{X} and \texttt{Y} are both plural occupation nouns (we got the candidate nouns from \citet{eisape2024systematic}), optionally modified by an adjective (e.g., \texttt{painters} or \texttt{polite agents}).
We only used nouns and adjectives that were a single token in GPT-OSS's tokenizer.
The syllogisms make an existence assumption: we assume that each mentioned category has a nonzero number of members, such that \texttt{All X are Y} entails \texttt{Some X are Y} and \texttt{No X are Y} entails \texttt{Some X are not Y}. Some premise pairs entail two conclusions; e.g., if the premise pair entails that \texttt{All X are Y}, then it must also entail \texttt{Some X are Y}. To avoid having multiple possible answers, the prompt specifies that the LLM should return the \textbf{first} of the four listed conclusions that follows from the premises; we always list the conclusions in the order shown (with stronger conclusions preceding weaker conclusions that they entail) such that the target answer is the strongest conclusion that can be drawn. Our dataset included 200,000 training examples, 5,000 validation sentences, and 5,000 test sentences.

\paragraph{Code execution:} The tasks in this category require predicting the output of a Python function call. All functions in question are manipulations of Python lists. The query first provides the definitions of two functions named \texttt{change} and \texttt{alter}, whose definitions vary from query to query. There are 5 types of functions that \texttt{change} and \texttt{alter} can be: (i) reversal (reversing the input sequence), (ii) sorting (sorting the input sequence in alphabetical order), (iii) repetition (repeating the input list $n$ times, where $n$ can be 2, 3, or 4; note that $n$ is a fixed constant inside the function definition rather than being an argument of the function), (iv) suffixation (appending a particular letter at the end of the list, where the identity of the letter is fixed inside the function definition rather than being an argument), and (v) prefixation (just like suffixation but with the extra letter added at the start of the list instead of the end). After the function definitions, two variables named \texttt{x} and \texttt{y} are defined; each is a list of letters of length 1, 2, or 3. Finally, a function call is given pairing one of the two functions with one of the two lists as its argument---i.e., the call will be \texttt{change(x)}, \texttt{change(y)}, \texttt{alter(x)}, or \texttt{alter(y)}. The intended output is the result of applying this function call. Note that the two functions can appear in either order (\texttt{change} then \texttt{alter} or \texttt{alter} then \texttt{change}), and similarly for the two list variables \texttt{x} and \texttt{y}; thus, one major component of this task is variable binding (associating each variable name with its definition).  Our dataset included 200,000 training examples, 1,000 validation sentences, and 1,000 test sentences.

\paragraph{Passivization:} The LLM must convert an active English sentence into its passive form. The input sentences were generated from a context-free grammar in which sentences always took the basic form of subject-verb-object, but there was substantial variability in the syntactic structure that the subject and object could have: they could be just \textit{the} plus a noun (e.g., \textit{the doctor}), or they could optionally include adjectives, prepositional phrases, and relative clauses (e.g., \textit{the illustrator who surprised the gardener next to the tall editor}). The example that is provided in the prompt includes a relative clause in order to make clear that only the main clause of the sentence should be passivized, not any relative clauses. Our dataset included 200,000 training examples, 5,000 validation sentences, and 5,000 test sentences. We ensured that all words used in the input sentences were processed as a single token in GPT-OSS's tokenizer. Within each sentence, we did not allow any content word (noun, verb, or adjective) to be repeated.

\paragraph{Tense reinflection:} The LLM must convert a past-tense English sentence into the present tense. Doing so requires the LLM to identify the subject of each verb in order to know whether the verb should take its singular or plural form (since the English past tense is not inflected for number while the English present tense is). The sentences take the same basic structures as those used in the passivization task. There is one wrinkle: The interplay of relative clauses and prepositional phrases can potentially create a scenario where syntactic ambiguity makes it unclear whether the verb should be singular or plural. E.g., in \texttt{The detective by the astronauts who helped the doctor recommended the lawyer}, it is possible for the subject of \texttt{helped} to be either \texttt{detective}---in which case the present-tense form should be \texttt{helps}---or \texttt{astronauts}---in which case the present-tense form should be \texttt{help}. To address this concern, we add the constraint that the object of a prepositional phrase must have the same number as the noun preceding the prepositional phrase; this does not remove the syntactic ambiguity, but it does ensure that there is no ambiguity about which linear string of words should be produced as the answer. Our dataset included 200,000 training examples, 5,000 validation sentences, and 5,000 test sentences. We ensured that all words used in the input sentences were processed as a single token in GPT-OSS's tokenizer. Within each sentence, we did not allow any content word (noun, verb, or adjective) to be repeated.

\paragraph{Question formation:} The LLM must convert a declarative English sentence into a yes/no question. The input sentences follow the same basic structure as those used for passivization and tense reinflection except that every verb is preceded by an auxiliary verb such as \texttt{would} or \texttt{can}.  Our dataset included 200,000 training examples, 5,000 validation sentences, and 5,000 test sentences. We ensured that all words used in the input sentences were processed as a single token in GPT-OSS's tokenizer. Within each sentence, we did not allow any content word (noun, verb, or adjective) to be repeated.

\paragraph{CFGs for syntax tasks:} For the three syntax tasks, sentences were generated from context-free grammars (CFGs). The CFGs can be found in the supplementary file on the project GitHub (Supplement S1).\footnote{\url{https://github.com/tommccoy1/discover/blob/main/discover_supplement.pdf}}

\paragraph{Harmony prompt format:} The prompts fed to GPT-OSS followed the Harmony prompt format expected by GPT-OSS (\url{https://github.com/openai/harmony}). That is, special tokens were included to demarcate components of the prompt. An example of a complete arithmetic prompt is in Example~\ref{ex:harmony}. It begins with two in-context examples, whose inputs are in the \texttt{user} field and whose answers are in the \texttt{assistant} field. It then includes the query that the LLM is expected to answer (\texttt{4 + -9 * 9}), and it should respond by producing tokens to continue after \texttt{final}---specifically, it should produce the special \texttt{<|message|>} token followed by its response.

\ex. \texttt{<|start|>user<|message|>-9 + 8 * 7<|end|><|start|>assistant<|channel|>final<|message|>\allowbreak 47\linebreak<|end|><|start|>user<|message|>5 * 7 + -2<|end|><|start|>assistant<|channel|>final\linebreak<|message|>33<|end|><|start|>user\allowbreak <|message|>\allowbreak 4 + -9 * 9<|end|><|start|>assistant<|channel|>\linebreak final}\label{ex:harmony}

\begin{figure}
    \centering
    \includegraphics[width=0.35\linewidth]{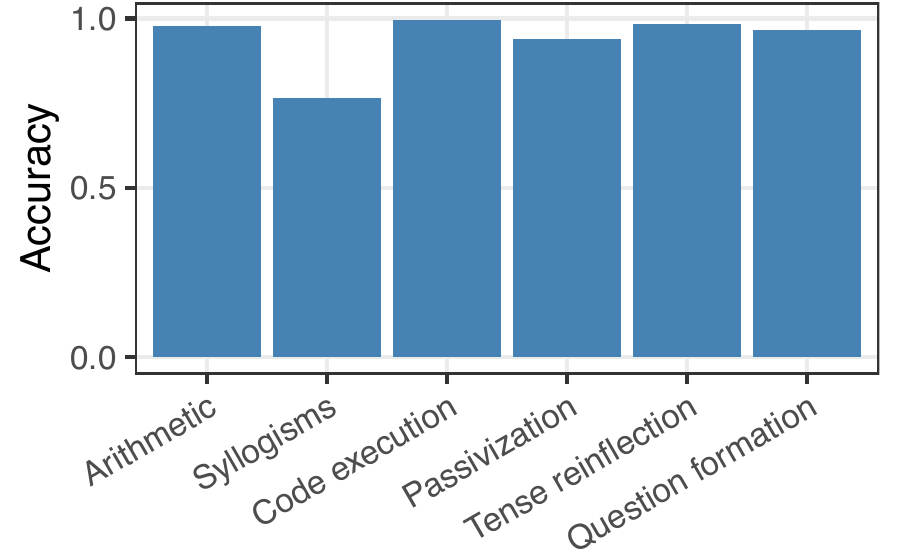}
    \caption{Performance of GPT-OSS on the six symbolic tasks that we consider.}
    \label{fig:llm_target_acc}
\end{figure}

\paragraph{GPT-OSS performance on the target tasks:}
Figure~\ref{fig:llm_target_acc} gives GPT-OSS's performance on the 6 tasks that we use in these experiments.

\subsection{Role schemes}\label{app:gpt_oss_role_schemes}

For our analyses of GPT-OSS performing symbolic tasks, we consider 5 types of role schemes: \texttt{bag-of-words}, \texttt{bidirectional (self)}, \texttt{bidirectional (all)}, \texttt{task-specific (self)}, and \texttt{task-specific (all)}. Table~\ref{tab:gpt_oss_roles} illustrates how these role schemes apply to one input for the passivization task. The \texttt{bag-of-words}, \texttt{bidirectional (self)}, and \texttt{bidirectional (all)} role schemes are all task-agnostic, so all of them apply in the same way to other tasks as they do for the passivization task. 
However, the task-specific ones are set up in different tasks. As noted in the main text, \texttt{task-specific (self)} is defined via a task-specific parsing algorithm that assigns each token a role that denotes its position in the context of that task (where the authors used their knowledge of these domains to hypothesize what the relevant notion of task-specific position is). In the \texttt{task-specific (all)} role scheme, each role is then the concatenation of two roles from \texttt{task-specific (self)}. Here are descriptions of the task-specific parsing algorithms used to define these role schemes for each task:
\begin{itemize}
    \item \textbf{Arithmetic:} The positions in arithmetic expressions indicate whether the token is a digit, a negative sign (the negative sign is always a separate token from the digit), or an operator (i.e., \texttt{+} or \texttt{*}); they also indicate the position of that element and the sequence of operators that have been seen so far, and for digits the role indicates whether the digit is negated or not. Thus, some example positions would be ``the second digit, which is negated, in an expression that thus far contains \texttt{+} and no other operators'', ``the non-negated first digit of an expression that thus far contains no operators'', or ``the second operator in an expression that, up until now, contained only \texttt{*} and no other operators.'' The operators are included in the roles because, due to the order of operations, they affect the structure of the computation to be performed.
    \item \textbf{Syllogisms:} The prompt is framed as containing 6 sentences (the 2 premises and the 4 candidate conclusions). Each token's role denotes which sentence that token appears in and its position in that sentence (such as subject noun, predicate noun, adjective modifying the predicate noun, quantifier, etc.) Some example roles would thus be ``the quantifier in the third candidate conclusion'' or ``the adjective modifying the predicate noun in the second premise.'' There are also roles corresponding to the linear positions of the tokens in the framing text that is not part of the premises and conclusions (\textit{Please return the first conclusion listed below that follows from these premises:}). 
    \item \textbf{Code execution:} For each token inside either of the two function definitions, the role for that token indicates the name of the function (\texttt{alter} or \texttt{change}), whether it is the first or second function, and what linear position inside the function definition is occupied by this token. For instance, one such role would be ``third token inside the definition of \texttt{alter} appearing as the second function in the prompt.'' The variable definitions also have roles indicating which variable the token appears inside, whether that variable is the first or second to be defined, and what position inside that variable definition this token occupies. For instance, one such role would be ``second element of the \texttt{y} list, which is the first list variable to be defined.'' Finally, there are roles for the elements of the function call (i.e., a role for ``function being called'' and a role for ``variable inside function call''), as well as roles for all the formatting material that appears in and around the components mentioned so far.
    \item \textbf{Syntactic tasks:} In all three syntactic tasks (passivization, tense reinflection, and question formation), each token's position is its syntactic position, such as ``subject noun'' or ``adjective modifying the object of the prepositional phrase that modifies the subject noun.''
\end{itemize}

\subsection{DISCOVER}\label{app:gpt_oss_discover}

We train DISCOVER to approximate all token representations except for those that are identical across prompts. Consider Example~\ref{ex:passive} below, which shows a complete input for the passivization task. Much of this input is invariant across stimuli: the in-context example \textit{The swimmer who the polite politician advised avoided the accountant near the clever gardener} is held constant across examples, as is the instruction \textit{Passivize the following sentence:}. The stimuli only become different once we get to the specific sentence to be passivized (here, \textit{The illustrator who surprised the gardener next to the tall editor helped the author in front of the famous drummer.}). Further, GPT-OSS proceeds in a left-to-right way, so the representation for a given token is only influenced by itself and the preceding tokens. As a result, all of the tokens that appear in the initial part of the prompt that is fixed across stimuli are guaranteed to have the same representation across stimuli, meaning that approximating them with DISCOVER would not be very interesting. Therefore, we only have DISCOVER approximate the tokens whose representations vary across stimuli---e.g., in this example, the tokens in the segment \textit{The illustrator who surprised the gardener next to the tall editor helped the author in front of the famous drummer.<|end|><|start|>assistant<|channel|>final}. Note that this includes formatting tokens at the end; although these tokens are fixed across stimuli, they are preceded by tokens that vary from stimulus to stimulus, meaning that their representations can be different in different stimuli---and we want to make sure that DISCOVER can capture this variation in case it is critical to how GPT-OSS operates for these stimuli.

\ex. <|start|>user<|message|>Passivize the following sentence: The swimmer who the polite politician advised avoided the accountant near the clever gardener.<|end|><|start|>assistant<|channel|>final<|message|>The accountant near the clever gardener was avoided by the swimmer who the polite politician advised.<|end|>\allowbreak <|start|>user<|message|>Passivize the following sentence: The poet who surprised the gardener next to the tall editor helped the author in front of the famous drummer.<|end|><|start|>assistant<|channel|>\allowbreak final \label{ex:passive}

We train one DISCOVER model for each of GPT-OSS's 25 layers. That is, we use the same learned DISCOVER parameters across tokens within a layer, but we use different learned DISCOVER parameters across layers.
For a given token, we hypothesize the same role-filler pairs for all representations of that token across layers (e.g., the role-filler pairs for \textit{poet} in the second-to-last row of Table~\ref{tab:gpt_oss_roles} would be used in the DISCOVER model for each of the 25 layers as the hypothesis underlying the DISCOVER approximation for the representation of \textit{poet})---though the learned DISCOVER parameters instantiating these hypotheses (e.g., the filler embedding for \textit{poet}) can vary across layers. It is likely that different layers in fact differ in what information they encode \citep[e.g.,][]{tenney2019bert}, but we leave investigation of such variation for future work, as the assumption of uniformity---flawed though it might be---produces effective approximations in our experiments. 
In relation to this point, it is worth recalling the ``third'' point from Section~\ref{sec:discover_comments}, which is that DISCOVER can succeed as long as the information in the DISCOVER hypothesis is a superset of what is actually encoded in the target model---it does not need to precisely match what is in the target model. Thus, the fact that we achieve successful approximations when adopting the same structural hypothesis across layers does not necessarily mean that all layers encode the same information; they might encode differing subsets of the information present in that hypothesis.

We trained our DISCOVER models with a batch size of 24 for arithmetic, passivization, tense reinflection, and question formation, and we used a batch size of 12 for syllogisms and code execution (the last two required smaller batch sizes because their prompts are longer, meaning that fewer prompts could fit on our GPUs at once). In all cases, we used a filler embedding size of 100, a role embedding size of 50, and an initial learning rate of 0.0001. We evaluated the DISCOVER models on the validation set after every 200 batches and saved the weights if they achieved the lowest validation loss seen so far. If we went through 20 such evaluations with no improvement to the validation loss, we halved the learning rate and reloaded the weights from the last saved checkpoint; we halted training at the point when the eleventh such learning rate reduction would have occurred.

\section{Technical details for causal interventions}

\subsection{Number of intervention examples}

For the letter sequence models and for the LLM period encodings, we used a dataset of 1,000 interventions.
For the GPT-OSS experiments, we used a dataset of 100 interventions for each condition, because the interventions were more computationally costly to carry out in this setting than in the others.

\subsection{LLM period encodings}\label{app:period_encoding_interventions}

To perform causal interventions on lists, we used only lists of length 5. To perform causal interventions on sentences, we used only sentences of the form \textit{The NOUN VERBed the NOUN, and the NOUN VERBed the NOUN.} 
We used these fixed templates because it made defining the interventions simpler than it would be if we allowed more variable input structures. 
For each of these two settings, we trained new period decoders on only stimuli that came from these fixed templates (length-5 lists or two-clause sentences), using the same procedure described in Appendix~\ref{app:training_period_decoders}. We then trained DISCOVER models on the encodings of the same restricted sets of stimuli. For the lists of length 5, we used the same setup as for the lists described in Appendix~\ref{app:period_discover}, except that the only role scheme we considered was \texttt{bidirectional}.
For the sentences, we used the same setup as for the subject-verb-object sentences described in Appendix~\ref{app:period_discover}, except that the only role scheme we considered was the two-clause analogue of the \texttt{subject-verb-object} role scheme (i.e., it had six roles: \textit{subject of first sentence}, \textit{verb of first sentence}, \textit{object of first sentence}, \textit{subject of second sentence}, \textit{verb of second sentence}, and \textit{object of second sentence}).
Those DISCOVER models were then the ones that we used to perform the causal interventions, which replaced one randomly-selected word (in the list condition) or one randomly-selected subject or object (in the sentence condition).

\subsection{GPT-OSS in symbolic domains}\label{app:gpt_oss_causal_interventions_details}

To perform causal interventions on GPT-OSS's representations in symbolic domains, we used the \texttt{task-specific (all)} role scheme for the domain in question. Whenever replacing a role-filler pair, we edited all tokens that would need to be modified by this change. For example, Table~\ref{tab:example_edits} shows how we would change \texttt{spy} (in position \texttt{subj\_n}) to \texttt{astronaut} (which would also have the role \texttt{subj\_n}). Under the \texttt{task-specific} role scheme, all tokens in the sentence from the position of \texttt{spy} onward are hypothesized to contain an encoding of \texttt{spy}. Thus, to edit \texttt{spy}, we will need to edit the representation of each of those tokens by subtracting out the role-filler pair that includes \texttt{spy} and adding back in a role-filler pair that instead contains \texttt{astronaut}. Note that the role to which \texttt{spy} (and later \texttt{astronaut}) is bound is different for each token position because the roles are the concatenation of the position of \texttt{spy} (which is always the same: \texttt{subj\_n}) and the current token (which varies).

\begin{table}[t]
    \centering
    \small
    \begin{tabular}{cp{3.8cm}p{4cm}} \toprule
         & Original & Edit to make \\ \midrule
        The & \{subj\_det-subj\_det:The\} & None \\ \midrule
        spy &  \{subj\_det-subj\_n:The, subj\_n-subj\_n:spy\} & $-$ subj\_n-subj\_n:spy \newline $+$ subj\_n-subj\_n:astronaut \\ \midrule
        helped & \{subj\_det-verb:The, subj\_n-verb:spy, \newline verb-verb:helped\} & $-$ subj\_n-verb:spy \newline $+$ subj\_n-verb:astronaut \\ \midrule
        the & \{subj\_det-obj\_det:The, subj\_n-obj\_det:spy, verb-obj\_det:helped, obj\_det-obj\_det:the\} & $-$ subj\_n-obj\_det:spy \newline $+$ subj\_n-obj\_det:astronaut \\ \midrule
        poet &  \{subj\_det-obj\_n:The, subj\_n-obj\_n:spy, \newline verb-obj\_n:helped, obj\_det-obj\_n:the, obj\_n-obj\_n:poet\} & $-$ subj\_n-obj\_n:spy \newline $+$ subj\_n-obj\_n:astronaut \\ \midrule
        . & \{subj\_det-punct:The, subj\_n-punct:spy, \newline verb-punct:helped, obj\_det-punct:the, \newline obj\_n-punct:poet, \newline punct-punct:.\}  & $-$ subj\_n-punct:spy \newline $+$ subj\_n-punct:astronaut \\ \bottomrule
    \end{tabular}
    \caption{Representational edits that would be made with the goal of changing \textit{spy} to \textit{astronaut}.}
    \label{tab:example_edits}
\end{table}

For the sake of space, additional information about the causal interventions carried out on GPT-OSS is presented in the supplementary file on the project GitHub page (in Supplement S2).\footnote{\url{https://github.com/tommccoy1/discover/blob/main/discover_supplement.pdf}} This additional information includes further information about how role-changing interventions are carried out as well as detailed examples and accuracy metrics for all of the categories of interventions that we use.

\section{Testing for structure-sensitivity of causal interventions in GPT-OSS}\label{app:intervention_structure_sensitive}

\begin{figure}[t]
    \centering
    \includegraphics[width=0.9\linewidth]{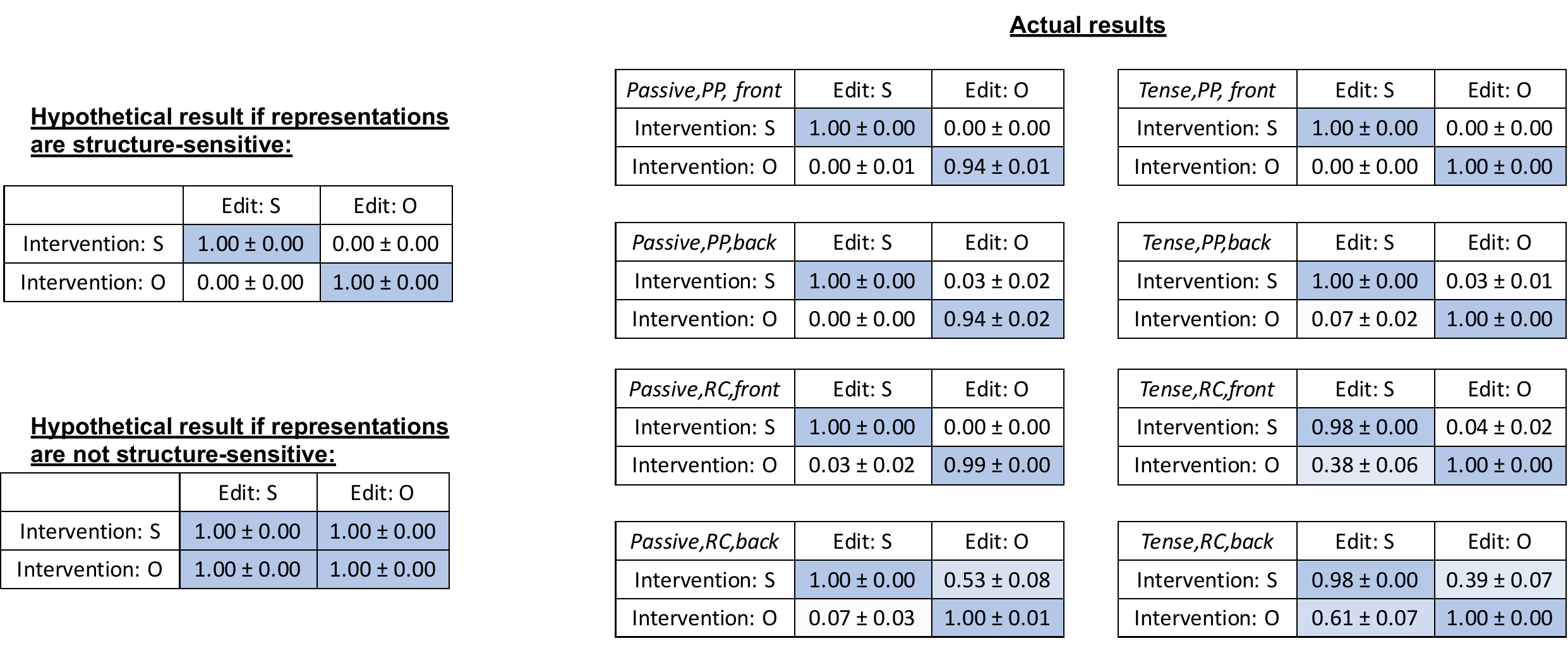}
    \caption{Analysis of whether causal interventions on GPT-OSS representations are sensitive to sentence structure or are purely driven by linear order. See the text for descriptions of the terms in the plots. All results (right) show at least some version of the diagonal pattern predicted if the interventions are structure-sensitive (top left) rather than the uniform pattern predicted if the interventions are not structure-sensitive (bottom left).}
    \label{fig:intervention_structure}
\end{figure}

A possible concern with our causal interventions on GPT-OSS is that they might only be intervening on the linear positions of tokens rather than on more abstract structural positions. One reason to have this concern is that a linear role scheme (\texttt{bidirectional (all)}) gave an approximation that was nearly as strong as \texttt{task-specific (all)} for GPT-OSS, particularly for the linguistic tasks (Figure~\ref{fig:llm_tpe_acc}; thus, if much of the structure can be captured in this linear way, perhaps the interventions succeed by editing only linear information. While it would still be interesting to demonstrate fine-grained control of linear information in an LLM, it is less interesting than demonstrating fine-grained control of more abstract structure. 
To address this concern, here we run an analysis that tests whether our intervention results can be explained via purely linear edits.

It is easiest to illustrate this analysis through an example. Consider the following two interventions for the tense reinflection task, both of which move the adjective \textit{popular} to a position later in the sentence:

\ex. \textbf{Intervention 1:}\label{ex:structure1}
\a. \textit{Output pre-intervention:} The dancer behind the \underline{popular} tutor recognizes the librarian.\label{ex:structure1a}
\b. \textit{Output post-intervention:} The dancer behind the tutor recognizes the \underline{popular} librarian.

\ex. \textbf{Intervention 2:}\label{ex:structure2}
\a. \textit{Output pre-intervention:} The dancer recognizes the \underline{popular} tutor behind the librarian.\label{ex:structure2a}
\b. \textit{Output post-intervention:} The dancer recognizes the tutor behind the \underline{popular} librarian.

\noindent
If these interventions are conducted in a way based on linear position, then both of them are making the same change: they are moving \textit{popular} from the linearly fifth position to the linearly eighth position. 
However, if these interventions are instead conducted in a way that is based on sentence structure, then they are making different changes: Intervention 1 is moving \textit{popular} from inside the prepositional phrase modifying the subject to the direct object, while Intervention 2 is moving \textit{popular} from the direct object to inside the prepositional phrase modifying the direct object.

Such intervention pairs---where the two interventions make the same change in linear terms but different changes in structural terms---form the basis of our analysis. Each intervention is conducted by adding some vectors into GPT-OSS's representations, where the vectors are derived from our trained DISCOVER model. 
The vectors that we generate are intended to capture sentence structure: for Intervention 1 we generate a vector that captures the shift from subject prepositional phrase to direct object (call this Edit Vector 1), and for Intervention 2 we generate a vector that captures the shift from direction object to object prepositional phrase (call this Edit Vector 2). 
Typically the way our interventions work is that we edit a representation using the edit vector that was motivated by that vector; i.e., we would edit Representation 1 (the representation of sentence~\ref{ex:structure1a}) using Edit Vector 1, and we would edit Representation 2 (the representation of sentence~\ref{ex:structure2a}) using Edit Vector 2. However, if the interventions are only sensitive to linear order, these edit vectors should be interchangeable since they instantiate the same linear movement; whereas if they are sensitive to sentence structure they should not be interchangeable.

The structure of our analysis, then, is to consider all four possible pairings of a starting representation (Representation 1 or Representation 2) and an edit vector (Edit Vector 1 or Edit Vector 2). If the interventions are structure-sensitive, each edit vector should only be effective on the representation that matches with it (Figure~\ref{fig:intervention_structure}, top left). In contrast, if the interventions are only sensitive to linear order, either edit vector should be effective with either input representation (Figure~\ref{fig:intervention_structure}, bottom left).

We consider four types of intervention pairs, shown below. To explain the terminology: The \textbf{PP} interventions are ones based on prepositional phrases, while the \textbf{RC} ones are based on relative clauses. The \textbf{front} interventions involve moving an adjective forward in the sentence, while the \textbf{back} interventions involve moving an adjective backward. The \textbf{S} interventions are ones in which the subject is modified by a prepositional phrase or relative clause, and the \textbf{O} interventions are ones in which the object is modified. We consider these interventions for both the passivization task (\textbf{passive}) and the tense reinflection task (\textbf{tense}):

\ex. \textbf{PP, front}
\a. \textbf{Intervention: S}
\a. \textit{Output pre-intervention:} The dancer behind the \underline{popular} tutor recognizes the librarian.
\b. \textit{Output post-intervention:} The dancer behind the tutor recognizes the \underline{popular} librarian.
\z.
\b. \textbf{Intervention: O}
\a. \textit{Output pre-intervention:} The dancer recognizes the \underline{popular} tutor behind the librarian.
\b. \textit{Output post-intervention:} The dancer recognizes the tutor behind the \underline{popular} librarian.

\ex. \textbf{PP, back}
\a. \textbf{Intervention: S}
\a. \textit{Output pre-intervention:} The dancer behind the tutor recognizes the \underline{popular} librarian.
\b. \textit{Output post-intervention:} The dancer behind the \underline{popular} tutor recognizes the librarian.
\z.
\b. \textbf{Intervention: O}
\a. \textit{Output pre-intervention:} The dancer recognizes the tutor behind the \underline{popular} librarian.
\b. \textit{Output post-intervention:} The dancer recognizes the \underline{popular} tutor behind the librarian.

\ex. \textbf{RC, front}
\a. \textbf{Intervention: S}
\a. \textit{Output pre-intervention:} The miner who the \underline{careful} professor stops advises the ambassador.
\b. \textit{Output post-intervention:}  The miner who the professor stops advises the \underline{careful} ambassador. 
\z.
\b. \textbf{Intervention: O}
\a. \textit{Output pre-intervention:} The miner stops the \underline{careful} professor who advises the ambassador.
\b. \textit{Output post-intervention:}  The miner stops the professor who advises the \underline{careful} ambassador.

\ex. \textbf{RC, back}
\a. \textbf{Intervention: S}
\a. \textit{Output pre-intervention:}  The miner who the professor stops advises the \underline{careful} ambassador. 
\b. \textit{Output post-intervention:} The miner who the \underline{careful} professor stops advises the ambassador.
\z.
\b. \textbf{Intervention: O}
\a. \textit{Output pre-intervention:}  The miner stops the professor who advises the \underline{careful} ambassador. 
\b. \textit{Output post-intervention:} The miner stops the \underline{careful} professor who advises the ambassador.

\noindent
The results are in Figure~\ref{fig:intervention_structure} (right). In all 8 cases, the results show at least some version of the diagonal pattern predicted by structure-sensitivity rather than the uniform pattern predicted by non-structure-sensitivity. In some cases, the diagonal pattern is extremely clear (e.g., \textit{tense,PP,front}). In others, it is not as stark, but it is still the case that the mismatched interventions result in substantially lower accuracy than the matched interventions (e.g., \textit{tense,RC,back}). We therefore conclude that these interventions are sensitive to sentence structure rather than being driven purely by linear position.

\section{Technical details for experiments testing for generalization to novel role-filler pairs}\label{app:ood_details}

\subsection{General note: Restricted templates}

In most of our main DISCOVER experiments, we train DISCOVER on stimuli that vary in structure because doing so is necessary for teasing apart candidate role schemes. 
For instance, if all inputs are sentences of the form \textit{The SUBJECT VERBED the OBJECT}, then a role scheme based on linear position would be equivalent to a role scheme based on syntactic positions because there would be a one-to-one mapping between linear positions and syntactic positions (e.g., the linearly-second word is always the subject). If the sentence structure is more varied, however, the different role schemes become deconfounded, enabling us to judge which one fits the target model best.

For the purpose of testing for generalization to novel role-filler pairs, however, experimental considerations point in the opposite direction, toward using fixed templates. Specifically, running these experiments requires us to withhold certain role-filler pairs during DISCOVER training. It is plausible that neural networks might sometimes use a fuzzy combination of multiple role schemes (e.g., their representations might be partially linear and partially syntactic). 
Such entanglement makes the withholding that we need to do somewhat tricky: Suppose we are withholding the filler \texttt{doctor} from syntactic role \texttt{subject} and then testing how DISCOVER generalizes to stimuli containing the pair \texttt{subject:doctor}. In our training set, the \texttt{subject} syntactic slot might overlap with a different syntactic slot (e.g., \texttt{object of a prepositional phrase following the subject}) with respect to the linear positions it can occupy. If the target model's representations are partially governed by linear position, then this shared linear position could result in the representations of these two syntactic roles being similar. As a result, even though we intend to withhold \texttt{subject:doctor}, we might not be doing so effectively because allowing \texttt{object of a prepositional phrase following the subject:doctor} to appear might have the effect of providing some leaky signal about \texttt{subject:doctor}. This possibility can be guarded against by using strict templates for the DISCOVER training stimuli such that, for all reasonable hypotheses we can make about what role scheme might be used by the target model, all would be in effect the same; in that case, withholding a role-filler pair as defined in one role scheme (e.g., \texttt{subject:doctor}) is guaranteed to also withhold a role-filler pair as defined in a different role scheme (e.g., \texttt{linearly-second:doctor}), ensuring that our evaluation is testing generalization to a withheld role-filler pair as intended.

Motivated by these considerations, the tests described below generally use rigid templates for training stimuli except in the case of letter sequence models, which are simple enough for us to be reasonably confident that the bidirectional withholding described below will address the same sorts of confounds that we otherwise address with strict templates.

\subsection{Letter sequence models}

For letter sequence models, we trained 10 re-runs for each setting, using the \texttt{bidirectional} role scheme since that role scheme reliably gave a strong DISCOVER approximation across tasks. We used the same training procedure as for the letter sequence DISCOVER models discussed earlier (Appendix~\ref{app:letter_sequence_discover_training}).

When training DISCOVER, we withheld the letter \texttt{A} from appearing in the position first from left, the letter \texttt{B} from appearing in the position second from left, etc., up to withholding \texttt{F} from appearing in the position sixth from left. We also withheld \texttt{A} from appearing as the sixth-from-right letter, \texttt{B} from appearing as the fifth-from-right letter, etc., up to withholding \texttt{F} from appearing as the rightmost letter.
We generated evaluation sets that contained some of these withheld role-filler pairs---i.e., out-of-distribution (OOD) evaluation sets. 
All elements in the evaluation sets were of length 6, such that each withheld role-filler pair appearing in the sequences had been withheld from both the left-to-right and right-to-left position that it appears in. 
We used the withheld role-filler pairs involving \texttt{B} and \texttt{E} for OOD validation for selecting regularization hyperparameters (Appendix~\ref{app:l21}), and we used the other 4 withheld role-filler pairs for OOD testing. Since we had 4 withheld role-filler pairs available for OOD testing, we made 4 OOD test splits containing sequences with 1, 2, 3, or 4 withheld role-filler pairs, respectively.

\subsection{Period encodings}

For the period encoding experiments, we used lists of length 5 and sentences of the form \textit{The NOUN VERBED the NOUN, and the NOUN VERBED the NOUN.} 

To create the list data, for each of the 5 possible list positions, we designated 10 nouns as withheld for purposes of OOD testing and a distinct set of 10 nouns as withheld for purposes of OOD validation. No nouns were withheld from more than one position. We then trained DISCOVER on a training set that had none of these withheld word-position pairs and used examples containing the role-filler pairs that had been withheld for validation for purposes of selecting the regularization hyperparameter (Appendix~\ref{app:l21}). We then tested the DISCOVER models on 5 OOD test splits, which contained examples with 1, 2, 3, 4, or 5 withheld role-filler pairs, respectively.
We used a role scheme of \texttt{bidirectional}.

To create the sentence data, for each of the four possible noun positions, we designated 10 nouns as withheld from appearing in that position. No nouns were withheld from more than one position. We did not have an OOD validation set; we used the lists condition for selecting a regularization weight and then used the same weight for the sentences, under the assumption that what works in one setting should work for the other given that they use the same types of encodings with the same type of normalization (z-scoring). We tested the DISCOVER models on 4 OOD test splits, which contained examples with 1, 2, 3, or 4 withheld role-filler pairs, respectively. 
In all OOD evaluations, we constrained both verbs in the sentence to be the same as each other.
For our DISCOVER approximations, we used a role scheme with roles for \textit{subject of first sentence}, \textit{verb of first sentence}, \textit{object of first sentence}, \textit{subject of second sentence}, \textit{verb of second sentence}, and \textit{object of second sentence}.

\begin{figure}
    \centering
    \includegraphics[width=0.9\linewidth]{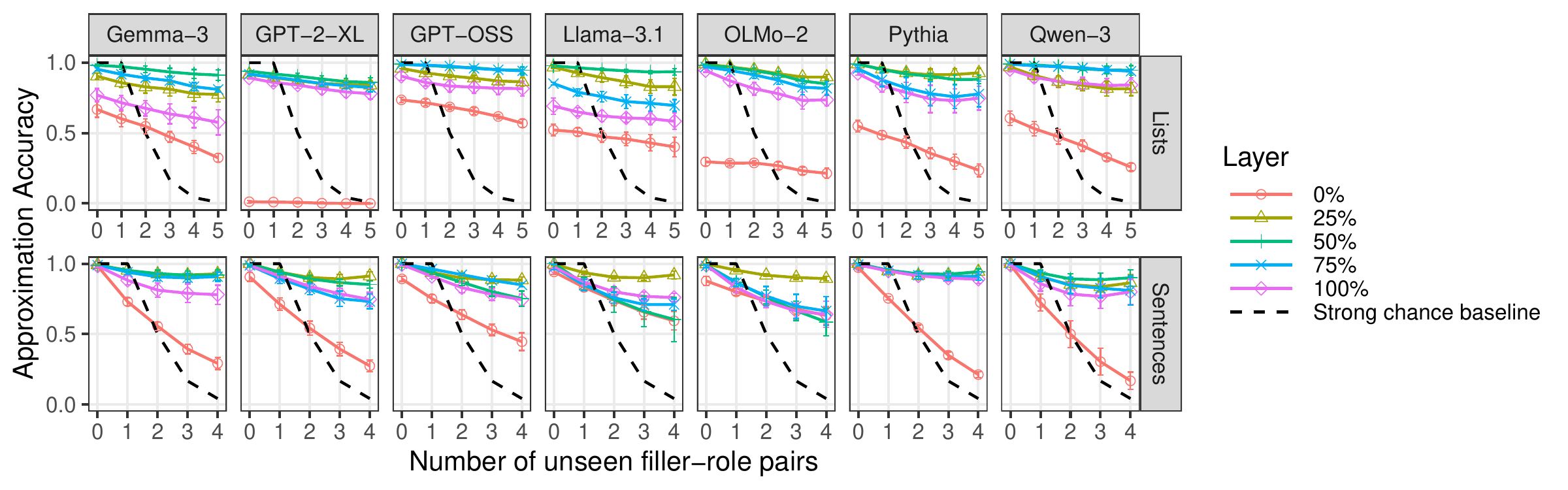}
    \caption{Testing how DISCOVER generalizes to novel role-filler pairs when it is trained to approximate LLM period encodings after lists of length 5 (top) or sentences of the form ``the NOUN VERBED the NOUN, and the NOUN VERBED the NOUN.'' (bottom). The dots show the mean across 10 reruns, while the error bars show two standard deviations (truncated at 0.0 and 1.0). The dashed lines show a strong baseline.}
    \label{fig:tpe_acc_ood_period_full}
\end{figure}

For both the lists and sentences, we used the same procedures to train DISCOVER models as described in Appendix~\ref{app:period_encoding_interventions}.
Full results for all 5 LLM layers that we considered are in Figure~\ref{fig:tpe_acc_ood_period_full}. All layers show robust generalization that is substantially above the chance baseline, except that in some cases the first layer (0\%) performs at or below baseline level.

\subsection{GPT-OSS performing symbolic tasks}

For arithmetic, we withheld the digits \texttt{1}, \texttt{2}, and \texttt{3} from appearing as the first number in the expression, \texttt{4}, \texttt{5}, and \texttt{6} from appearing as the second number in the expression, and \texttt{7}, \texttt{8}, and \texttt{9} from appearing as the third number in the expression. The positive version and the negative version of each digit in question were both withheld from the specified position.
The withheld role-filler pairs involving \texttt{3}, \texttt{6}, and \texttt{9} were used for OOD validation, and the remaining withheld pairs were used for OOD testing. 

For syllogisms, we restricted ourselves to only a subset of the syllogism templates to guarantee that (i) all templates had the same number of words and (ii) the nouns in the template had the same relative ordering (i.e., a syllogism always has one noun that is repeated across both premises and two others that are unique; we selected only templates of the form ``QUANTIFIER1 NOUN1 are NOUN2. QUANTIFIER2 NOUN3 are NOUN1.'') These choices were made to ensure that the positions we withheld were consistent linear positions, so that each instance of withholding withheld the noun in question both from its structural position and from its bidirectional position. For each noun slot (NOUN2, NOUN3, and the repeated NOUN1), we designated 5 nouns as withheld for OOD validation and 5 as withheld for OOD testing.

For code execution, we only used lists of length 3 as the values for the variables \texttt{x} and \texttt{y}. We also always used only the \texttt{suffix} and \texttt{prefix} functions (with one of each appearing in each prompt). We also always placed \texttt{x} as the first variable to be defined. These provisions were to ensure that the withholding that we did applied to consistent linear positions as well as consistent structural positions. For each of the 6 possible list positions (3 in \texttt{x} and 3 in \texttt{y}), we designated 2 letters as withheld for OOD validation and 2 different letters as withheld for OOD testing. No letter was withheld from more than one position.

For passivization and tense reinflection, we always used sentences of the form ``The NOUN PREPOSITION the NOUN who VERBED the NOUN VERBED the NOUN PREPOSITION the NOUN WHO VERBED the NOUN.'' (e.g., \textit{The librarian behind the painter who advised the reporter encouraged the magician near the actor who recommended the photographer.}). Question formation used the same template except with an auxiliary verb before each verb. For passivization and question formation, for each of the six noun positions, we designated 5 nouns as withheld for OOD validation and 5 as withheld for OOD testing; for tense reinflection, for each of the six noun positions, we designated three singular nouns and three plural nouns as withheld for OOD validation, and the same number for OOD testing. No noun was withheld from more than one position.

We trained each DISCOVER model with the relevant task's \texttt{task-specific (all)} role scheme. We used the same traning procedures as described in Appendix~\ref{app:gpt_oss_discover} except that we decreased the learning rate after 5 non-improving evaluations (instead of 20) and halted training at the point that would have brought the 3rd such weight decrease (rather than the 11th), since we found that the more constrained templates led to faster training in these settings than in the previous ones.

\section{Regularization for OOD generalization}\label{app:l21}

\begin{table}[t]
    \centering
    \small
    \begin{tabular}{lcccc} \toprule
        Architecture & Task & Best $\lambda$ & \multicolumn{2}{c}{OOD accuracy} \\ & & & Unreg. & Reg. \\ \midrule
        MLP & Copying & 0.0 & 1.0 & 1.0 \\
         & Reversal & 0.0 & 1.0 & 1.0 \\
         & Interleaving & 0.0 & 1.0 & 1.0 \\ \midrule
         GRU & Copying & 0.001 & 0.77 & 0.98 \\
         & Reversal & 0.00001 & 1.0 & 1.0 \\
         & Interleaving & 0.001 & 0.70 & 0.88 \\ \midrule
         Bottleneck & Copying & 0.001 & 0.99 & 0.99 \\
         Transformer & Reversal & 0.001 & 0.90 & 0.92 \\
         & Interleaving & 0.001 & 0.54 & 0.97 \\ \midrule 
         Transformer & Copying & 0.0 & 1.0 & 1.0 \\
         & Reversal & 0.0 & 1.0 & 1.0 \\
         & Interleaving & 0.0 & 1.0 & 1.0 \\
         \bottomrule
    \end{tabular}
    \caption{OOD validation set results of the hyperparameter search for the $\lambda$ hyperparameter that determines how much to weight the $L_{2,1}$ regularization loss term for letter sequence models. The accuracies are on examples with 2 withheld role-filler pairs. We show accuracies in the unregularized setting (`unreg.'---i.e., $\lambda=0$) and in the regularized setting (`reg.'---i.e., with the best $\lambda$).}
    \label{tab:l21_hypsearch_letter_sequence}
\end{table}

\begin{table}[t]
    \centering
    \begin{minipage}{0.5\textwidth}
    \centering
    \small
    \begin{tabular}{ccccc} \toprule
        Model & Task & Best $\lambda$ & \multicolumn{2}{c}{OOD accuracy} \\ & & & Unreg. & Reg. \\ \midrule
        Gemma (31) & Sentences & 0.001 & 0.00  & 0.86  \\
        Llama-3.1 (16) & Sentences & 0.0005 & 0.00  & 0.90  \\
        OLMo-2 (20) & Sentences & 0.001 & 0.00  & 0.86  \\
         \bottomrule
    \end{tabular}
    \end{minipage}
    \hfill
    \begin{minipage}{0.45\textwidth}
    \centering
    \small
    \begin{tabular}{cccc} \toprule
        Task & Best $\lambda$ & \multicolumn{2}{c}{OOD accuracy} \\ & & Unreg. & Reg. \\ \midrule
        Arithmetic & 0.0001 & 0.08  & 0.43  \\
        Syllogisms & 0.00005 & 0.79 & 0.94  \\
        Code execution & 0.0001 & 0.93 & 1.00 \\
        Passivization & 0.0001 & 1.00 & 1.00 \\ 
        Tense reinflection & 0.0 & 1.00 & 1.00 \\ 
        Question formation & 0.0 & 1.00 & 1.00 \\ 
         \bottomrule
    \end{tabular}
    \end{minipage}

    \caption{OOD validation set results of the hyperparameter search for the $\lambda$ hyperparameter that determines how much to weight the $L_{2,1}$ regularization loss term for LLM period encodings (left) and for GPT-OSS performing symbolic tasks (right). We show accuracies in the unregularized setting (`unreg.'---i.e., $\lambda=0$) and in the regularized setting (`reg.'---i.e., with the best $\lambda$).
    On the left, the accuracies are on examples with 5 withheld role-filler pairs, and on the right the accuracies are on examples with 2 withheld role-filler pairs. In the \textit{model} column on the left, the numbers in parentheses indicate the layer we were analyzing.
    }
    \label{tab:l21_hypsearch_period}
\end{table}

When testing how well DISCOVER can generalize to novel role-filler pairs, there is a potential problem. Specifically, if the role embedding size is at least as large as the number of possible roles and the filler embedding size is at least as large as the number of possible fillers, then it is possible for DISCOVER to effectively learn an independent embedding for every role-filler pair, rather than having any consistency between role-filler pairs that share a role or filler. (To see this, suppose that each role and filler has a 1-hot embedding. Then, their tensor product will yield a 1-hot matrix, with a 1 indicating that role-filler pair and zeroes elsewhere. Then, the weight matrix $W$ that linearly transforms the tensor product is effectively an arbitrary embedding matrix for role-filler pairs). In principle this problem could be circumvented by making the filler and role embedding sizes smaller, but if they are too small, then they will fail to capture the structure of the target model.

A potential solution is to add a regularization term that encourages the role and filler embeddings to be small but without enforcing a specific embedding size. For this, we use an $L_{2,1}$ regularization loss. The basic DISCOVER loss is the mean squared error between the DISCOVER approximation and the target model's encoding vector; this regularization adds two more loss terms, one for the role embedding matrix and one for the filler embedding matrix. The overall loss then becomes the following, where $W_{emb,f}$ and $W_{emb,r}$ are the filler and role embedding matrices and $\lambda$ is a hyperparameter that determines how the loss terms are weighted (note that we fix $\lambda$ to be the same across both $L_{2,1}$ loss terms):
\begin{equation}
    L_{reg} = L_{MSE} + \lambda L_{2,1}(W_{emb,f}) + \lambda L_{2,1}(W_{emb,r})
\end{equation}

\noindent
The $L_{2,1}$ loss takes in a weight matrix and first computes the $L_2$ norm of each row, giving a vector of size equal to the number of rows. It then computes the $L_1$ norm of that vector. The matrices to which we apply this are embedding matrices in which each column is an embedding. This loss thus encourages certain dimensions in embedding space to remain unused: the $L_1$ loss encourages the $L_2$ norms of the dimensions to be 0, and the $L_2$ norm can only be 0 if the entire row is 0, meaning that the dimension corresponding to that row is unused. 

Using this regularization requires us to select a value for $\lambda$. To do this, for each experiment we create an OOD validation split and an OOD test split. That is, the training of the DISCOVER model has certain role-filler pairs withheld from it. A subset of those withheld role-filler pairs are used for OOD validation, while the rest of the withheld role-filler pairs are used for OOD testing (the withheld pairs used for OOD validation and OOD testing do not overlap). We select the $\lambda$ that works best on the OOD validation set and then train models with that $\lambda$ and evalaute them on the OOD test set. See Appendix~\ref{app:ood_details} for details for each model type that we analyze.

\subsection{Letter sequence models}

Table~\ref{tab:l21_hypsearch_letter_sequence} shows the results of our hyperparameter search for $\lambda$ in the letter sequence models. There are some settings in which the unregularized accuracy is at ceiling, but there are others where the regularization makes a major difference (e.g., in the interleaving bottleneck Transformer, raising OOD accuracy from 0.54 to 0.97).

\subsection{LLM period encodings}

Table~\ref{tab:l21_hypsearch_period} (left) shows the results of our $\lambda$ hyperparameter search for LLM period encodings. We ran the hyperparameter search over just 3 of the LLMs (using one layer from each); they all gave similar results (with $\lambda = 0.001$ performing best or similarly to the best $\lambda$), so we adopted that value for all other models and layers. (In the table, note that 0.001 is not the best $\lambda$ for Llama-3.1, but 0.001 performs nearly as well as the best $\lambda$: 0.896 vs.\ 0.898. Thus, since 0.001 seems strong across settings, it is the one that we adopt in general. 
For these experiments, the regularization seems to be critical, making a qualitative difference in DISCOVER generalization performance (Table~\ref{tab:l21_hypsearch_period}, left).

\subsection{GPT-OSS performing symbolic tasks}

Table~\ref{tab:l21_hypsearch_period} (right) gives the results of searching over $\lambda$ for applying DISCOVER to GPT-OSS's representations of symbolic inputs. We ran this search separately for each of the six tasks.

\section{Generalization to novel role-filler pairs in white-box models}\label{app:ood_whitebox}

Our experiments with generalization to novel role-filler pairs are built on the assumption that, if DISCOVER successfully generalizes to novel role-filler pairs, this is evidence that the target model's representations incorporate systematic binding. 
This assumption can be supported solely from principle (DISCOVER cannot generalize to something unseen unless there is some sort of systematicity that it can leverage to determine how to handle the unseen units). However, as an additional source of support, here we also run an empirical experiment to validate this assumption.

The way we test this assumption is to run DISCOVER on two types of white-box models (where a white-box model is one whose internal structure is known, in contrast to the more standard situation in which neural networks are black boxes). We use one type of white-box model that we know has systematic role-filler binding and one type of white-box model that we know does not. The one that has systematic role-filler binding is a sequence-to-sequence model with an encoder that is explicitly a TPR, and a decoder that is a Transformer; the one that does not have systematic role-filler binding is a sequence-to-sequence model with an encoder that has an atomic embedding for each role-filler pair (which we call an Atomic Pair Encoder) combined with a decoder that is a Transformer. The Atomic Pair Encoder's representation for an input is simply the sum of its embeddings for the role-filler pairs in the input.

There is one wrinkle: It is possible that, through training, the Atomic Pair Encoder would coverge to systematic embeddings. That is, even when the architecture is not structured systematically, it is possible for learned embeddings to converge to systematic structure due to training pressures \citep{mikolov2013linguistic}.
Therefore, we leave the encoders of these two models untrained---the TPR Encoder and the Atomic Pair Encoder are left with the weights that they have at random initialization, with the Transformer decoders trained to read out from these frozen encodings. It is still possible that the Atomic Pair Encoder would have systematic structure through astronomical good luck, but that is so unlikely that we are comfortable dismissing the possibility.

We first train these sequence-to-sequence models on the task of copying, and both architectures perform nearly perfectly, with accuracies above 0.99 for all 10 reruns for both architectures. 

\begin{figure}[t]
    \centering
    \includegraphics[width=0.6\linewidth]{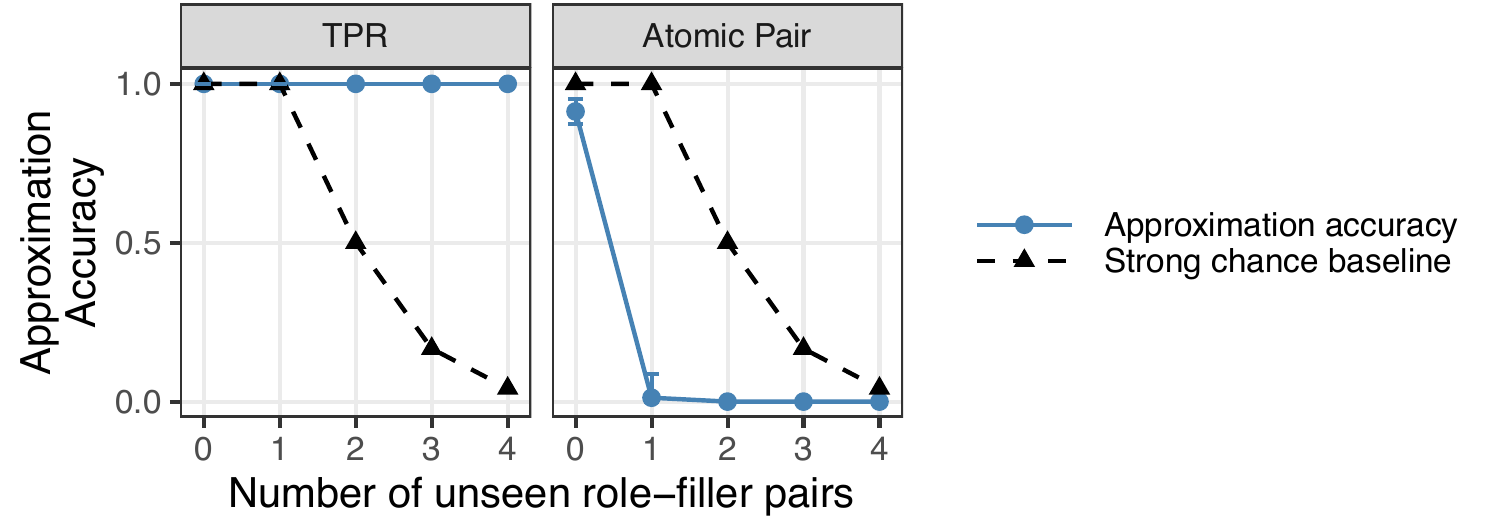}
    \caption{DISCOVER performance on two white-box models: one that is known to have systematic role-filler binding (``TPR'') and one that is known not to have systematic role-filler binding (``Atomic Pair'').}
    \label{fig:whitebox}
\end{figure}

We then apply DISCOVER to the two sequence-to-sequence architectures, training it with withheld role-filler pairs in the same way as the black-box models in Section~\ref{sec:ood}. The results are in Figure~\ref{fig:whitebox}. Two points about these results are worth noting. First, in the cases where there are no unseen-filler role pairs---i.e., where DISCOVER only needs to handle role-filler pairs that it has been trained on---DISCOVER works well for both architectures, even the one that has no systematic binding. This illustrates that we indeed cannot draw strong conclusions about the presence of systematic binding based on in-distribution DISCOVER results, since this plot gives an existence proof of DISCOVER giving high accuracy even when the target model definitely lacks systematic binding. 
Second, however, we also see that testing for generalization to novel role-filler pairs addresses this concern: We see strong generalization for the white-box TPR model that is known to have systematic binding, while we see a complete failure to generalize in the white-box Atomic Pair model that is known not to have systematic binding. Thus, this plot supports our methodological decision to use generalization to novel role-filler pairs as the key type of evidence for systematic role-filler binding.

\section{Is DISCOVER performance merely driven by the number of roles?}\label{app:rolecount}

Many of our results involve comparing multiple role schemes, where some role schemes work better than others. 
Our interpretation of the results typically assumes that the extent to which a role scheme provides a close approximation quantifies the extent to which the role scheme matches the implicit structure of the representational space being analyzed.
However, there is an important confound to address: In many cases where role scheme A outperforms role scheme B, role scheme A also involves a larger vocabulary of roles than role scheme B, meaning that the DISCOVER model for A has more parameters than the DISCOVER model for B. Since machine learning models typically perform better when they have more parameters, differences in the number of roles raise the concern that DISCOVER performance might merely be driven by the number of roles rather than by a role scheme's structural fit to the target model.

Luckily, there are many data points that provide reason to reject this possibility. First, consider the letter sequence results in Figure~\ref{fig:letter_seq_tpe_acc}. For each role scheme used in those results, Table~\ref{tab:letter_seq_rolecount} (left) shows how many roles are used (the number of roles is the same across all architectures and tasks). In every condition in Figure~\ref{fig:letter_seq_tpe_acc}, \texttt{bidirectional} roles outperform \texttt{Wickelroles} despite using far fewer roles (21 vs.\ 729). Note that \texttt{Wickelroles} are not quite sufficient to uniquely identify every possible sequence (e.g., \texttt{A, A, B, A, A, A} and \texttt{A, A, A, B, A, A} both receive the same \texttt{Wickelrole} representation). This factor is unlikely to fully explain the low performance of Wickelroles, since there are cases where it produces a fairly high DISCOVER accuracy (such as the reversing bottleneck Transformer), showing that at least in principle it can score reasonably high on DISCOVER but usually does not; however, the other examples in this section involve comparisons where this issue is not relevant because all role schemes considered are sufficient to uniquely identify the sequence in question.
Further, \texttt{left-to-right} and \texttt{right-to-left} roles use the same number of roles (i.e., 6), yet they often show substantially different approximation accuracies; in this case, both are sufficient to uniquely identify the sequence.

\begin{table}[]
    \centering
    \begin{minipage}{0.35\textwidth}
    \centering
    \begin{tabular}{cc} \toprule
        Role scheme & Number of roles  \\ \midrule
        Left-to-right & 6 \\
        Right-to-left & 6 \\
        Bidirectional & 21 \\
        Wickelroles & 729 \\
        Bag-of-words & 1 \\ \bottomrule
    \end{tabular}
    \end{minipage}
    \hfill
    \begin{minipage}{0.6\textwidth}
    \centering
    \begin{tabular}{ccc} \toprule
        & Bidirectional (all) & Task-specific (all) \\ 
        Task & role count & role count \\ \midrule
        Arithmetic & 94 & 185 \\
        Syllogisms & 1,956 & 1,772 \\
        Code execution & 1,329 & 3,771 \\
        Passivization & 993 & 1130 \\
        Tense reinflection & 993 & 1130 \\
        Question formation & 1038 & 1283 \\ \bottomrule
    \end{tabular}
    \end{minipage}
    \caption{Number of roles for the DISCOVER models trained in the letter sequence experiments (left) and for the \texttt{bidirectional (all)} and \texttt{task-specific (all)} role schemes used in DISCOVER approximations of GPT-OSS (right).}
    \label{tab:letter_seq_rolecount}
\end{table}

Now consider the period encoding results in Figure~\ref{fig:tpe_accs_period}. In the lists condition, the \texttt{bidirectional} role scheme uses 12 roles while the \texttt{predecessor} role scheme uses 301 roles, yet \texttt{bidirectional} consistently outperforms \texttt{predecessor}. Note that, in this setting, we use lists in which all elements are unique, so the \texttt{predecessor} role scheme does in principle provide enough information to uniquely identify the sequence being encoded. 

Finally, consider the GPT-OSS results in Figure~\ref{fig:llm_tpe_acc}. Table~\ref{tab:letter_seq_rolecount} (right) shows the number of roles used in the two competitive role schemes for each task. In all cases, \texttt{task-specific (all)} achieves a stronger DISCOVER approximation than \texttt{bidirectional (all)}. Nonetheless, in the syllogisms task, \texttt{task-specific (all)} uses fewer roles; further, in code execution, \texttt{task-specific (self)} (99 roles) slightly outperforms \texttt{bidirectional (all)} (1,329 roles) despite using substantially fewer roles. Thus, the differences in DISCOVER accuracy cannot be fully explained by the number of roles available---though it is indeed the case that, in the 5 of the 6 conditions, the better-performing \texttt{task-specific (all)} has more roles than the worse-performing \texttt{bidirectional (all)}.

As one other relevant point, the main reason to be concerned about the settings where \texttt{task-specific (all)} uses more roles than \texttt{bidirectional (all)} is that it raises the risk that we are misinterpreting the results by concluding that \texttt{task-specific (all)} better reflects the structure of the target model; what if the target model's structure actually follows \texttt{bidirectional (all)}, but in a way that is masked in our analyses by the smaller role count for that role scheme than for \texttt{task-specific (all)}? This concern is particularly acute in the linguistic tasks (passivization, tense reinflection, and question formation), since in those conditions \texttt{task-specific (all)} achieves only slighter better approximation accuracy than \texttt{bidirectional (all)}. However, for two of the linguistic tasks, we have an alternative source of evidence---the structure-sensitivity of causal interventions (Appendix~\ref{app:intervention_structure_sensitive})---as evidence that these representations are indeed driven more by the task-specific (i.e., syntactic) structure than by the task-agnostic \texttt{bidirectional} structure.

Overall, then, we conclude that differences across role schemes in DISCOVER approximation accuracy are not primarily driven by the number of roles used, supporting the type of inference drawn throughout the paper (namely, that DISCOVER approximation accuracy with a particular role scheme indicates the extent to which that role scheme captures the causal structure of the target model's representations).


\end{document}